\documentclass{article} 
\usepackage[small]{caption}
\usepackage{graphicx}

\usepackage[american]{babel}

\usepackage{natbib} 
\usepackage{mathtools} 
\usepackage{booktabs} 
\usepackage{tikz} 

\usepackage{subfigure}
\usepackage{float}
\usepackage{hyperref}
\usepackage{color}
\usepackage{algorithm}
\usepackage{algpseudocode}
\usepackage[]{mcode}

\title{Model of models - Part 1}

\begin{document}

\maketitle

\begin{center}
	\author{\href{mailto:<cm5099@yahoo.com>}{Shimon Komarovsky}{, Technion, Haifa, Israel}}	
\end{center}

\begin{abstract}
  This paper proposes a new cognitive model, acting as the main component of an AGI agent. The model is introduced in its mature intelligence state, and as an extension of previous models, DENN, and especially AKREM, by including operational models (frames/classes) and will. This model's core assumption is that cognition is about operating on accumulated knowledge, with the guidance of an appropriate will. Also, we assume that the actions, part of knowledge, are learning to be aligned with will, during the evolution phase that precedes the mature intelligence state.
    
  In addition, this model is mainly based on the duality principle in every known intelligent aspect, such as exhibiting both top-down and bottom-up model learning, generalization verse specialization, and more.  
  Furthermore, a holistic approach is advocated for AGI designing and cognition under constraints or efficiency is proposed, in the form of reusability and simplicity.
  
  Finally, reaching this mature state is described via a cognitive evolution from infancy to adulthood, utilizing a consolidation principle.
  
  The final product of this cognitive model is a dynamic operational memory of models and instances.
  
  Finally, some examples are presented, and some preliminary ideas for the evolution phase to reach the mature state.
\end{abstract}

\section{Introduction} \label{sec:MOM_intro}



Our consistent goal is to construct a basic realistic model for \textit{AGI} (Artificial General Intelligence). 
First we illustrate how AGI is perceived by us. It is this data processing tool, without any awareness or consciousness. In addition, it thinks, perceives and acts as human. It also have different channels for input and output, again as human. All for the purpose to make an agent that understand humans and humans understand it.
To accomplish that, there exist many ways to approach the \textit{AGI} problem, e.g. via emulation.
Although, a human brain emulation (copying the brain into digital form) is a good idea, there are some deficiencies to this approach:
\begin{enumerate}
	\item the brain is a very complex organ, especially when it is accessed at its mature state.
	\item there is no direct access to the human experience (consciousness) with correlation to the brain activity (or neural processing). It is mostly performed in a very indirect process, like questioning a human participator. Also it is done in a very coarse resolution of the brain (large clusters of neurons).
	\item it is difficult to model the brain, since it works throughout the whole network. Hence, simple models cannot be extracted out of it by local isolation. See more in \url{https://medium.com/nontrivial/neuro-nonsense-2acc209d42c3}
\end{enumerate}

Hence, here the \textit{AGI} problem is tackled from a human design direction instead, though neuroscience can act as a source for ideas and inspiration \citep{zhao2023brain}.
This design is a gradual process with many versions along the way.
Therefore, this paper presents \textit{MOM} (Model Of Models), the next version of \textit{AKREM} (Associative Knowledge Representation). See its short version \citep{10.1007/978-3-031-19907-3_6} or its full version \citep{EasyChair:7921}. This paper is actually the full version of a preliminary short paper \citep{komarovsky2023purposeful}.

\textit{AKREM} is a mature-state knowledge representation model, based mainly on the assumption that communication is about encoding the sender's will into a sequence of words (a message), and then decoding it by the recipient. The model proposes a representation of any message, in a hierarchical form based on grouping, by generating some essence in a given level, from details in the lower level. The lowest level details are founded upon some DNN (Deep Neural Network), generating the basic concepts and actions (from which the details are made of) from unstructured input.

\textit{AKREM} is also backed 
by neuroscience, such as the idea of memorizing techniques such as the Memory Palace, which turns random objects into memorable ones by converting them into some consistent sequence or tale. Also, \citep{hahamy2023human} 
supports the idea that story or any message is partitioned into micro-events, which are also inter-connected to allow comprehending current part of the story based on previous parts of it (a phenomena called "replay").

Finally, while the will concept exists in \textit{AKREM}, it will be expanded upon in this paper.

The will is very fundamental to our intelligence. It is the source of everything. It dictates our actions, our interpretation of reality, thus affect the meaning of things to us. Nevertheless, it is almost not mentioned in AI literature. Our main cognitive will is to comprehend the reality, in order to manifest other types of will in it. More generally, it is expressed via problem-solving state of mind, in which we react in reality. Nevertheless, will can be also affected by reality.

Furthermore, we assume the will is a field, similar to the "will force" phrase, i.e. it is represented as vector in some state space. Then, in order to affect reality, this will has to create some connection to knowledge. We assume that this connection is learned through experience, and it is manifested by alignment with the actions operating on objects that comprise the knowledge. This way, whenever we have a will, the appropriate actions are triggered, similarly to when we want to raise our arm - the appropriate physical actions operate.

However will is an intriguing phenomenon. It can be realized via laws of physics, or derived from emotions, or even derived deeply from high (moral) values, such as justice. This makes the full knowledge model of anything, as observed from outside, very stochastic, due to these hidden factors.

Following this, new additions to the presented model, including new associations, operationability, modeling, consolidation, and reusability, are introduced.
First, while \textit{AKREM} assumes that the learned \textit{elements} are either objects or static actions (verbs), new associations are introduced: object's attributes and relations. Next, these connections are all static representations of knowledge, i.e., the hierarchies cannot be changed.
Therefore, operationability introduces a new type of association to objects: actions that act upon them, thus producing new knowledge \textit{elements}. This makes the connections in \textit{AKREM}'s hierarchies dynamic, hence it allows the freedom to update and create new hierarchies.
Next, modeling introduces some basic cognitive operations, e.g. abstraction and grouping\footnote{Philosophically, these operations drive us to theory of everything, and religiously to the notion of God.}. Both gather many details into fewer. Grouping specifically is about connecting \textit{elements} via some common property. It could be for example a chronology in a plot or other common properties/actions grouped into classes.
Finally, consolidation is a process in time, that collapses a huge amount of possibilities into small set of patterns, of any kind.

All the operations above are considered to be bidirectional, i.e. everything lies in some range between extremes, i.e. everything has its inverse (dualism)\footnote{Can also be regarded as symmetries or invariants}. In grouping, it is from the whole to its parts and vice versa, and in abstraction, it is from instances to classes and vice versa. Consolidation is an operation in time, creating models and memory, while its inverse is forgetting, which is also an operation in time.
Will also lies in the dichotomy of determinism and randomness.

Moreover, if in the early epoch of AI, symbolic reasoning was dominant, and nowadays connectionism dominates, then we come to a new era \citep{sheth2021duality}, where we should combine and include many conflict perspectives, in cooperation and competition. Our holistic perspective embrace this duality and other dualities also \citep{reggia2013rise}. An exhaustive but not complete list is presented:
\begin{itemize}
	\item top-down and bottom-up model learning, 	
	\item consolidation verse forgetfulness, 
	\item generalization verse specialization, 	
	\item grouping and dismantling,
	\item determinism and randomness,
	\item problem-solving verse designing perspectives,
	\item connectionism verse symbolism,
	\item soul and brain (idealism),
	\item convergent verse divergent thinking, 	
	\item object-oriented verse functional programming paradigms,
	\item past verse present perceptions,
	\item induction verse deduction,
	\item recognition verse creativity,
	\item serial thinking verse parallel thinking (as described be de Bono, e.g. in \citep{de2016parallel}),
	\item left-brain verse right-brain theories,
	\item neuron excitation verse inhibition,
	\item exploration verse exploitation, and more.
\end{itemize}


The product of these cognitive operations and dualities is a dynamic memory of models, formed as a semantic network of \textit{elements}. It is dynamic by being evolving through time, by including spatio-temporal modeling, and by being operational, i.e. by possessing the possibility to update the models.
Additionally, it encourages a holistic approach for \textit{AGI} designing: one simple system for multiple functions, such as short-term memory (STM) and long-term memory (LTM), problem-solving, communication, learning, and any cognitive function. This approach is also encouraged in neuroscience, i.e., though the brain seems complex, but its basic mechanism is simple \citep{mountcastle1997columnar}. This approach is especially needed when resources are limited. By using this approach, our \textit{MOM} puts everything in one place, instead of common cognitive architectures (CAs) approach to separate into modules. Here there is a dynamic memory, which is also used for thinking, planning, imagining, and more. While the Working Memory (WM) is simply loading old memories of LTM "into the surface".

Following this dynamic representation of knowledge - its evolution is presented. That is, operational modeling presents \textit{AGI} agent's intelligence in its mature state, which is the state of how its knowledge should be represented. However, to accomplish this state, a cognitive evolution over time is required, which utilizes the consolidation principle.
Along with this primary consolidation process, that act in many different things, there are also some fundamental learning approaches. These approaches are the learning from examples and the learning from logic, use consolidation as a tool, to reach the desired mature state we present in this paper.

\begin{figure}[H]
	\centering
	\includegraphics[width=0.7\textwidth]{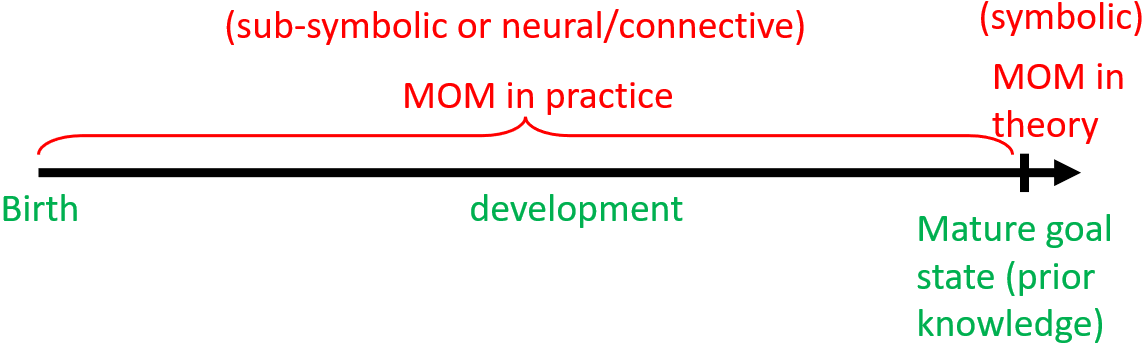}
	\caption{\textit{MOM} two parts.}
	\label{fig:MOM_2parts}
\end{figure}

The two parts of the \textit{MOM} cognitive model are illustrated in Fig.~\ref{fig:MOM_2parts}. They represent the duality above, especially our Neuro-Symbolic approach.
\textit{MOM} in theory includes objects, relations, actions, abstractions, etc – mostly logic and classic AI.
On the other hand, \textit{MOM} in practice based mostly on \textit{DL}, and includes concepts/principles like consolidation, attention, splitting \& merging as in \textit{DENN}, and more.

Eventually, this framework has different approach to learning compared to usual \textit{DL}, such as Large-Language-Models (LLMs), in the journey to reach \textit{AGI}. \textit{DL} based on learn-everything-at-once kind of approach, or batch learning. In this approach the expert AI agent supposed to learn everything mixed up with different topics and complexities. Hence, the way to reach \textit{AGI}, is by larger models and data. Conversely, we support incremental curriculum type of learning as in humans. The learning is built in layers or stages. When the current layer is established, then and only then that we can proceed to a new layer on top of the previous layers. Usually it is done by extending analogies (distant connections), which can be regarded as breadth extension, and abstraction, which can be regarded as extension in height. 

Finally, after establishing a firm theoretical framework of \textit{MOM}, a proposed implementation is presented. That is, the implementation of the evolution phase that strive to reach the mature intelligence state. It is based on the principle of model separation/isolation, then the use of attention splittability feature to realize this, then the use of multi-version to allow flexibility in learning, finishing with program-search to model methods of consolidated classes/objects.

\subsection{Motivation and Problem definition}

\subsubsection{General motivation}

The general motivation is to reach \textit{AGI}. We consider LLMs as the first step towards this goal. However, the deficiencies of these LLMs is what inspire us to succeed where they fail. For example, ChatGPT performs quite well in many tasks, hence GPT can be regarded as general purpose AI model, and can generalize to new tasks. However, it hallucinates with confidence, and do not perform well in reasoning tasks, that require logic and step-by-step thinking. One of the reasons for this, is that masking or next word prediction is not enough as a prior knowledge induced in the neural network architectures, to align with humans, even though some studies in neuroscience showed correlation of brain activity and LLMs.

Also LLMs do not have a world model (unlike JEPA \citep{dawid2023introduction}). They simply mimicking text or images (like Midjourney) \url{https://medium.com/@ignacio.de.gregorio.noblejas/world-models-the-next-frontier-in-our-path-to-agi-is-here-ecab17042d1e}.
Although some papers use probing of the LLM, to prove that some sort of model/representation is learned in the LLM.

Many papers have tried to improve reasoning. They used many methods, e.g. Chain-of-thought (CoT) prompting, self-consistency, role prompting, voting by agreement, Tree-of-Thoughts (ToT), Graph-of-Thoughts (GoT) \citep{besta2023graph}, and other prompting techniques. However, these do not show ultimate solution. Another example, in \citep{ling2023deductive}, they perform CoT but with accompanying an evaluation for each step. Specifically, perform a deductive reasoning through evaluating each step. In deductive reasoning, a (premise, conclusion) pair is given, and the goal is to determine whether the conclusion follows from the premises. This method demonstrates also the need in a much stronger prior, since the proposed method searches for “fixes” to the CoT method, though the CoT method in LLMs is fundamentally wrong approach to model reasoning.

Additionally, there was an effort towards alignment with human intentions, e.g. via Reinforcement-Learning through Human Feedback (RLHF) in LLMs as GPT. However, it is mostly used for constraining the ChatGPT replies, e.g. being polite, filter out harmful requests, being unbiased, avoiding hateful, violent, or rude speech and more.


A recent paper by Google Brain \citep{schuurmans2023memory} proves how adding a memory to LLMs can produce Turing complete machine, i.e. simulate any algorithm. Similarly, is the inclusion of vector databases with LLMs. This demonstrates that memory is definitely part of the needed prior, but it is not all of it.

In our opinion, this prior should integrate different brain’s cognitive capabilities/functions. Such a prior would allow the agent to learn and act more rationally, thus produce the most valuable trait, that's missing in today's (mostly black-box) AI: explainability (through external communication with it. Not by analyzing its internal model).
Hence, a more promising approach we advocate for and try to apply, is the hybrid approach of Neuro-Symbolic AI. It is a sub-field of AI, where the strengths of the two disciplines are combined to overcome their weaknesses.

Finally, the lack of understanding affects also vision comprehension, which results sometimes in a workaround like LIDAR sensors in autonomic driving vehicles, see \url{https://futurism.com/the-byte/elon-musk-furious-autopilot-tried-kill-him}.

\subsubsection{Specific motivation}

This paper is the continuation of the previous paper's model \textit{AKREM}. The context of the previous and the current paper is the same: assuming some module that converts sub-symbolic data to symbolic one and vice versa. In \textit{AKREM}, it is a temporal hierarchical DNN, that processes the incoming data from senses via different temporal rates, where the slowest processing layers suppose to deal with thinking, then finally produces the appropriate output.

In this paper however, this module is more vague. On the one hand it can be part of the whole system proposed in the paper: sub-symbolic to symbolic and vice versa can use the same mechanism of learning as described in Section~\ref{sec:Learning_the_modeling}. On the other hand, we can assume an external module similar to \textit{AKREM} settings, where some DNN, either flat or also temporal hierarchical, is responsible for some fundamental processing of incoming and outcoming raw data. During this processing neurons are triggered in the DNN. These neurons stimulate classes of objects and actions, representing agent's knowledge map. This is where a transition from connectionist AI to classic/logic AI occurs. 

Our main goal in this paper is to concentrate on the logical part, or the knowledge map, to model the thinking/cognitive processes, as it is in our opinion the bottleneck for \textit{AGI}, or the more urgent issue in AI nowadays, such as it appears in the problems revealed in LLMs.

This modeling, \textit{MOM}, presented as the mature state of an intelligent agent, is described in details in Sections~\ref{sec:Associations}, \ref{sec:OperationalModeling}, \ref{sec:Examples}.

Of course, this model is only the ideal intelligence state, hence how to reach this state is discussed in Section~\ref{sec:consolidation}.

Additionally, we figured out that thinking is very dependent upon will, either the will encoded in the perceived message or the agent's own will. This topic is elaborated in Section~\ref{sec:will}.

Finally, we define \textit{thought} as operating in the knowledge map, e.g. by inferring class features/attributes, or by performing actions on objects, or by overall supervising the thought process.

\subsection{Contribution}

The contribution of this paper is expressed within two comparisons: the comparison of the proposed model to its previous versions, see Section~\ref{sec:Cognitive_model_comparison}, and the comparison to other CAs, see Section~\ref{sec:Comparison}.

\textit{What we propose:} In essence, what is presented in this paper is a cognitive model, or a model describing how thinking works. In this model we describe how different thinking processes occur, including learning and knowledge representation.
Even if it is not how our brain works, it is still can be utilized to build an AI agent, like other virtual assistants: ChatGPT or MIT's ELIZE or Amazon's Alexa or Apple's Siri. These assistants are communicating with people and perform different tasks.

\textit{What exists:} From the early days of AI, there are expert systems which are based on classic AI, that is mainly based on logic. Nowadays, there are LLMs that are based on Machine Learning (ML) but more specifically on Deep Learning (DL).
Each of them has advantages and disadvantages. In one relevant aspect to us they differ: the perception of will and intention throughout communication. The logic-based systems are based almost solely on cold and precise logic. This is mechanical and technical perspective on intelligence. The deep-learning models are usually do learn will and intention from data. However, it is done very implicitly\footnote{It can be learned when will is expressed in the data, e.g for data the includes will in its input while the output is the appropriate response. Therefore, LLMs for example are good at task generalization, while image label/caption classifiers are not. See good discussion about it in \url{https://medium.com/swlh/what-makes-neural-networks-fragile-676fe7cf230a}.}, especially lack the logical structure prior, hence cannot be generalized to any random and flexible conversation. It makes these both approaches lack of common sense and especially of full understanding and comprehension of the situation and of the requests from the human user, in both directions. Both in understanding human's intent and in replying, in human-like logical way.

\textit{What is the novelty here:} The new thing compared to existing approaches in AI, is that we explicitly learn will in the models. The models that we construct in our mind while interacting throughout our lifetime: from inanimate objects like basket and table, through animals, humans, upto abstract theories.

This will becomes essential part of any model, and in order for it to be part of our physical reality and affect things, it is aligned with our knowledge, be it in a logic-based form or embedded in neural networks.

This alignment is useful, since it creates intuition between problems and admissible actions. It provides kind of heuristics for solving new problems.

\textit{MOM's assumptions:} There are many assumptions or fundamental axioms in \textit{MOM}, similarly as in any model. Some of these are embedded and still hidden, while others are more explicit and scattered in this article. Some of them are also expressed within the different definitions. Here are some assumptions:
\begin{itemize}
	\item Causality and action affecting the environment: basic assumption here is that actions applied from any object or living thing are affecting things in the environment. Thus causality is derived. Moreover, due to this assumption we do not necessary need explicit RL mechanism for learning.
	\item Will and knowledge: the assumption here is that will and knowledge are two separate entities. However, they can be aligned for effective operation of the agent, to accomplish its goals.
	\item Multiple modalities: the original vision and audio channels are assumed. But eventually it can be extended to any other modalities, such as digital domain (sending/receiving data files), action domain (what is/was done), and more.
	
	Action domain means that we can input the AGI agent also full experiences, skills, without actually go through all the steps and all the training it involves. Similarly, instead of sketching on a piece of paper, hand-waving and long explanations, the digital domain can transfer whole books, presentations, images, and any digital file. Perhaps since this domain is so general, it can also contain action domain within it.
\end{itemize}

\section{Timeline}

The following Sections are ordered like this:
\begin{itemize}
	\item Section~\ref{sec:will} starts from some terminology, then discuss will as it was presented in previous model: \textit{AKREM}, then continues with will expression in a constrained environment, especially in problem solving and design, then presents will as a field and its consequences/implications, and then finishes with different levels or types of will.
	\item Section~\ref{sec:Associations} describes briefly the new Associations added to our current model.
	\item Section~\ref{sec:OperationalModeling} is about operationability and modeling, two extensions of the previous \textit{AKREM} model. First it describes them both as simple classes/frames, then details about each of them in our framework. Finally, few definitions are introduced and then will derivation is discussed.
	\item Section~\ref{sec:Examples} is engaging in further details about \textit{MOM}'s knowledge structure, and then displaying several examples of \textit{MOM} model.
	\item Section~\ref{sec:MOM_evolution} discusses temporal processes of consolidation in its multi-faced forms, reusability principle as an additional form of consolidation, and finishes with learning approaches.
	\item Section~\ref{sec:Iplementation} proposes one possible implementation of ideas presented in previous chapters. It is founded upon the model separation capability, then the utilization of attention mechanism to realize this capability. Then the multi version for flexibility is used, and finally the program search for the actions is discussed.
	\item Section~\ref{sec:PriorWork} is reviewing relevant literature in different topics, e.g. Neuro-symbolic AI, CAs, and hierarchical planning and AI. Finally it 
	is comparing previous models with the current proposed \textit{MOM} model.
	\item Section~\ref{sec:Conclusion} is a summary of the paper including some key takeaways, and later 
	discusses future work.
	\item Finally, Appendix~\ref{sec:appendix} includes: discussion of advantages for adopting a hybrid approach for AGI; how will can be learned; existing knowledge representation compared to \textit{MOM}; the issue of Bias in any intelligence; a the role of language in the AGI background; DNN's equivalence to programming languages; general evolution in \textit{AKREM}; the aspects of Logic and Creativity in \textit{MOM}; and AGI additional characteristics.
\end{itemize}

Note that the paper is not so rigorous as any serious academic paper. It is mostly philosophic, since this theory is foremost presented for the sake of acting as a starting point for discussion and adaptations. It is not that important to define every detail in the theory just yet. Nevertheless, along the paper few definitions are spread, to ease on the reader, i.e. to make things more clear and less ambiguous.

Either way, the reader should come open-minded, since this is not technical paper. We invite and welcome notes, suggestions, reviews, corrections, and any type of comments and discussion.

\section{Will} \label{sec:will}

In this section, the importance of will as an essential element in human intelligence\footnote{One clue for this found in \url{https://theconversation.com/great-mysteries-of-physics-5-will-we-ever-have-a-fundamental-theory-of-life-and-consciousness-203127}, where they try to characterize how the living different from non-living structures. They show that living structures for some reason can defy entropy and can produce even more complex structures through evolution, while non-living cannot. This may indicate the external factor of will is what can cause this difference.} is elaborated upon, starting from some general motivation, then continuing from the previously presented model, \textit{AKREM}. The detailed discussion about how will is expressed in a constrained environment is performed, via problem-solving setting. Finally, some basic types of will are presented.


We first start with some terminology, so that the reader would have clear context of the discussed in the rest of this chapter and other chapters.

\subsection{Terminology} \label{sec:terminology}


First we collect the relevant interpretations from the dictionary:

Will = power of the mind in choosing/controlling one's own actions/emotions.

Goal = the result or achievement toward which effort is directed.\\
Purpose = the reason for which something is done, exists, created, made, used, etc; something set up as an object or end to be attained; a subject under discussion or an action in course of execution; an intended or desired result; end; aim; goal.\\
Objective = something that one's efforts or actions are intended to attain or accomplish.
Intention = an act or instance of determining mentally upon some action or result; a determination to act in a certain way; the end or object intended; meaning or significance; purpose.

Similarly, our definition of will is the source of actions.
We refer purpose as a goal. However, its other interpretations such as a reason or a meaning of something and an action in course of execution are related. Since it is like describing the how it got to be, or the will that derive it, or the will in the process.

And taken from social studies \citep{marco1996terminology, gollwitzer1993goal, reker1987meaning}:  

Will = refers to an individual's mental faculty or capacity to make choices and decisions. It involves the conscious and deliberate exercise of one's volition or intention in order to initiate or refrain from certain actions. Will is often associated with personal agency and the ability to exert control over one's behavior, thoughts, and emotions.\\
Goal = a desired outcome or result that an individual or a group of individuals strive to achieve. It represents a specific target or destination towards which efforts are directed. Goals can be short-term or long-term and can be related to various aspects of life, such as personal development, career aspirations, relationships, health, or academic achievements. They provide a sense of direction and purpose, guiding individuals' actions and decisions.\\
Purpose = a sense of meaning or significance in one's life. It involves having a clear understanding of one's values, passions, and overarching aims. Purpose goes beyond specific goals or achievements and often involves a deeper reflection on one's identity, values, and contribution to the world. It provides a sense of fulfillment and a motivation to engage in activities that align with one's purpose.\\
Objective = a specific, measurable, and time-bound step or milestone that is designed to contribute to the attainment of a larger goal. Objectives are concrete and quantifiable, allowing individuals or organizations to assess progress and success. They serve as the building blocks that help individuals or groups move closer to their overall goals.\\
Intention = a conscious and deliberate mental state characterized by a planned or anticipated course of action. It involves the formation of a specific purpose or objective in one's mind and a commitment to act in accordance with that purpose. Intentions can range from simple everyday actions to more complex and long-term endeavors. They play a crucial role in guiding behavior and directing efforts towards the achievement of desired outcomes.

Interestingly, these definitions are quite different from the dictionary's ones. Here they are more human-inclined or social, while the dictionary is more general.

Finally, note that we use \textit{logic} most often, throughout this paper. However, we mean the most basic type of logic, that is involved in all cognitive operations. This is not includes the highly abstract rigorous mathematical/scientific logic, which involves proofs, exact definitions, and more. This is because we assume that this type of logic is acquired or learned. I.e., not embedded as a necessary element, in our cognition throughout its all different forms.

\subsection{Motivation}
\begin{figure}[H]%
	\centering
	\subfigure[Single action, e.g. in lower level]{%
		\label{fig:a}%
		\includegraphics[width=0.45\textwidth]{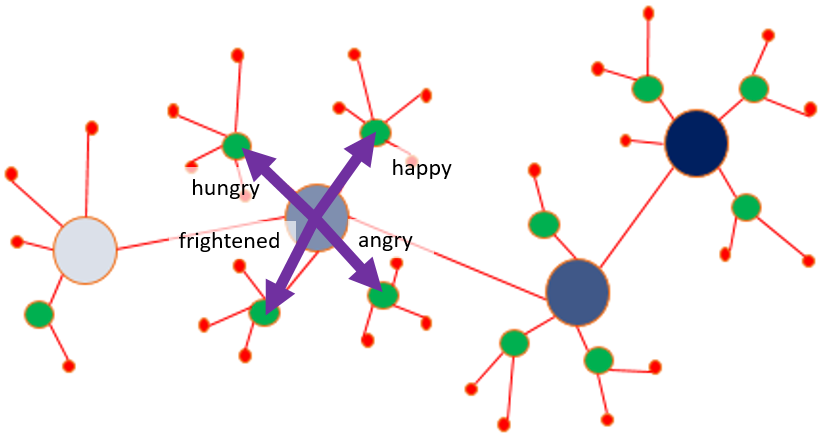}}%
	\hfill
	\subfigure[Group of actions, e.g. in higher level]{%
		\label{fig:b}%
		\includegraphics[width=0.45\textwidth]{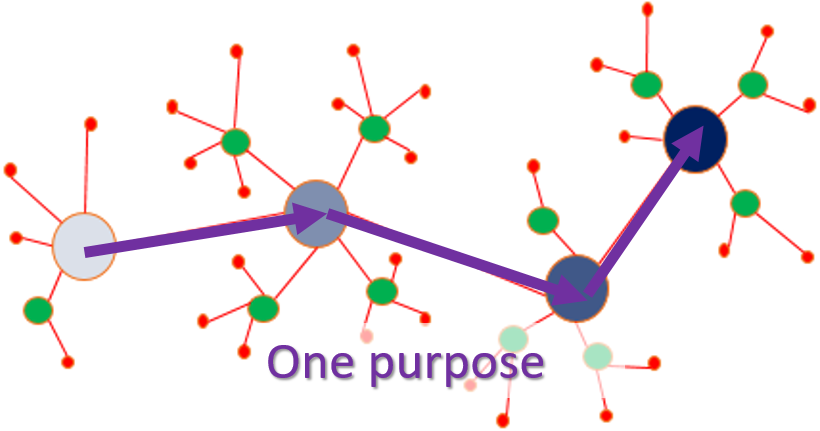}}%
	\caption{Actions representing a meaning/purpose}
	\label{fig:single_verse_multiple_actions}
\end{figure}

One aspect of will, is that it determines our behavior, or more precisely - our actions.
Will derives or creates somehow a set of actions, i.e. like a plan, for example a will causes for constructing an algorithm such as 
search in a knowledge base or in an \textit{AGI}/brain network.
Inversely, will can be inferred from an action or from a sequence of actions. This process of producing actions and retrieving their original will is expressed in the sequence encoding and decoding in \textit{AKREM}, respectively. As seen in Fig.~\ref{fig:single_verse_multiple_actions}, will or purpose is expressed both in single action (individuality), and in a sequence of actions or a trajectory (grouping). 
More generally, it is in the different levels of \textit{AKREM}.

Another aspect of will, is its affect on our knowledge.
Philosophically, will + objective data produce meaning and bias \citep{nietzsche1974gay}, or more generally perception and interpretation. Or:
\begin{equation}
	\text{will + objective data = subjective data.}
\end{equation}
In other words, meaning is not an objective concept in non-human reality, but rather it is "tailored" specifically to humans, or more accurately it is a function of will - or derived from it. I.e. it is related to the function that the concept has to the human.
For example, "cat" perceived as a pet while wild cat excluded from this concept since it is not under human control. Similarly, any tool like a hammer or a screwdriver has a meaning as a function for fulfilling some will, and not just as a particular shape and matter properties. Along with this, it is natural that meaning would have a bias. Hence, the bias and fairness which researchers battle with, is impossible to overcome, since there will always be some bias, i.e. some preferred opinions over others, again - as a function of the will. Best thing we can do is externally change the will, which will yield different outcomes.

Moreover, meaning is the will's manifestation, or its destination. It is the end-product of will, like role in RRG model \citep{van1995role}, or the target/objective in optimization (the destination of some process) \citep{lewis2012optimal}. Hence, meaning imply some designer's (accomplished) will, e.g. a chair is designed for sitting (and this is its meaning). See meaning as the end-product of a message in a conversation in Fig~\ref{fig:will_in_constraint_environment}(b).

Additionally, will is an important element in human intelligence, that is almost not mentioned in the modern AI literature.
It is expressed in today's machines, via the requirement from the designer to express his will in a very limited way. For example, in control systems or in software - everything must be designed in a very clear and accurate manner. It is expressed in the form of an objective function in control systems or in AI.

In Natural Language Processing (NLP) will exists in its highest hierarchy, in speech acts, above descriptive-only logical forms. Speech acts contains the will in the message and not only its informative part, which forces the receiver to react accordingly (to a request/question/claim/promise, etc).

Deep Learning (\textit{DL}) tries to override inclusion of explicit will by supplying examples to express will. It is hidden in the need for big data and attention in \textit{DL} models, disguised as an implicit will that determines what should be the correct output. It is also hidden in the curiosity-driven Reinforcement Learning (RL) models, where curiosity is only one specific type of will. 
Instead, will should be addressed directly and explicitly in AI.

One intuitive way to insert will is externally, in a specific input channel only designed for that. This is the explicit way. Complementary or as an alternative, the communication channel can be used, accompanying with its recent context, 
to infer the will from the message that an \textit{AGI} agent receives.
This is the implicit way. It is practiced via prompting in language models (LMs) and it is supported in \textit{AKREM} and in \textit{MOM}.

\subsection{Extending will beyond \textit{AKREM}} \label{sec:AKREM}
Will in \textit{AKREM} is represented in the levels of any specific hierarchy. 
Starting from the most detailed aspects of will, at the lowest level, and finishing at the most abstract will or its essence, at the top. The top level represents some kind of experience uniqueness to differentiate it from other memories that use the same low-level structures.

\begin{figure}[H]%
	\centering
	\subfigure[3-level problem-solving in state space]{%
		\label{fig:a}%
		\includegraphics[width=0.97\textwidth]{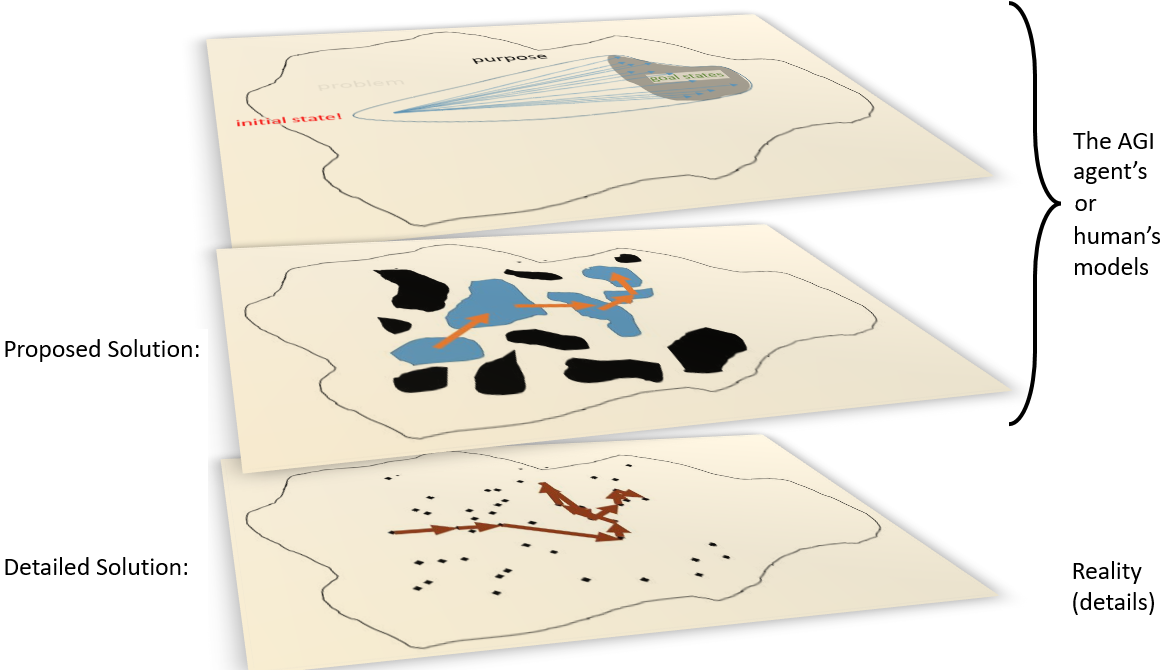}}%
	\hfill
	\subfigure[Communication model]{%
		\label{fig:b}%
		\includegraphics[width=0.94\textwidth]{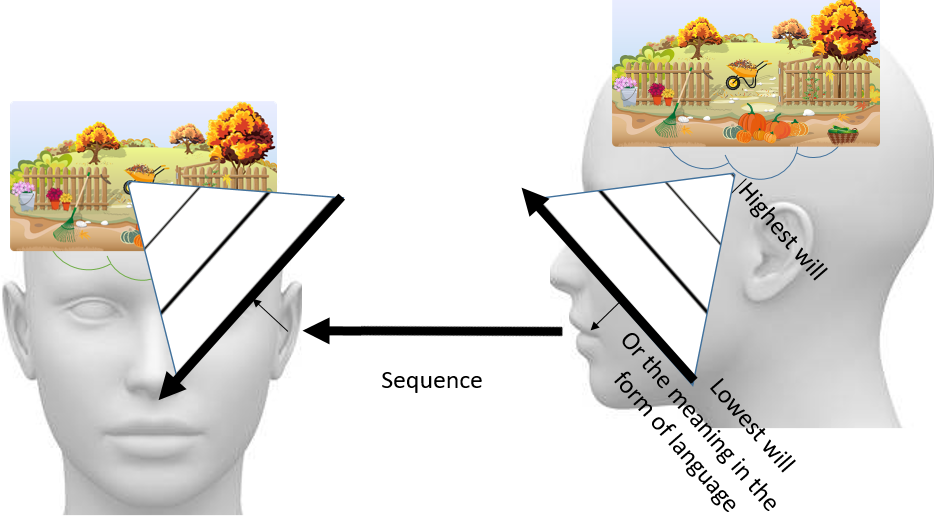}}%
	\caption{Two cases of will in a constraint environment}
	\label{fig:will_in_constraint_environment}
\end{figure}

This hierarchical will is especially demonstrated in a constrained environment, as our reality, for topics like problem-solving and communication. 
In problem-solving, the main will produces sub-wills in lower levels, till it reaches the final solution at the bottom level (Fig~\ref{fig:will_in_constraint_environment}(a)). The final result is a plan or a sequence of actions. See more in Section~\ref{sec:problem_solving}.
Similarly, in communication, the sender encodes/converts his will into a sequence of actions (in a language form), while the recipient on the other side decodes the intention/will from this sequence (Fig~\ref{fig:will_in_constraint_environment}(b)).
In both cases, evaluation is necessary, hence this top-down process is cyclic and non-linear.

All the above are specific cases of will, but there is also a more general will.
As recipients of reality, our main cognitive will is to find the most appropriate/simple model to fit all the pieces/details in the right place or to make the most sense of them\footnote{Making sense is also important for explainability, which can be provided via storytelling modeling. Hence, explainability should be generative and flexible, in its most general conception. In other words, it is a problem-solving task, yielding a sequence of words.}. This is similar to decoding a will from a message/mystery/riddle, and it can be rephrased as a general problem-solving task to comprehend reality \citep{black1980story}. First, it is done internally, by reorganizing our models (mostly during a sleep phase), and later it is done externally, in any kind of problem-solving, or in understanding a message/story/riddle/situation/phenomena.
Moreover, it is claimed in psychology ({\tiny \url{https://www.psychologytoday.com/us/blog/the-resilient-brain/202302/the-many-mental-benefits-of-decluttering}}): \textit{"Not only is it hard to physically function in a cluttered space, but clutter bombards the mind with excessive stimuli. Addressing the heaps of paperwork, laundry, and thoughtfully organizing helps to calm the mind."}. Meaning that the physical cluttering generate our discomfort, since our inner cognitive objective is to reduce entropy, or organize everything in our mind.
The best model will allow us to move from place to place in it easily, perform new actions, and produce conclusions/solutions with less effort. In summary, everything can be considered as a problem-solving task. This is also known as the problem-solving hypothesis, originated in Global Problem-Solver (GPS). For example in communication: in converting will into action and vice versa, in answering a question or in explaining, and in organizing knowledge to make sense of it. More about in Section~\ref{sec:problem_solving}. 

This main cognitive will results with an understanding, or the ability to control any aspect in the complex model. So in a sense, we have two wills governing our cognition: controlling and subsequently making sense. In other words, to accomplish what you want you need to learn how to manipulate your environment. Obviously, these wills are enforcing each other.
It also relates to the learning process. Either it should be performed via learning some model (e.g. by observation), and then learning its control model (by interaction/intervention), or they can be learned simultaneously, i.e. learn a model efficiently so that it would be utilized in the future.


\subsection{Problem-solving and Designing}  \label{sec:problem_solving}

\subsubsection{Problem-solving}

\textbf{Problem-solving via System 1\&2 perspective}

A particular topic that involves will is \textbf{problem-solving} \citep{blocks2010towards}. It is the expression of human will in a constrained reality (otherwise any will could be realized immediately). It is a broad topic, which is about handling any given situation, and not only solving puzzles/mysteries/science. 
Unlike classical AI methods to solve problems, e.g. by searching for similar ones in memory (recording cases), or by finding similar ones and adapt to them in various ways (case based reasoning), here it has much more flexibility.
In a given problem, a response could be either to recognize a previous similar pattern (System 1\footnote{Based on \citep{daniel2017thinking}}), and apply automatic reaction, i.e. immediate resolution, or try to generate a new solution (System 2).

System 2, in our opinion, is representing the cases when wrong top-down prediction requires higher level adjustments, see \citep{hawkins2007intelligence}, i.e., System 2 is the learning mode, tackling new or complex situations, requiring special attention. After enough training and repeating of the same problem in system 2, it becomes automatic and decline down to system 1 for fast response. 

Moreover, System 2 is generative. It means that more generally, problem-solving is generative, while only if possible is discriminative (in System 1). This generation can be expressed for example in explaining. One evidence for generative nature can be seen in \textit{DL}, where explainability is expressed sometimes via special DNN, e.g. in image captioning \citep{vinyals2015show}, where the caption is the explanation of the image. However, explanation may include also summarization or finding the essence of a given group of elements. Hence, another evidence for the generative nature is in dimensionality reduction techniques, as Principal component analysis (PCA) \citep{abdi2010principal}, where high-dimensional input is mapped into a lower one which in turn predicts the correct class, thus the low-dimensional space acts as the explanation space. Similarly, sparse or denoising auto-encoders \citep{bank2020autoencoders} reproduce the input after corrupting it and compressing it. These two examples from \textit{DL}, strengthen our \textit{AKREM} (or \textit{MOM}) hypothesis about grouping. It is since we regard essence of a group to be a lower dimension of a higher one that includes all the details in the group. Note that, explanation or summarization in such group hierarchy could be done in any level of the hierarchy. That is, most concise is in the higher level, while the most detailed is in its lowest level.

System 1 can be viewed as an effortless bottom-up triggering system, coming externally to us, while system 2 requires our effort, i.e. it involves our own will coming from above, so it is a top-down system in this sense.
Note that system 1 is not stimulus$\rightarrow$action, or simple model recognition and a response, but rather there is always a will present and a hierarchy, representing the context of the situation, only that we decide whether to take the fast response (system 1) or the slow one (system 2).

\begin{figure}[H]
	\centering
	\includegraphics[width=0.7\textwidth]{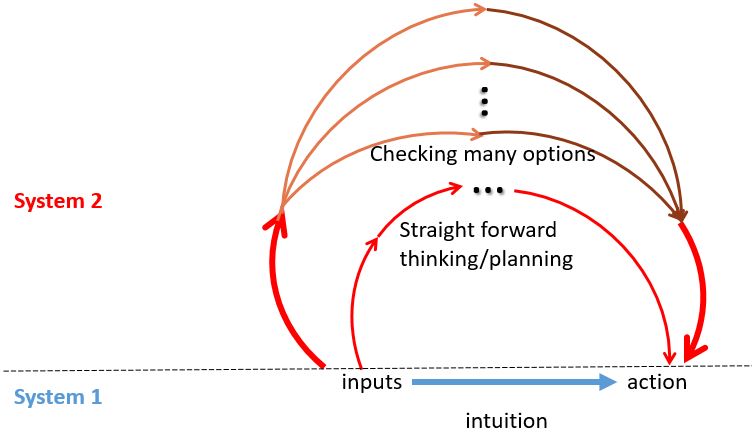}
	\caption{Short to Long paths between inputs and action}
	\label{fig:system1_and_2}
\end{figure}

In summary, as illustrated in Fig.~\ref{fig:system1_and_2}, you can either decide upon a set of actions, without any cognitive processing, a kind of straight-way response (System 1), or you can have possible cognitive operations, before deciding upon the most proper response (System 2). 
System 2 can be regarded as problem solving, i.e. via first divergent thinking, finding possible solutions/approaches, then applying convergent thinking, to decide upon most appropriate solution.\footnote{Similarly to competing \textit{AGI} modules over WM attention, in Global Workspace Theory (GWT).}

A side note, is that we can view System 1 as automatic first response, that occurs always, while System 2 can be activated afterwards if necessary. Methods like Relaxed-Planning-Graph (RPG) from classical AI can be used to find an unconstrained fast solution of shortest path between current and goal state, to act as illuminating the different paths from the initial state to a goal one.

Divergent thinking could be counter-factual one, i.e. “what if” planning, like in chess, where you contemplate about what will happen if you decide to do some action.\footnote{In this case we are aware of it. In other cases it could be sub-conscious.} Which is why in our modeling the knowledge is dynamic, and can always produce new outcomes by the set of admissible actions each model has.
This thinking could obviously form a tree-like branching, since we could perform few steps of "what if", in series. Also it can be hierarchical, as any memory, evole not also in breadth (tree-like), but also in height.

As can be seen in Fig.~\ref{fig:system1_and_2}, intuition is the shortest path for response.

\textrm{\\}
\textbf{Problem-solving in \textit{MOM}}

\begin{figure}[!htb]%
	\centering
	\subfigure[1: the problem]{%
		\label{fig:a}%
		\includegraphics[width=0.44\textwidth]{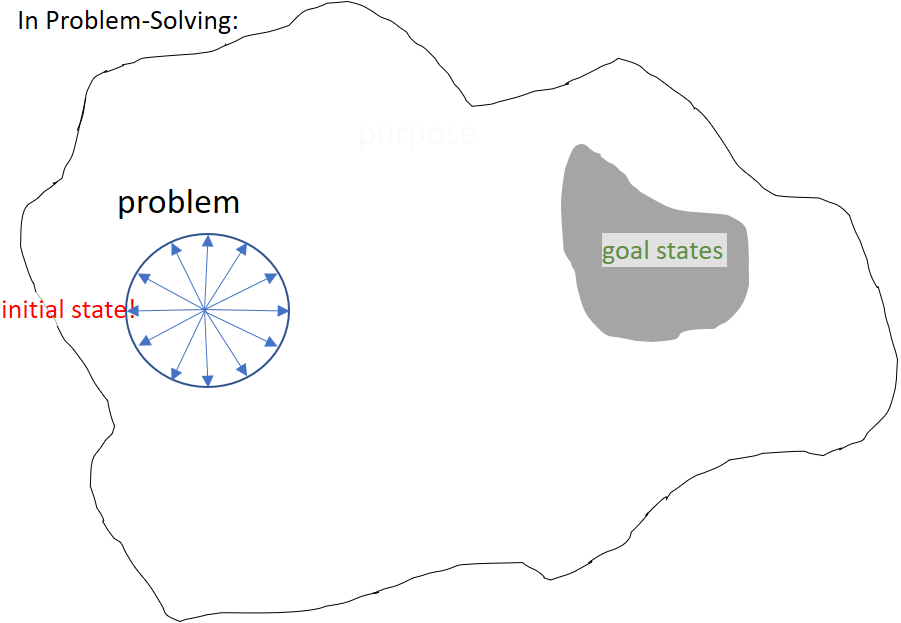}}%
	\hfill
	\subfigure[2: will turns into a purpose]{%
		\label{fig:b}%
		\includegraphics[width=0.44\textwidth]{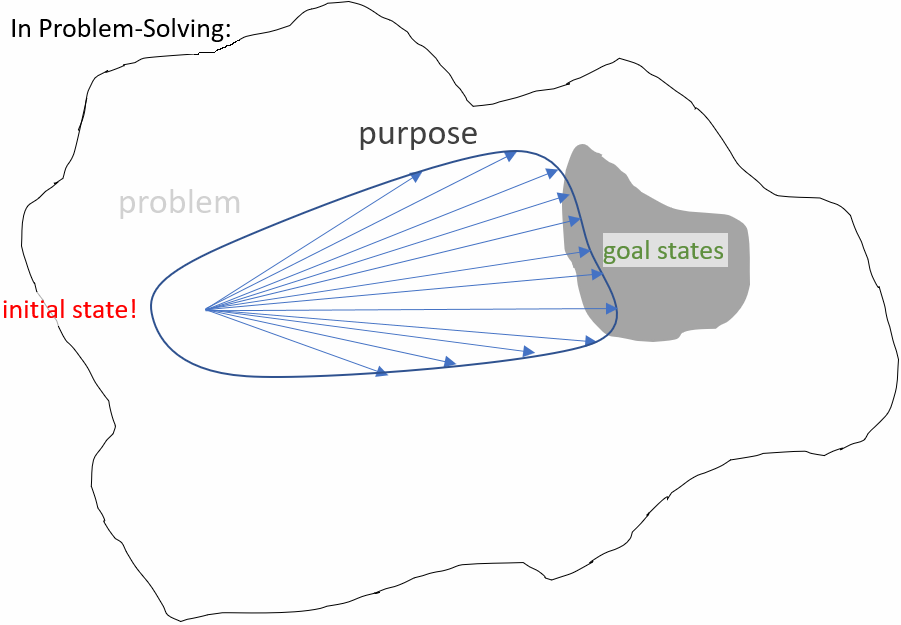}}%
	\caption{Gradual will refinement in problem-solving}
	\label{fig:Phases_in_problem_solving}
\end{figure}

In this section, the System 2 response is discussed, since System 1 response is a simple pattern-matching.
This process is modeled via 2-phase process in state space:
\begin{itemize}
	\item Refinement: starting from a general will to get out of a problematic state, then deciding upon some goal states to be reached. It makes the will more definite (directional/purposeful), see Fig~\ref{fig:Phases_in_problem_solving}.
	\item Realization: searching for a solution via coarse-to-fine hierarchy of models (Fig~\ref{fig:will_in_constraint_environment}(a)).
\end{itemize}

In refinement, a vague will turns to be more precise and directional.
Since will derives action(s), it is represented similarly to an action in the state space - as a vector, transmitting one situation to another. Meaning, will defines the direction the agent wants to move, before it found the admissible/legitimate/allowable way to realize it, in the given environment.
Finally, the agent starts to plan how to solve the given problem, under given constraints, i.e. where one cannot fulfill its will directly, but instead is looking for some legitimate way to accomplish this, in the given circumstances.

In summary, Fig~\ref{fig:Phases_in_problem_solving} describing the gradual change, where at first the purpose is vague, hence a large space can represent the goal state. Then, as the will is refined, it means more constraints are added or more descriptions of the goal. In other words, the goal is shrink from something general and vague, to something more specific.
In perception it is the opposite. We perform inverse engineering. The basic understanding is to know what happened/was done. Better understanding is to know what was the purpose (behind the actions). But the full understanding is also from what problem this purpose was yielded/created from. 

Note, that will, after refinement, is merely a sketch of initial and goal states. To realize it, the will is used to scan for models of different levels. It is like an attentive beacon(s), focusing on potential models, see more in \ref{sec:attention}.

The Fig~\ref{fig:will_in_constraint_environment}(a) figure demonstrates a coarse-to-fine hierarchy, where the will along with its refinement, is placed at a top level. This level is vague, since nothing is perceived clearly about the ground level. However, descending the levels reveal more and more details, and get the will more closely to realization. It is similar to the process of zooming in on a geographical map.
Higher levels propose general models as a potential stations in a possible trajectory from a problem state to a goal state. Then at descending, finer models (with more details) are proposed, consistent with the upper levels, to move from a problem to a goal state.

This idea is similar to the idea used commonly in problem-solving \citep{zelikman2023parsel}: breaking a problem to smaller sub-problems, and then solving each separately, and finally combining them into one whole solution.

The search at any level can be performed by any heuristic/learned model, such as back/forward chaining, Depth-First Search/Breadth-First Search techniques \citep{al2015comparison}, or any combination of those. However, more generally, it is dictated by will. 

Note the mismatch between the lowest model level (proposed solution) and data level (detailed solution), in Fig~\ref{fig:will_in_constraint_environment}(a). It is due to our inclination towards abstracting, i.e. memorizing the essences and less the details \citep{ref for memory changing will retreiving it}, which is essential for efficient learning.\footnote{Philosophically, it extends into separating models from reality totally. For example, the assumption of real figures behind the shadows in Plato's cave, or the idealism where presumed ideal objects out of this world, where only coarse instances of those are experienced in our world. Or the space-time curvatures of Einstein's general relativity, a model that cannot be detected directly.} It is described also in \ref{sec:consolidation}, where it is better to learn several patterns than to loose yourself in a non-pattern realm, where all we see are details.

Also note that there are other configurations of problem-solving, such as inference: given specific start state and specific end state, to infer what happened in between. For example, at 12:00pm child left alone in a room with a closed cookie jar containing a cookie, and at 12:05pm it is observed that the jar is closed, but the cookie is nowhere to be seen. This inference also happens in every information perception (as in a story), where we required to fill in missing details. Also, in retrieval from memory, many details fade with time, e.g. time indexes, hence inference is used to extract the time something happened (e.g. if it happened in elementary school, with Sarah as teacher, the I must have been in 4th grade). 
But most likely, inference occur not only in interpolating (inner missing information), but also in extrapolating, i.e. forecasting how the plot develops, or proposing several plausible developments, while reducing the wrong ones as the plot is revealed. Just like investigators.

All these examples are also supported in neuroscience, e.g. \citep{hahamy2023human}, where just like as in \textit{AKREM}, \textit{MOM} is also inferring by traveling the different hierarchies/memories, within them and between different ones.

Other configuration could be also in cases where the problem requires a long-term process, such as succeeding in educational course. These long-term goals, can be partly active in the background, either to be solved (or partially solved) during a mundane tasks, or through acquisition of new data and extracting the relevant information from it to the background problem. 

In summary, this approach is non-local, i.e. similar to Means-Ends Analysis \citep{sweller1982effects}, it is looking simultaneously at the whole region, only within different resolutions.
It is also cyclic and non-linear, both in the will-refining stage and in the realization stage. At will refining, it is since sometimes the goal states cannot be reached, so other states are needed to be generated, sometimes as a compromise. At the realization stage, it is since descending in levels might result in conflicts or failures, due to misalignment between the lowest models of reality and the actual reality. Hence, returning to higher levels for trying different solutions is needed.


\subsubsection{Designing}

\begin{figure}[!htb]%
	\centering
	\subfigure[1: Design task]{%
		\label{fig:a}%
		\includegraphics[width=0.44\textwidth]{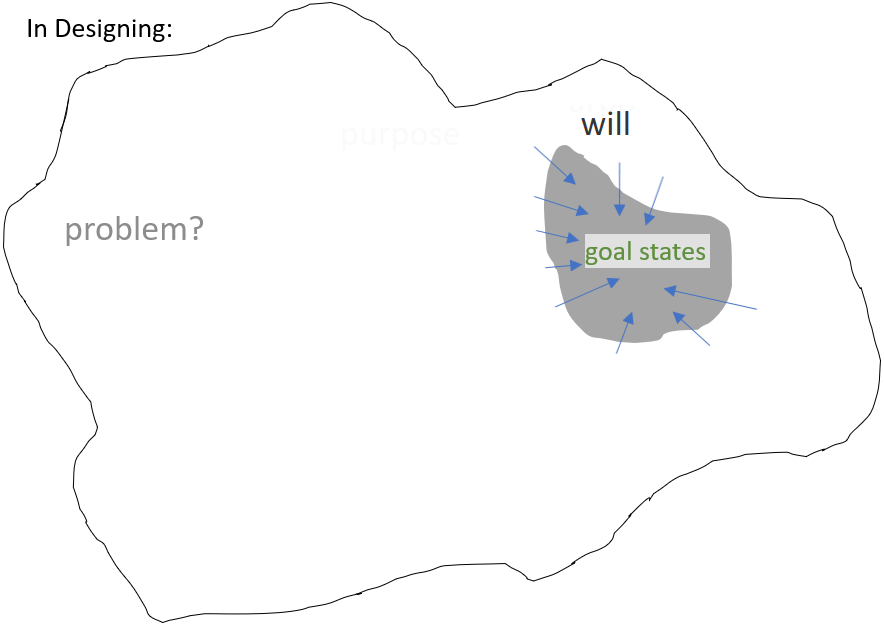}}%
	\hfill
	\subfigure[2: Problem-solving task]{%
		\label{fig:b}%
		\includegraphics[width=0.44\textwidth]{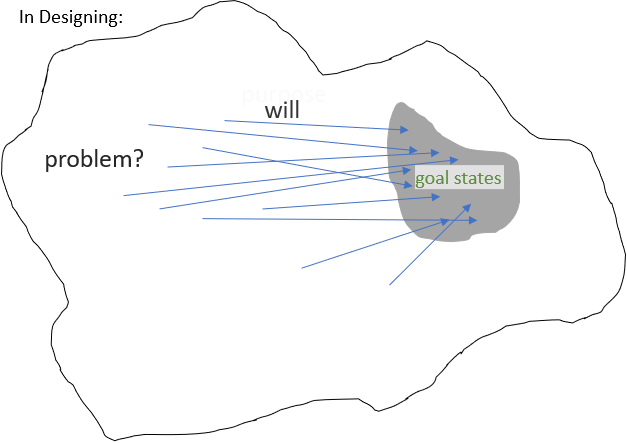}}%
	\caption{Designing approach turns into problem-solving task}
	\label{fig:designing}
\end{figure}

While in problem-solving, the will was generated from a problem, i.e. growing from an initial state, in \textbf{designing} it is the opposite. Here, instead, it is growing from the final state(s), searching for the best state to start the full solution from. It is like creating a story backward: starting from the end, to reach some beginning (Fig~\ref{fig:designing}).
In this case the final states are given, while unlike problem-solving the initial states are extended into a region (of possibilities). 

In designing there is a goal and a will to go there, but no specification of the problem or the initial states.
Hence, it is an iterative process, see general algorithm~\ref{designing_algorithm}. Each iteration starts from searching for a problem to reach the goal, then continues with a specific will connecting the problem with the goal, resulting with a problem to solve, thus formalizing it as a usual problem-solving task (Fig~\ref{fig:designing}).

\begin{algorithm}[H]
	\caption{Designing Algorithm}
	\begin{algorithmic}
		\While{not satisfied}
		\State search a problem to reach the goal
		\State try to resolve problem-solving task
		\EndWhile
	\end{algorithmic}
	\label{designing_algorithm}
\end{algorithm}

\subsubsection{Summary of Problem-Solving and Designing} \label{sec:summary}


One can spot a duality here also.
On the one hand, problem-solving is an analytical perspective, looking for a resolution or finishing a problem, by usually a systemic view, breaking it to parts and then looking for some appropriate solution, that serves as a better state than the one we started with (problem state).
On the other hand, designing is based on a holistic perspective, where instead of finding a fast/analytical resolution to a problem (“to make everyone happy and go on with our lives”) it is about empathy/consideration, i.e. it is about looking for the roots of the problem, and not just shutting it down quickly. It takes the opposite approach: instead of reducing the problem, it tries to track its sources and thus solving the causes that generated this problem state/situation in the first place. By doing so, it searches for a better problem to solve, a root problem, which solves multiple other problems too.

Nevertheless, both problem-solving and design are non-linear in the sense that at each stage, we can skip backward to previous stages or cycle through the same stage several times. Also they are both have evaluation mode where the agent stops what it does to evaluate the whole process efficiency. For example, by checking if the current state can be considered as the final state, i.e. whether it holds with all the requirements for the solution. Or by checking how far the current states of the different trajectories to the solution. This non-linear cycling is like moving up and down in the hierarchy of wills.

Additionally, we can see designing as specific and private case of problem-solving, only now the “given” is different.

In summary, will 
is either starts from getting somewhere (goal) or getting out of something (problem), similarly to our basic wills to feel good and to get away from what is considered bad.
In terms of explaining, "problem" is explaining the will's outward directions, while "purpose" is explaining the will's inward directions (or its realization process).

Side note: searching for a trajectory to solve a problem can be also performed by inversion. For example, proving by negation, or explaining something not by usual causality but via contrastive manner, i.e. how other things are invalid in order for deriving the desirable outcome \citep{gilo2022anytime}.

\subsection{Will Field} \label{sec:WillField}

As been mentioned previously, if actions are vectors, then will is also a vector. This fact may imply that the direction property of the action is will, i.e. will is aligned with actions, to solve problems efficiently and fast. Thus it creates a field in the state space, which helps us to solve the problems we encounter. Hence, we have separate attributes of an action: consolidation measure and directionality of an action. It is like electrical/magnetic field, where the problem state can be considered as "+" charge for example, see Fig~\ref{fig:Phases_in_problem_solving}(a), and a goal state can be considered as "-" charge, see Fig~\ref{fig:designing}(a). Consequently, when both appear, it defines problem-solving task, as seen in Fig~\ref{fig:Phases_in_problem_solving}(b). Of course, more complex structures are allowed, i.e. with multiple "particles".

\begin{figure}[H]%
	\centering
	\subfigure[System 1 solution]{%
		\label{fig:a}%
		\includegraphics[width=0.49\textwidth]{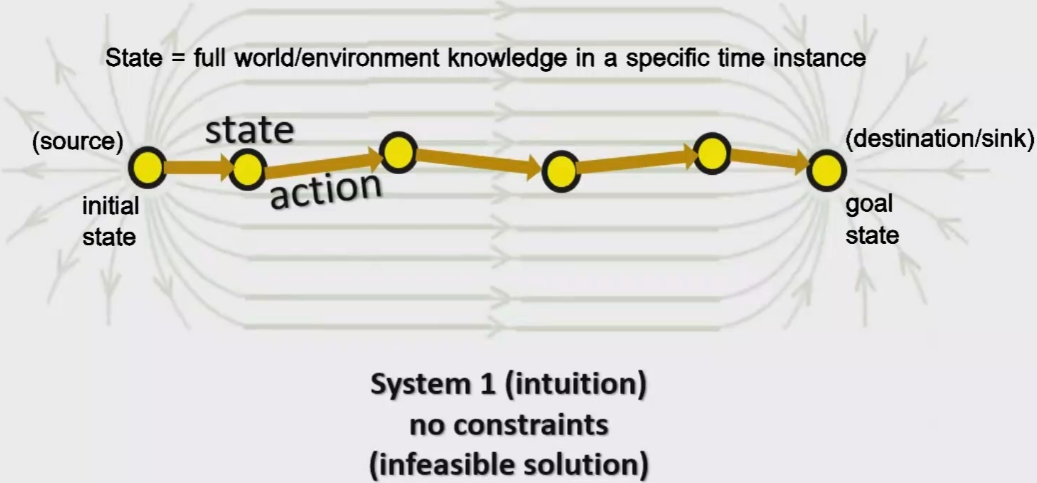}}%
	\hfill
	\subfigure[System 2 solutions]{%
		\label{fig:b}%
		\includegraphics[width=0.49\textwidth]{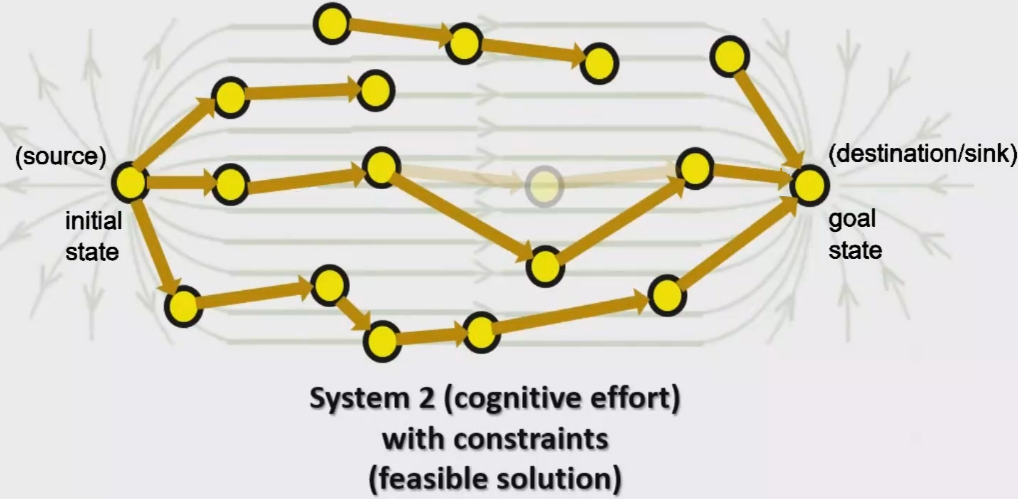}}%
	\caption{Planning – From Will to Knowledge models.}
	\label{fig:system1&2_solution}
\end{figure}

We can see demonstration of will field in searching for solutions in a given problem-solving task, as described in \ref{sec:problem_solving}. See Fig~\ref{fig:system1&2_solution}(a) for System 1 solution. It is the first most intuitive solution that pops up, and it is almost the shortest path from problem to goal. However it is not considering constraints of the problem and of logic.

Hence, for the real (or a better) solution we try different paths, either close to the intuitive solution or far from it, see Fig~\ref{fig:system1&2_solution}(b) for System 2 possible solutions.
 
Note, that in this figure we see the alignment of will and knowledge models, in the same state space (see the will field in the background). This alignment support organization of knowledge (models), which is necessary for effective problem solving, instead of blind search (such as A*, BFS, DFS, Means-end analysis and more). So the AI works hard in organizing/aligning the actions, in order to work with minimum effort in the future, for fast response.

The directionality can be in high dimension, since will can be a complex multi-facet vector with different aspects. Similarly, it could be like Capsule Nets \citep{sabour2017dynamic}. Still you should note, that we represent will as an arrow/vector, thus assuming its basic form: either moving away from something or moving towards something.

This idea can be attached to different human feelings, such as disgust, horror, nostalgia, relief, surprise, and more. 
See more in Fig~\ref{fig:will_types}.
\begin{figure}[H]
	\centering
	\includegraphics[width=0.99\textwidth]{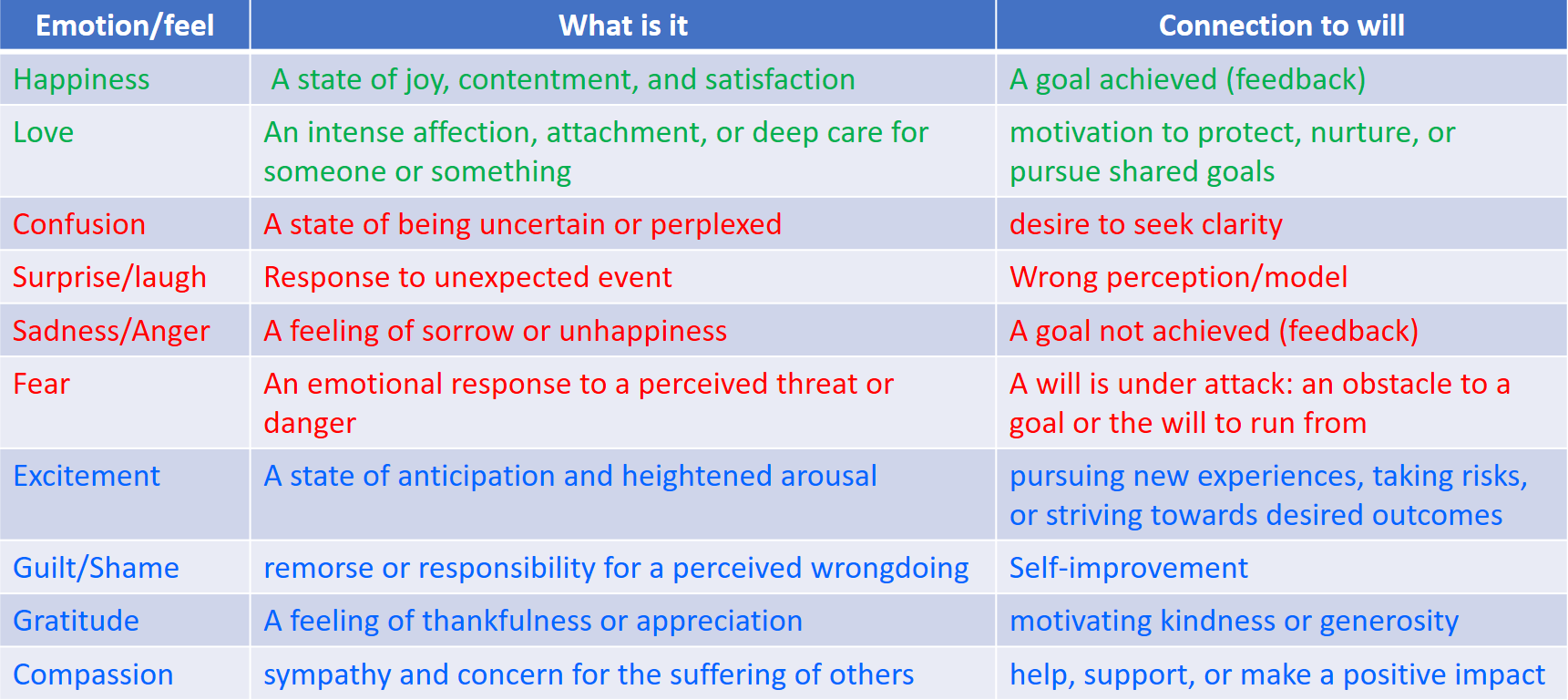}
	\caption{Human's emotions and their relation to will}
	\label{fig:feelings}
\end{figure}
where in green: "+" will, i.e. will to go to (goal); in red: "-" will, i.e. will to go from (problem); in blue: unrelated emotions to AGI, like meaning of life, etc. Also the sense of self model, within AGI and other agents.

This idea may explain why most of our problems are already aligned with will from previous experiences, mainly being instructed or taught by other people.
Only very limited amount of problems are actually being solved from scratch, without alignment with will, which makes it feel like searching in the dark. 
It is an exhaustive search, where we are the first pioneers to create this alignment for others. As if we chart/pave a new path in a geographic map for others to use.

This idea may be contemplated philosophically or religiously, claiming that the real human is a non-physical spirit/soul, attached to a physical body. This spirit has will. Thus in order to manifest this will, the spirit must somehow align it to the models it learns from the physical reality\footnote{There is the materialist theory, claiming that consciousness is settled in the brain. It is supported by neuroscience studies that show correlation between mental experiences with specific neural activities. However, it can also support dualism, since the alignment is definitely yield such correlations.}.

Another issue, is what constitutes a state in state-space. Probably it is a collection of classes (and their states), but then how a collection of objects acquire a direction, and how is it implemented to its items?
One option to solve this question: actions are already operate on several objects, so we could say that they have direction, to change the state, even if partially, towards the goal. Hence, it is ok to define single action’s direction. Alternatively, we can set a direction to sequential action, i.e. action that is a recipe-like (composed of sequence of actions).

Regarding problem-solving, after a will is settled, the first reaction is of System 1 type. This intuitive solution is usually very close to the real solution. It is often only needs a small tweaks.
But if it is not enough, System 2 starts to search for partial paths (Divergent thinking) which finally combined into one solution (Convergent thinking).

This alignment is what makes organization of knowledge (models), which is necessary for effective problem solving, instead of blind search. Hence we work hard in organizing/aligning the actions, off-line, in order to work with minimum effort in real-time (on-line), for fast response.

More basically, the idea of will being aligned with actions is evident in new born infants. Their basic learning is about relating the will to body control, e.g. when one wants to raise its arm.

Finally, as seen in Section~\ref{sec:problem_solving}, there are two separate spaces: models state space and will space. However, just as complex processes occur in models state space, there could be complex processes in will space. For example, the current will may derive from primitive emotions or from higher values, such as justice, respect, empathy to others (consideration, cooperation), success, equality, and more. 

Nevertheless, our hypothesis that the knowledge map itself is dynamic and dependent on will imply also that anything can be justified. That is, the basic problem-solving platform most often can find a trajectory to accomplish this will.
Hence, the will dictates our perception and conclusions, while the rationality is merely a tool to realize and justify will.

\subsection{Basic Wills}  \label{sec:basic_wills}

Although there are wills to solve problems, as discussed in previous sub-sections (will field and problem solving), there are fundamental wills that guide throughout our whole live in acquiring knowledge. It starts from the moment we born and continuous all the time. Our curiosity and wonder drive us and thus generate basic questions, e.g. Wh-questions (where, who, by whom, what, how, when, what, why, etc) or any other type of questions. These questions represent our basic wills to understand what we perceive. For example, in reading comprehension of young students.
This idea resembles to the RRG method for language parsing \citep{van1995role}.

It means, that we induce these basic wills during perceiving a story/message, even when no goal states exist. However, they are also guide us when a problem is conceived. In such case they can act as 
micro-wills, serving the macro will in the scope of the whole problem.
These basic wills serve as guiding wills in inference, especially for completing missing information and more generally for the most important reason of all: understanding. It is like anything we perceive is like a mystery or a riddle, and our mind always tries to make sense of everything in the perceived message. Part of this process is also in cases where inference is confident and tries to predict what comes next, in order to evaluate how the current models are really confident (which if not - require some resolution or model update).
With such interpretation we can picture these basic wills as small scanning/scouring devices/bots or probes, that are being sent first to examine the surface (see best analogy in \textit{"Minority Report"} 2002 movie with Tom Cruise, sending spider robots to scan people's eye for identification). In our case, scanning the surface is for all relevant facts and knowledge that can be gathered to solve the given mystery. Then these probes fill in missing information and inconsistencies.

Perhaps these basic wills are the most primeval wills, rooted deep into what defines us human. Hence, these are usually aligned in early development, and frequently aligned also throughout our lives. Moreover, since they can be thought of as preceding the knowledge acquisition, they can be predefined in \textit{MOM}, as prior knowledge of the system.

\begin{figure}[H]
	\centering
	\includegraphics[width=0.95\textwidth]{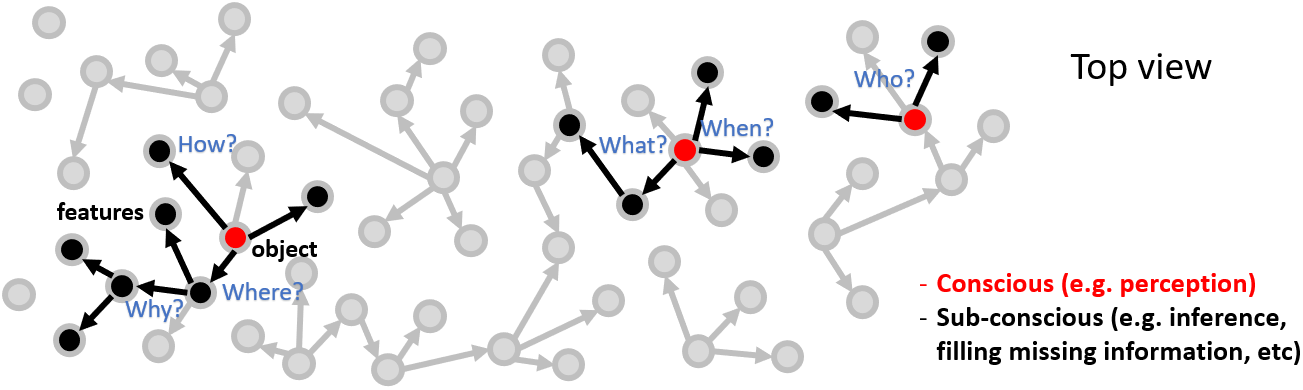}
	\caption{Inference while perceiving}
	\label{fig:inference}
\end{figure}

See how inference is executed while perceiving data in Fig~\ref{fig:inference}.

Given a knowledge map (gray network in the figure), we assume that at some instance we perceive an input, which triggers some object in this map (the circles in red).
Then an inference process starts to characterize this object, by propagating and activating relevant features\footnote{We are a bit ahead of ourselves: this is Forms or \textit{OOP} representation of knowledge, with objects such as concepts, and their attributes/features that characterize them. More about this formulation is in \ref{sec:Modeling}.}, either via 1 step or more.
But the inference scanning is not blind. As mentioned above, it is usually also guided by will, the basic will, in a form of a simple question.
This inference occurs not only in understanding, but also in planning, specifically to comprehend the problem and the goal.

As we will see later, these basic question-like wills are guiding us not only in the micro of inference, but also in stitching larger components in a given story, to make sense also in the macro. Meaning, not only its function is in filling missing common data, but also in filling missing parts of a story, finally resulting with a fully-explained story.

This explains why children are so often asks a lot of questions: for the purpose to align the micro and macro inference processes with the will. While as adults we continue to ask questions, to solve bigger problems.
Also, in the opposite, many psychological methods are help people to get out of problematic or stuck situation, by guiding questions they ask themselves. This may be, how previous alignment, may illuminate the knowledge hidden and darken that already exists with us. 

This also explains the motivation to design curiosity loss in AI agents, such as embedding it in their reward via RL. Or presenting an extra reward for asking questions in a conversational settings, such as in dialogue or in some game.

Finally, there are many guides for optimal prompts for LLMs, it is mostly tuned to response in a service orientation, i.e. to accomplish what is was asked for. However, usually it is very hard to understand exactly what was asked. Hence prompt engineering is such a difficult skill to acquire. Instead, what humans do in these situations is simple: they do not assume that everything clear right from the first interaction, and ask questions for clarification. Similarly an AI agent should ask questions to align itself with the user will.

\subsection{Types of Will}  \label{sec:will_types}

\begin{figure}[H]
	\centering
	\includegraphics[width=0.95\textwidth]{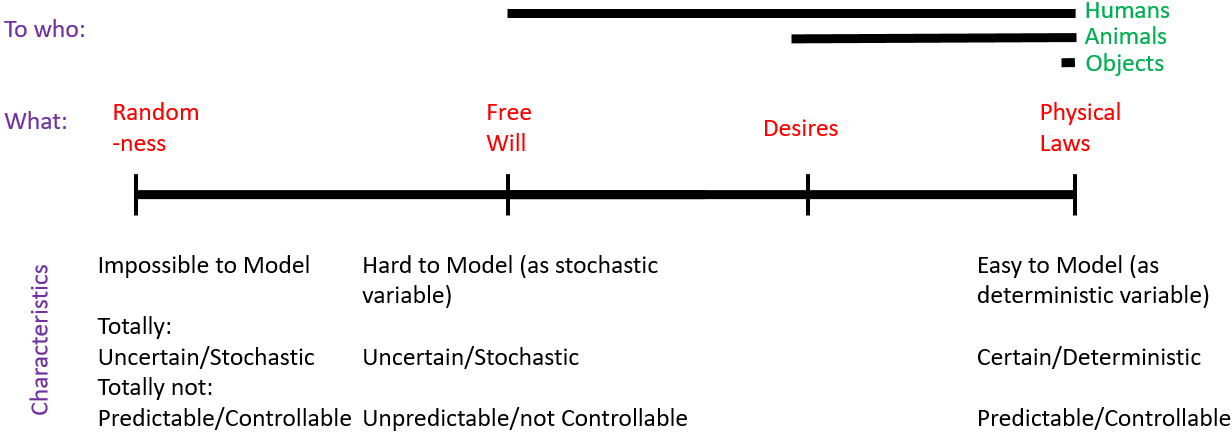}
	\caption{Types of will}
	\label{fig:will_types}
\end{figure}

Additionally, there are different categories of will, such as chronology, causation, and purposefulness. In stories, they are very intertwined/mixed. It is because purposefulness is a higher manifestation of will (usually applied in humans), while causation is a lower one, usually applied to animals/objects (e.g. \textit{"A causes B"}), and chronology is simply the way will is implemented: in a delay. You first want, and then you try to accomplish it. Or in the case of causation, there is a law, as a fixed kind of will (e.g. gravity), and then it is realized. 
See more in Fig~\ref{fig:will_types}.

These types of will are different qualitatively: objects do not will at all. It is an enforced will, that lacks any freedom to resist it. Animals have instincts which at least externally present some kind of choice, or at least some degrees of freedom, even if it is highly predictable and physically-constrained. Humans on the other hand, have much higher freedom (free-will), to be aware of their will in the first place, and then to govern it.


In other words, we could say that \emph{Desire} is "a strong wish", which is usually not controlled by you, but instead controls you, hence it is more closer to physical level, since it is more predictable, and less vague/random. While will is assumed to be free, i.e. uncontrolled-by/independent-from the state of environment. Therefore, will can be an internal variable in a model (similar to Einstein's famous rejection to Quantum Mechanics, assuming that there are hidden variables). It is a hidden independent variable, due to the assumption that it is not affected or dependent on other variables, such as input variables perceived by the senses.
But as discussed above, its independence has measure (from being highly independent in free-will, to being highly dependent with physical phenomena).

Finally, will is everywhere, in all of our daily interactions. We try to figure out animals' intentions or people's hidden will, to make sense, or in our case to make our model complete, i.e. enable prediction.\footnote{Model completeness drive us to search also meaning of life and existence, found usually in religion or philosophy.} 
Since a specific human has his own will, he tries to figure out other people's will to enforce its own will over theirs. Sometimes it is via competition (if wills are in conflict), while sometimes it is through cooperation (common will, when it is more beneficial to work together then solely or in competition).
Hence the rise of lying/deceiving phenomena (either learned from observation or inherited), which introduces new models to learn. It is beyond good or bad. It is simply modeling for the purpose of control, or the manifestation of my will, whatever it is.
Moreover, will might be a complex phenomena. That is, it is dynamic like the wind (as spirit in spirituality). For example, it often can be a mix of wills, even wills in conflict \citep{heckhausen2008motivation}. Nevertheless, in this paper will is considered as a single entity at each moment, to simplify the discussion\footnote{Although it can be modeled simply by superposition, i.e. by summing different wills as any mathematical fields.}.

Also, will can be fundamental, i.e. in the background of all discussed instances of will above. For example, for animals and humans: survival which includes protection and gaining energy mainly by food (again the going away from and to types). In humans it can be expressed via livelihood or a comfort and pleasant life, etc.
These wills can be long-range, and without explicit actions derived from them, though these wills can be regarded as suspended, i.e. not always relevant. For example, while solving some problem, their are ignored.

However, how is will actually included in \textit{MOM}? It is discussed in \ref{sec:Modeling} and Section~\ref{sec:attention}.

\begin{figure}[H]
	\centering
	\includegraphics[width=0.75\textwidth]{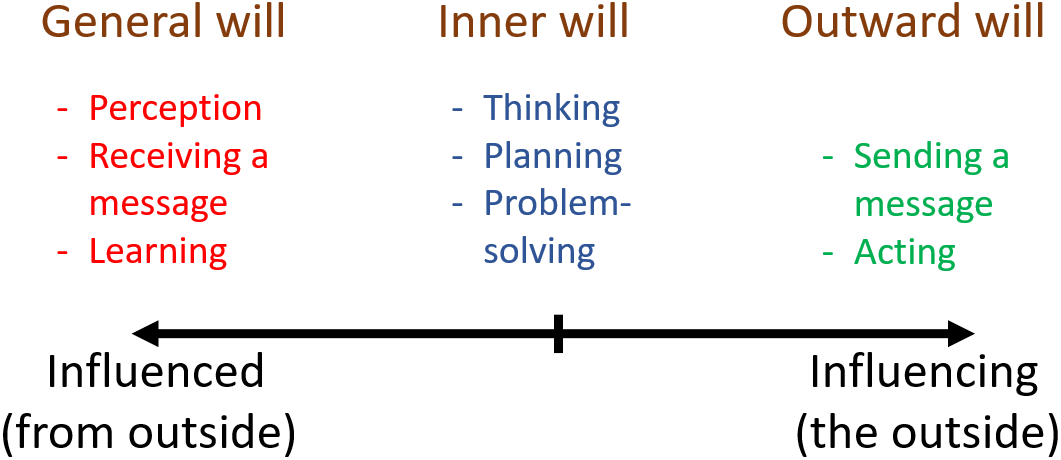}
	\caption{Cognitive operations under a category of the main will.}
	\label{fig:main_will}
\end{figure}

Another possible categorization of will is represented in Fig.~\ref{fig:cognition}, see the summary sub-section below. But additionally, we can have other categorization of will. In this case, we assume that will or some kind of objective is always presented in human/agent, in all cases. Hence, we are not passive while hearing/learning/perceiving. Similarly, there is no encoding/decoding will, as illustrated in  Fig~\ref{fig:will_in_constraint_environment}(b), since our main will is always exists, and the deciphered will is still under this main will. See this kind of categorization in Fig~\ref{fig:main_will}.

Example for outward will is explaining. It is like going over the trajectory of an hierarchy (see \textit{AKREM}). While in acting, it is like performing physically or executing a given hierarchy.

\subsection{Will chapter summary}  \label{sec:op_and_modeling_summary}
\begin{figure}[H]%
	\centering
	\subfigure[Cognition parts]{%
		\label{fig:a}%
		\includegraphics[width=0.47\textwidth]{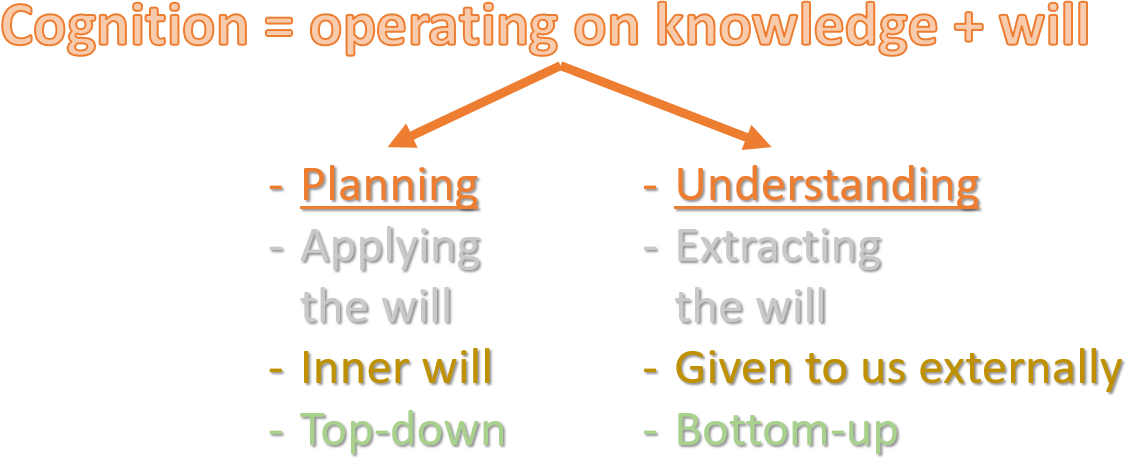}}%
	\hfill
	\subfigure[planning verse understanding]{%
		\label{fig:b}%
		\includegraphics[width=0.47\textwidth]{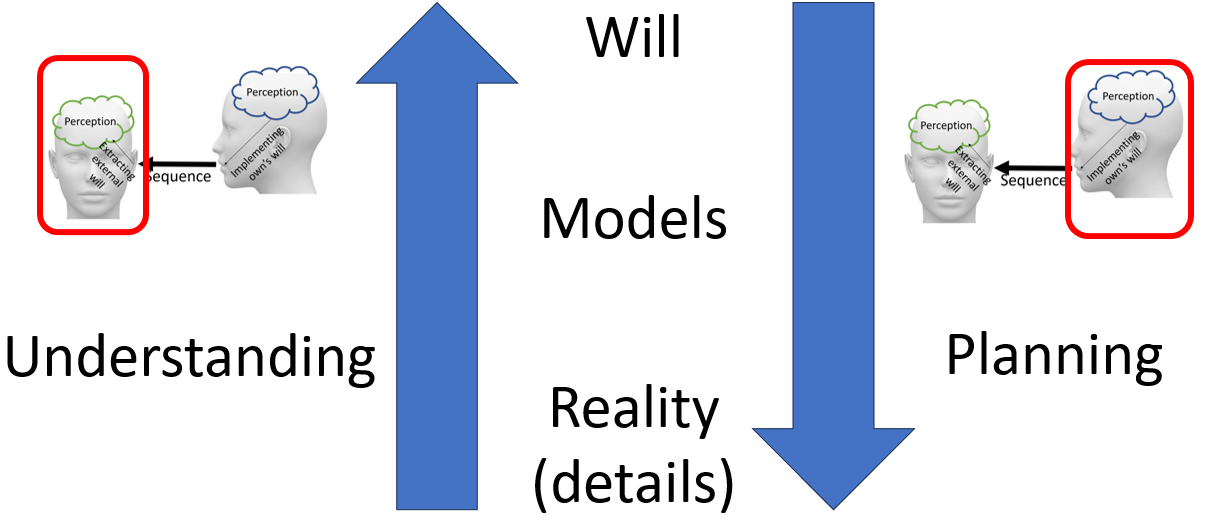}}%
	\caption{Cognition definition in \textit{MOM}.}
	\label{fig:cognition}
\end{figure}

In conclusion of this chapter, we can define cognition a shown in Fig.~\ref{fig:cognition}. 

Following are some additional definitions.
\begin{itemize}
	\item Will = what affects the actions in state space.
	\item State = full world information at a given time instance.
	\item Action = transform a given state to a new state.
\end{itemize}
All these elements are coexist, aligned, in the same state space.
These definitions are typical definition in many science fields, such in classic AI, RL, and control theory.
Eventually, will, like action, are both vectors in state space, only we assume that usually will is affecting the preference of which action to be selected to change the current state.
Note that the states are discrete while actions can be either continuous or discrete also, and the will is acting like a filter to reduce the amount or the range of possible actions we should apply to change the current state. This allows a more efficient and faster problem solving, since the solution space is reduced significantly.
Also note that the state represent only all relevant elements of knowledge, and not everything that is in the world, since we have a limited WM.
Later, in \ref{sec:Summary_op_modeling}, we will define the full model, which defines also the states and actions themselves, which are composed of basic objects/actions.

\section{New Associations}  \label{sec:Associations}

Inspired by semantic nets, some general connections are proposed, such as inheritance (is-a relation), instance property (is-instance-of relation), part-whole property (part-of relation), an attribute of an object property (has relation), assigning a value to attribute (value relation)\footnote{Actually the is-a, instance, and assigning relations can be grouped to one type, since these operations are all represent specification of some general concept.}, synonyms, antonyms/opposites and more, see \citep{Wiktionary}. 
All these connections are static and can represent stories as separate items, as it is in \textit{AKREM}. But to connect these details, as a sequence of applied actions, operationability is introduced.

\section{Operational modeling}  \label{sec:OperationalModeling}

The two principles added to \textit{AKREM} are operationability and modeling. Both are formalized via frames or classes in \textit{OOP} (Object-Oriented Programming) \citep{rentsch1982object}. Operationability represents the methods that can operate on classes and change their state, while modeling in our terms is the abstraction capability\footnote{
	In philosophy, abstraction is like idealization \citep{dunham2014idealism}, i.e. like "treeness" is the perfect idea of a tree, while all trees we actually see, are some instances of it. Similarly is the idea of perfect geometric shapes, like circle or cube.}.

Both are utilized to construct a general model that can implement different cognitive functions humans have, and the different models representing interactions, see \textit{"Models in communication"} section in \citep{EasyChair:7921} and Fig.~\ref{fig:Modelling_machines}. 
To reach this utilization, this model representation should be transformed into a learnable model such as DNN.

\subsection{Operationability}  \label{sec:Operationability}

It is hypothesized that thinking is operational. Meaning, it is a process, generating new facts along the way, via some set of actions.
Hence, the first principle added to \textit{AKREM} is operationability, which turns it from static to dynamic knowledge representation. That is, unlike static connections as in \textit{AKREM} and knowledge/scene-graphs (representing facts or a scene shot) - action connections in \textit{MOM} can also be productive (produce new \textit{elements}).
I.e., other knowledge representations, including \textit{AKREM}, have pre-defined (or independently learned) actions and objects, constructing their knowledge base. Similarly is in rule-engine of an expert system, where there are rules and data they operate on.

Operationability also allows for maximum flexibility in reasoning, problem solving, or in finding the proper response in a given situation. It does it by associating each concept with maximum related operations and attributes. Thus creating highly connected network of possible actions/paths for huge amount of scenarios to deal with.

\begin{figure}[H]%
	\centering
	\subfigure[Previous \textit{AKREM} model]{%
		\label{fig:a}%
		\includegraphics[width=0.47\textwidth]{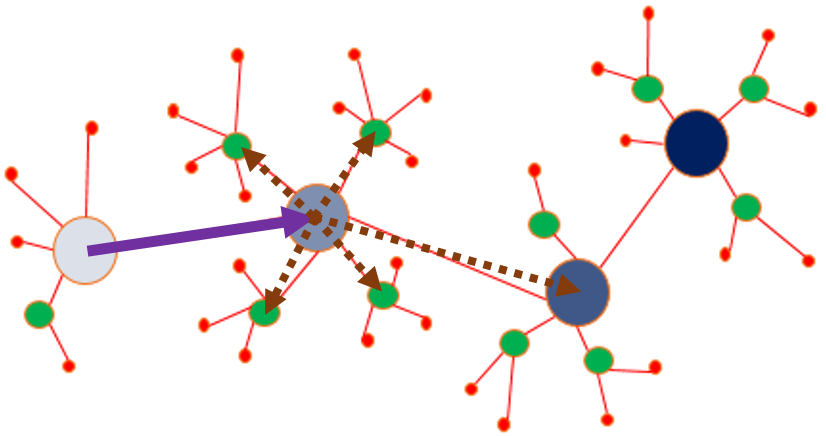}}%
	\hfill
	\subfigure[Current \textit{MOM} model]{%
		\label{fig:b}%
		\includegraphics[width=0.47\textwidth]{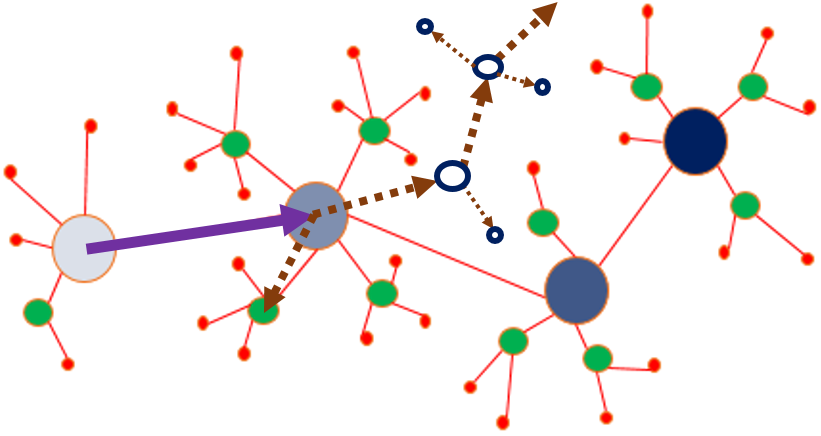}}%
	\caption{Current model allow more degrees of freedom}
	\label{fig:prev_verse_curr}
\end{figure}

Consequently, operationability adds degrees of freedom to the current cognitive model, to move in new directions along the hierarchy, i.e. to create new hierarchies on the fly or update old ones, via admissible actions. It can be seen in Fig.~\ref{fig:prev_verse_curr}(a), where the previous model, \textit{AKREM}, had the ability to move on static memories/hierarchies (instances of associations). But the current model, is not limited to the options in a given instance, but rather it can always apply any of its available operations, 
thus it can create new associations, see Fig.~\ref{fig:prev_verse_curr}(b). This demonstrates also the effect of 
past memories changing over time.

To achieve operationability, a minimal set of primitive operations is proposed, to function as basic operations, which can be the building blocks for more complex and composite operations/actions. 
Operations such as logical relations (AND, OR, NOT, "for all" universal quantifier, "there is" existential quantifier, (in)equalities, exists, count), flow operations\footnote{Note that loop operators and if operators can be formalized as classes, with general methods and properties. See for example at the bottom of the website: \url{https://www.w3schools.com/python/python_iterators.asp}, where an iterative class with \_\_iter\_\_() and \_\_next\_\_() methods is defined.} like loop operations (while, for) and if-else conditionals, mathematical operations (+,-,*,/,min,max,norm,log), and other relations.

Note, that loop operations are temporal ones, which should enable both using the past (current/past variable calculations) and the future (current/future variable calculations) for planning/simulating goals. For example, future calculations performed by halting the perception from senses, or ignoring it, and continuing in an imaginative state of mind. Although it exhibits the problem of also predicting future sensory input to this artificial reasoning process. Neuroscience claim a different mechanism to determine real verse not (imaginative/planned): by the strength of the input signals, i.e. exceeding some threshold \citep{dijkstra2023subjective}, based on the assumption that imagery is generally weaker, or less vivid, than perception.

Note, that "for all" and "there is" operations are first-order-logic binary quantifiers. These operation could become continuous features of classes, defined in some continuum, such as "all", and "there is" is a range includes: "most", "some", "few", "one"; and finally "none". Thus traverse it into fuzzy logic.
Using De-Morgan laws we can deduct that "all" and "there is" complete each other, while "none" is the opposite of "all".

\begin{figure}[H]
	\centering
	\includegraphics[width=0.95\textwidth]{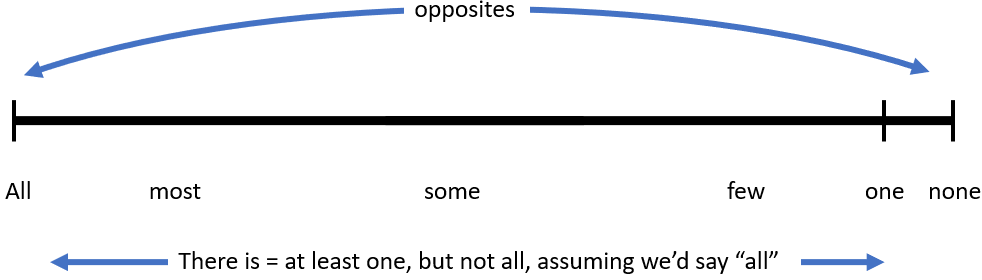}
	\caption{Quantifiers as features in a continuum}
	\label{fig:quantifiers}
\end{figure}

See in Fig~\ref{fig:quantifiers}.

Similarly is in space and time, such as "always", "often"/"usually", "sometimes", "seldom"/"rarely", "never". And in belief: "necessarily", "possibly" (are complete each other like "all" and "there is"). And in not operation: from "yes" to total "no". Eventually, all these examples, and different measures (such as of consolidation, certainty, relevance, etc), are all demonstrate continuous features. This fact may drive us to rethink \textit{DL}-only representation.

This set of tools described above, can replace DNN units and DNN's fixed structure, in a program-search process. It can be implemented for example via DL \citep{chollet2019measure} 
or via other AI tools. For example, via Reservoir network, a random mix of basic rule components/blocks, yielding an algorithm best describing an operation, within models or between them.  Another way to perform program-search is via Genetic Programming (GP). Also we could use the AlphaDev framework \citep{mankowitz2023faster}. Another option is Probabilistic Program Induction, which defines a Domain-Specific Language (DSL). DSL consists of primitives and grammar for programs that learn to generate data, similar to GP.  Another option is Neural Module Networks \citep{andreas2016neural}, where small modules perform on specific tasks, and they rearrange on the fly depending on the current problem. Another option is DreamCoder \citep{ellis2021dreamcoder}.
We could also consider the idea of Liquid Neural Networks \citep{hasani2022closed} or Neural Ordinary Differential Equations (NODE) \citep{chen2018neural} as an alternative to the modeling of temporal operations. These NNs learn spatio-temporal space at the core, unlike regular DNNs or RNNs.
Other possibilities are described in the Subsection~\ref{sec:FP}.

One work tried the Bayesian program search did it in the Omniglot challenge \citep{lake2015human}. 
However few limitation discovered: it needs lot of built-in primitives, lot of data to learn prior and likelihood distributions, and a lot of search (it is combinatoriaclly explodes), especially as the DSL becomes more expressive. One solution proposed is a neuro-symbolic one \citep{feinman2020generating}.

Another way is CopyCat \citep{hofstadter1995copycat} 
that uses current knowledge and top-down and bottom-up processes to answer an analogy task. The top is the concepts in the knowledge, while the bottom is the perceived information.

At the same time, prior knowledge is needed to be inserted, within operations and other \textit{elements}, by including: number of visits, uncertainty, rate of update, measure of dependence, and measure of consolidation. These variables are needed for meta-cognition \citep{ackerman2017meta}. The measure of consolidation is to prioritize different options associated to some \textit{element}, to separate the relevant from the irrelevant, e.g. admissible actions, which can be used inversely in creativity mode - by picking the less expected directions to follow.
Also, evaluating actions (basic or composite) in planning can be embedded in such relevancy measures.
Similar idea is presented in neuroscience, e.g. \citep{anderson2022investigating}, where the researches state: \textit{"Strong connections involve highly connected hubs of information-processing that are established when we learn about the world and become adept at solving familiar problems, while Weak connections have fewer neural linkages but enable flexibility and adaptive problem-solving"}.

Note the difference between the level of uncertainty of a class (feature or object) and unknowing. In the former, it is the dilemma when several options are possible (like multiple versions), and you cannot rule between them. The latter is similar, only the number of options is different. We can either regard it as the case of no options, or all possible options are allowed. This notion is important for formulation of goal states. For example, "x=?" means x is an object, it has a feature of equality (=) but the other side of it is unknown. See math example for this in Fig.~\ref{fig:math_problem}. Similarly, is in the riddle problem: \textit{"what am I? I don't talk, but I reply when spoken to"}. In this case there is a general class with the assignment feature to some unknown specific class, e.g. a watermelon or a cat.
This idea can extend further to any inference through exploration in knowledge map. That is, the filling of missing information, there are small problem-solving tasks, in which the missing information of some feature is unknown and in the process of become known.
Consolidation, see Section~\ref{sec:consolidation}, can be regarded as the level of (un)certainty in the class itself.


Moreover, action's admissibility is needed for several reasons. Firstly, due to the elimination of entry conditions necessary for an action to be performed, e.g. on which types (integer, string, etc). This is common in languages like C\#, where all variables' type must be pre-defined. 
Secondly, it is due to the ability to use 
High-Order Logic (HOL) \citep{miller2012programming}, as in $\lambda$-calculus, which removes any restrictions on an object's slot or action's argument. For example, there could be actions on actions, e.g. function as an argument in a function or a class decorators in python which are classes that can modify functions. Hence, relevancy is needed to constrain action's admissible space. Additional constraints can be embedded in relevancy: such as different rewards/costs for applying an action, time/resources constraints, and more. Note that the problem specifications determine the constraints, while the will (which can also be extracted from the problem state) determines action's features such as relevancy and reward.
And thirdly, it limits the search space in a given problem/request/situation, by restricting the number of possible sequences of actions. Hence, previous experience comes to help. Other action's features include previous evaluation measures.
Thus, perhaps, this is the trade-off between system 1 and 2, when automatic response is based solely on previous experiences, while system 2 comes to play when you want to rely less on these experiences, and to create new ones.

\subsection{Operationability in Deep Learning}

In this section we present an alternative approach, that exists in literature, to learn objects and actions - via embedding learning.

\begin{figure}[H]%
	\centering
	\subfigure[Separate learning in \textit{DL}]{%
		\label{fig:a}%
		\includegraphics[width=0.47\textwidth]{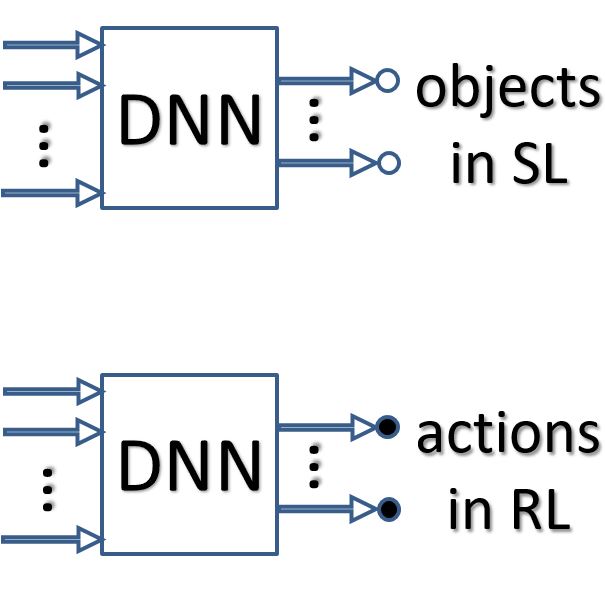}}%
	\hfill
	\subfigure[Updating both DNN of basic models and their complex structures]{%
		\label{fig:b}%
		\includegraphics[width=0.4\textwidth]{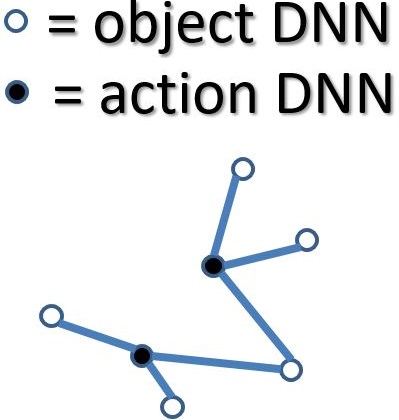}}%
	\caption{Object and Action learning}
	\label{fig:operational}
\end{figure}

In \textit{DL} objects or classes are usually learned via Supervised Learning (SL) and actions are learned via RL, see Figure~\ref{fig:operational}(a).

However, they should be learned together, where they are inter-wined (unlike scene graphs with pre-defined actions). Moreover, DNN models should be learned/updated on the micro and the macro levels, i.e. models of simple objects/actions or their complex combinations, see Figure~\ref{fig:operational}(b).

\begin{figure}[H]
	\centering
	\includegraphics[width=0.65\textwidth]{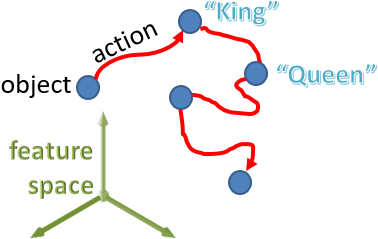}
	\caption{Operational features space, e.g. in word embeddings}
	\label{fig:operational_feature_space}
\end{figure}

If they are learned jointly, then the next challenge
, is how to represent operational data, i.e.: objects, actions, attributes, etc. See e.g. in Fig.~\ref{fig:operational_feature_space}, where the representation is embedded in some feature space, consisting of objects as points or other shapes, and actions as vectors.

One known theory \citep{hawkins2007intelligence}, is that neocortex was evolved from or on the basis of the old brain, i.e. the basic/instinct/primitive/reptile brain. Similarly, later in \citep{hawkins2019framework}, Jeff Hawkins extends the physical senso-motor system to the abstract thinking. In our opinion, infant’s first training is over the same principles in a physical environment. It is what prepares him for next level of abstracting similar principles beyond physical realm (as in Piaget's development theory \citep{piaget2003part}).

Interestingly, in \citep{bordes2013translating} they perform embedding learning over dataset of triples, each consisted of two objects $(h,t)$ and a relation $l$ between them. $h,t,l$ are high-dimensional vectors, or tensors. This embedding tries to learn in a high-dimensional space where each triplet will follow the distance equation of $h+l\approx t$.

Note, that what discussed above is a fixed embedding (such as word2vec). This is representation of knowledge (e.g. words, sentences) in a vector form. Which can make tasks like text retrieval or usage much more efficient than regular keyword search. However, contextual embedding is preferred, as in BERT, 
though it is difficult to encode such a space, as dynamic one.

Nevertheless, this is encoding, i.e. implicit representation of knowledge, of sub-symbolic features, not interpretable to us. Hence, \textit{MOM} proposes a more explicit representation, of symbolic features. It may be better, since perhaps all actions cannot be mapped into one feature space, if some actions have no connection between them (although perhaps all actions could be encoded in one highly dimensional space, providing a huge amount of features, such as SDR and VSA, see explanation below).

\textit{MOM} represents similar hypothetical space, 
where individual actions change the system's state. It supports different and unrelated types of actions, such as temporal-only, spatial-only, spatio-temporal, logical, etc.

However, this type of state-space representation is comfortable for transforming it to other representation, by a simple mapping. This could be beneficial for example in cases when to solve a particular problem, an appropriate representation is required.

More generally, this is just one option to represent state space, actions and will. Other options are also possible.

For example, the vector space representation of state space above, with vector operations within it is one option. Another representation could be by attaching some function to every state and thus representing the state space as some high-dimensional manifold or surface, where problem solving is occurs not via going from state to state but by optimizing this function and moving along its gradient to reach some optimum.
If we include the will field, than this function become a vector field function, where each state returns a vector, and the movement is according to the field and not according to the gradient. Finally, this function, as a measure of state, is dynamic, and it changes as a function of will, constraints, settings, etc. Eventually, yielding a manifold that contains all information of the current problem (will, constraints, settings) in its shape. This idea resembles the idea of connectionism verse waveness in neuroscience \citep{pang2023geometric}, where they claim that in macro the brain operates more in propagating waves then message-passing among neurons. Similar idea exists in general relativity, where the (gravitational) force field is the result of mass distribution in the universe.

Note that though gradient-descent (GD) method, in function representation, is easier to reach the goal state, compared to classic search algorithms, however, one enormous problem with it is local optimums and saddle points. Hence, one way to deal with it, is the ability to see the function not only locally, but also in higher resolution, if we climb to higher levels.

\begin{figure}[H]
	\centering
	\includegraphics[width=0.99\textwidth]{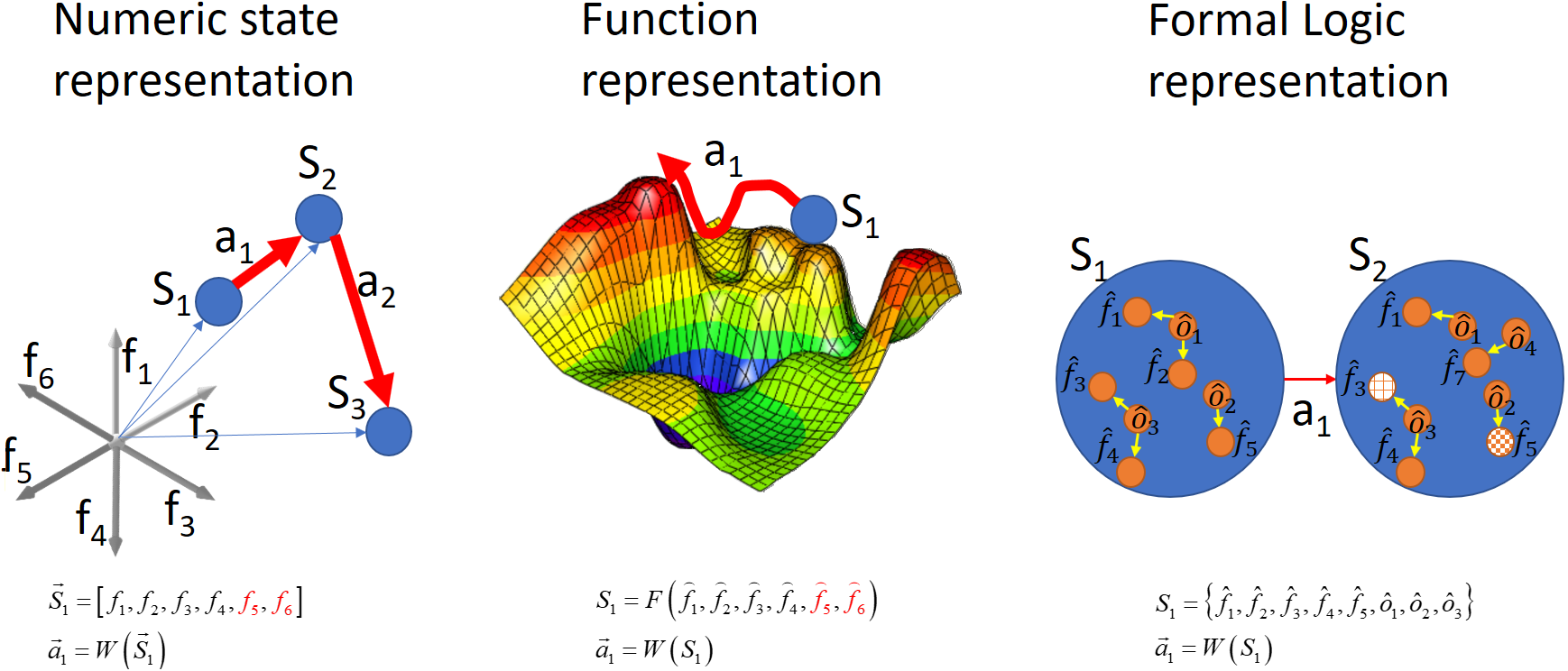}
	\caption{Different representations of state space, actions and will.}
	\label{fig:different_representations}
\end{figure}

All proposed representation so far is summarized in Fig.~\ref{fig:different_representations}.

\subsection{Modeling} \label{sec:Modeling}

If operationability is considered, as the addition of freedom to move in a 2D knowledge representation, then modeling is an addition of a new dimension, i.e. converting it to 3D.
It is about extending the "is-a" operation into a programming abstraction, as in \textit{OOP} (Object-Oriented Programming), or in abstract mathematics, such as algebra or category theory\footnote{
	Group properties include identity object to use compositionality of an operation and its inverse operation. This implies another duality.}.
Meaning, that while usually semantic networks represent this operation in a 2D graph, here the instances are totally separated from classes, and from classes of classes, and so on. Resulting with a multi-level of abstractions, while for simplicity two types of levels can be distinguished, in the final LTM, see Fig.~\ref{fig:Modelling_machines2}.

\begin{figure}[H]
	\centering
	\includegraphics[width=0.99\textwidth]{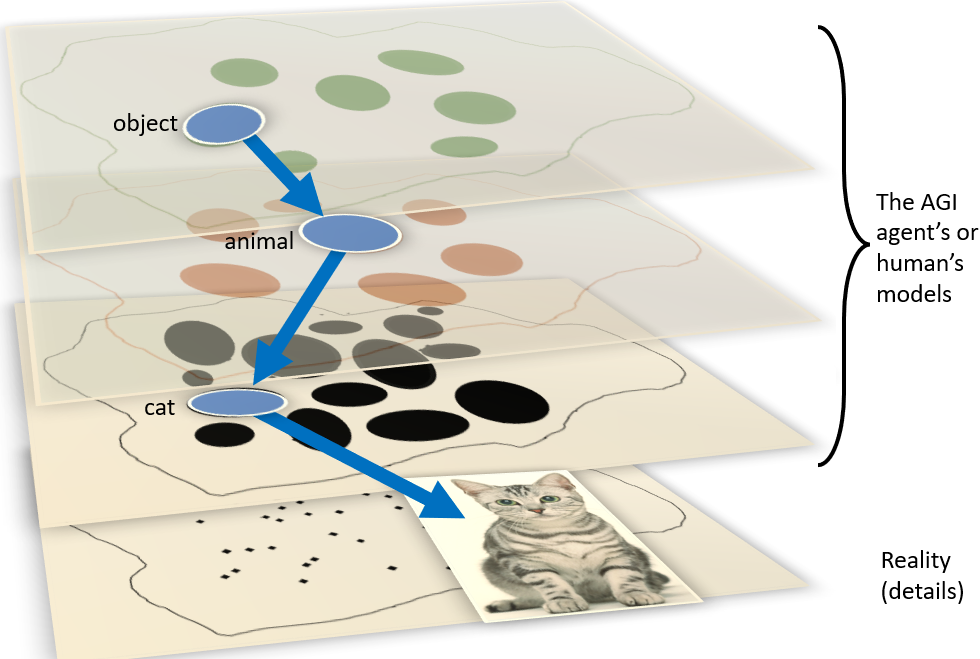}
	\caption{Example of how the models ordered in a hierarchy by abstraction or grouping.}
	\label{fig:models_hierarchy}
\end{figure}

We can see these classes and instances also in problem-solving or planning. If we disregard for a second the will level in Fig~\ref{fig:will_in_constraint_environment}(a), that directs all the models below, then we can see that the models are also comprised of levels, which are separated by the amount of abstraction or grouping. For example, we can see an abstraction or inheritance of classes, and an instance in the reality below them. We see the cat image in reality, then above it the “cat” class, then “animal” class above it, then “object” class above it. See this example in Fig.~\ref{fig:models_hierarchy}.

In summary, at first all different associations including operationability are describing objects, and then abstraction extends objects as instances into classes, which represent models. Especially, for complex cases like learning-to-learn, i.e. meta-learning, where abstraction is over similar tasks, or in transfer learning among tasks.
Consequently, unlike \textit{AKREM}, these full models are important for natural communication, e.g. for context-based conversations, where undelivered missing information (common sense) is needed to be filled.
For example, the sentence \textit{"The car was uncomfortable. The seat was too low."}
, indicates that "the seat" refers to the seat of the car.
Or in the sentence \textit{"The kid put his ball in a basket. Then the basket was moved to another room. Where is the ball?"}.
Or in the sentence \textit{"Nick is in the kitchen. Jim is in the garden. Later Jim went to the kitchen. Who is in the kitchen?"}\footnote{Note that in all examples above will can be regarded also as an object, with its features, such as the objects they are applied on.}.
These both cases involve inference to fill up the missing information, which can be regarded as mini problem-solving tasks, as mentioned in the ending summary of Section~\ref{sec:AKREM}.
Note that common sense is necessary also for intention comprehension, i.e. when we request something for an AI agent, in the usual minimalistic form, it should infer all the relevant constraints to accomplish its goal, e.g. for minimizing SPAM, it will not try to kill all people for that matter.
\footnote{Similar idea is developed by Shannon, defining information as a measure of surprise. Meaning, that while communication and perhaps while remembering, we tend to obtain and emphasize what is new.}

Moreover, to establish context, we must define "settings", or the context describing the overall situation. Settings can have also sub-settings and so on. For example, we can catch a ball moving fast in a game, since we define the settings we are in. In driving a car, we react fast with controlling the vehicle and attending the environment, since we first aware of our settings. This idea also aligns with the temporal hierarchy in \citep{EasyChair:7921}, where in parallel there are the lower layer of high-frequency (fast) perception-response, then above it and based on it is the mid-frequency (fast) perception-response layer, and so on. It means that the fast temporal scale work efficiently if the settings perceived in the layer above are correct, or fully "ready" for any data entering. Similar are the settings above the mid layer, and so on.
Settings include also the producing/receiving of language.

The settings are highly important to interpret correctly the situation and respond appropriately. In other words, the settings or the context dictate which features are relevant to interpret objects, and it relates or uses only the relevant actions for these objects. 

However, we could replace the "settings" with general planning for broad class of situations. Meaning, the driving or the sports game above can represent examples of abstract classes and actions. For example, in driving: \textit{"if I see a stop sign I must stop; if I see a crosswalk I must watch for pedestrians and slow down"}, and so on.

This extension has several implications.
\begin{itemize}
	\item First, in a broader perception from senses: from the basic recognition of instances in \textit{AKREM}, to a multi-level recognition of instances and classes.
	\item Next, every \textit{element} is learned and can be abstracted (into class), i.e. objects, actions, relations, and attributes.
	
	For example, action with attributes (such as bi- or uni- directionality of an action), relation with attributes such as strength, numeric attribute/values as class and its different types as sub-classes (integer, real, complex) in different bases, group types (strings, sets, tuples, lists, arrays, dictionaries) with their group operations (slicing, concatenation, union, intersection, difference, indexing, searching, sorting, replacement, etc.), and more. Although different types have also their specific attributes (e.g. set is a collection which is unordered, unchangeable, and unindexed), and their specialized operations, such as lower/upper case for strings, ordering operations for non-set types, key-value operations and nesting in dictionaries, etc.
	
	Moreover, there could be also non-deterministic classes, such as fuzzy sets and stochastic variables (characterized by some distribution, such as uniform/Gaussian distribution).
	\item Next, related to 
	attributes of objects or classes, treated as variables, we assume that there is no garbage collector in the WM. Hence, legitimate variables could be only attributes of classes, and no temporal variables are allowed (unless they are part of temporal classes/methods). Subsequently, our system's state is defined as the state of all active classes in WM, which leaves the methods as purely functional/declarative (not imperative), see more about Programming Paradigms \citep{Chen, van2009programming}. 
	
	Moreover, as stated here \citep{SCALFANI}, Functional Programming (FP) is much better for maintenance and updating of existing code. Which in our case means the ability to learn with as least modifications to previous learned "code", because naturally it requires too much effort.
	\item Next, this extension enables answering the question about how will is implemented in \textit{MOM}, see question at the end of Section~\ref{sec:will}.
	Alternatively to \textit{AKREM}'s hierarchy by will, which it is not clear how it can be implemented, it could be generated by abstraction, while the will/intention is serving as an additional and independent variable in the models that construct the hierarchy.
	The assumption here is that humans have will independent of their situation, while if will is deprived from \textit{AGI} agent, then this variable is a function of its sensory input, since it is embedded in the request from the human user.
	
	Additionally, feeling measures could be included, as influencing will. Our six basic emotions \citep{snoek2023testing} are: happy, surprise, fear, disgust, anger and sadness. Alternatively, they can be represented as some high-dimensional vector, e.g. via valence-arousal model in 2D \citep{cui2023emotion}.
	Although one paper \citep{coppini2023experiments} claims that the six universal basic emotions are too simplified to describe the wealth of realistic range of emotions, hence propose special ontology for that.
	
	Moreover, since there are several levels of will, correspondingly there are the main variable and secondary variables representing these wills, perhaps with different significance intensities, depending on the abstraction level.	
\end{itemize}

An example for different abstractions could be demonstrated in the word "bridge" \citep{mitchell2023abstraction}. The simplest form is learned from physical examples of bridges, but a higher one could extend into function and purpose. Hence it can be detected also as a log connecting two river shores or ants that form a bridge to fill a gap, or more abstract notions such as bridging a social gap.

Note, that both modeling (abstracting) and instantiating are localized energy-efficient memory structures. Meaning, that we do not separate episodic memory from the base or knowledge memories, nor the need to have separate memory for abstractions. All of those reside together. This may explain why it is not so simple to recall events (since they are all embedded in the classes and only small additional unique cues distinguish specific memories from them). Similar idea is stated in \citep{crossley2023circuit}: \textit{"To conserve resources, a brain must therefore be able to distinguish when it’s worth the cost to form a memory and when it’s not."}. It may mean that we have the trade-off between being based on current knowledge and updating it.

Finally, the novelty here, is that unlike \textit{DL} which performs program-search in an un-interpretable way, here however, additional inductive bias is introduced: separating of models (Section~\ref{sec:ModelSeparation}) and performing program-search to relevant actions (Section~\ref{sec:FP}), in consistency with other models and actions. This makes \textit{MOM} both usable and interpretable.

Moreover, \textit{DL} copes with data complexity by simple memorization, hence it requires huge amount of data, to represent all different scenarios in the world. Humans on the other hand, try to learn the basic ingredients that are common to all the knowledge they receive, to construct any new information, such as through inference and aggregation (grouping and abstraction). This is what separating models means, along with compositionality to construct complex models.

Side note: Meta-cognition is a similar phenomena to abstracting, since it is also in levels. The difference, that it is level of thoughts. See also Marvin's "Emotion machine" \citep{minsky2007emotion} about the levels of thinking\footnote{This yet again enforcing the idea of humans as being merely observes. Not only of physical reality through senses, but also of the thoughts themselves, which is why the thoughts can get hierarchical. Since we can view always from an upper layer on any thought.}.
First-order thinking is simple experience of sensory input, while second-order is analyzing/contemplating a memory in which such experience occurred. Eventually high-order thinking is just pilling up memories of memories (or hierarchies), watching one over each other. Hence, it is like extension of the hierarchy representing a memory, into a hierarchy of hierarchies. This occurs also in stories and any messages, e.g. \textit{"He said that he was thinking about sailing"}.
Note that meta-cognition, as probably any hierarchical process (as abstracting), evolve gradually from childhood.

Alternatively, high-order thinking can be regarded as any hierarchy, i.e. connected to different parts, and not separated. For example, the sentence \textit{"Bill claims his cat chases mice"}, is attaching Bill to a group object, which consists of the details in "his cat chases mice". See this example illustrated in Fig.~\ref{fig:high_order_thinking}.

More generally, for simplicity, there are only two cognitive operations with memories: construction and retrieving. This way the examples above can be explained as following:
\begin{itemize}
	\item 1st order thinking (observing) \hspace{5.0em} $\rightarrow$ construction, from basic LTM concepts. 
	\item 2nd order thinking (recalling a memory) \hspace{1.5em} $\rightarrow$ retrieving or traveling the different hierarchies/memories.
	\item 3rd order thinking (meta-cognition) \hspace{3.3em} $\rightarrow$ construction, attached to past memories.
\end{itemize}

Finally, just as any abstract thinking, e.g. "what if" counterfactual thinking, or high-order thinking (meta-cognition), knowing what others knows, etc - are all based on the same mechanism of constructing in height (hierarchy). This operations used in many cases, e.g. in comparing old knowledge to new incoming knowledge, and update it accordingly, or in any problem-solving scenario, such as imposing time constraints for solving a problem in high-order hierarchies.

\subsection{Summary of operational modeling} \label{sec:Summary_op_modeling}

\begin{figure}[H]%
	\centering
	\subfigure[Frames or classes in \textit{OOP}]{%
		\label{fig:a}%
		\includegraphics[width=0.87\textwidth]{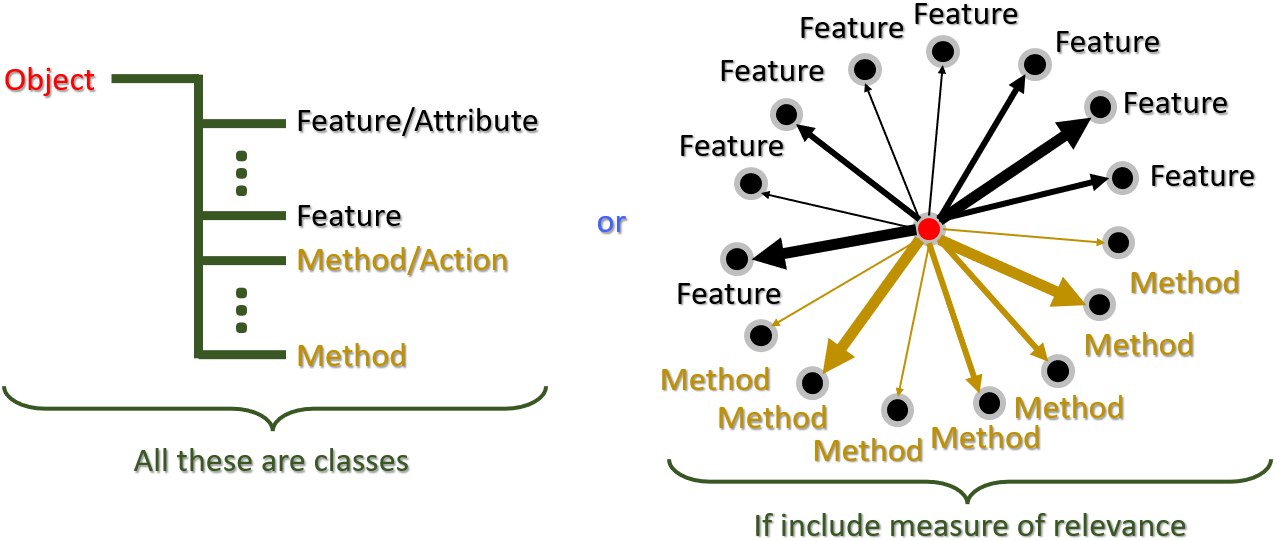}}%
	\hfill
	\subfigure[State and action definitions]{%
		\label{fig:b}%
		\includegraphics[width=0.77\textwidth]{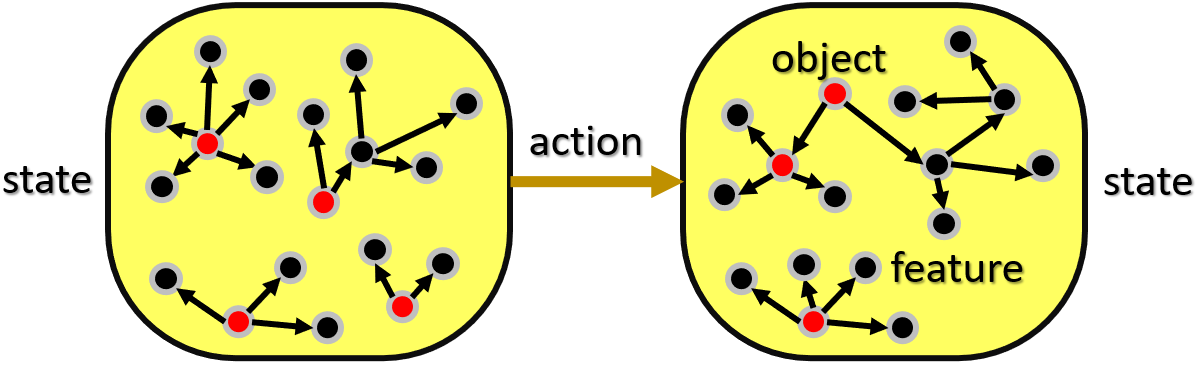}}%
	\caption{State, object and action definitions in \textit{MOM}.}
	\label{fig:oop}
\end{figure}

In \textit{MOM}, we represent each object or class along with its features and actions, or in the equivalent graph view on the right, see Fig.~\ref{fig:oop}(a). Also we can see how each element of an object has its own measure of relevance, or more generally it could be other measures, all treated as features in the spoken class.

Also, as in \ref{sec:op_and_modeling_summary}, we present in the following some additional definitions.
\begin{itemize}
	\item State = group of relevant object classes and their feature classes.
	\item Action = learnable algorithmic class.
	\item Will: derived implicitly from state, and it dictates relevant actions.
\end{itemize}

After the definitions of knowledge basic elements as described in \ref{sec:op_and_modeling_summary}, we can represent a state as group of objects, and actions as those that operate on one or more objects in a state.
The actions are learned, and the will is specified usually in the first states.
How will is derived from the state is elaborated upon in \ref{sec:Will_derivation}. 
As we showed previously, the will and the knowledge models are aligned, so that the will would be guiding the state to select the most relevant actions.

Additionally, actions operate in a continuous spatio-temporal space. Meaning, just as it can act in any direction and region in space, similarly it is in time: it can be either immediate or include past/future of any scale. This also enables causality modeling.

\begin{figure}[H]
	\centering
	\includegraphics[width=0.85\textwidth]{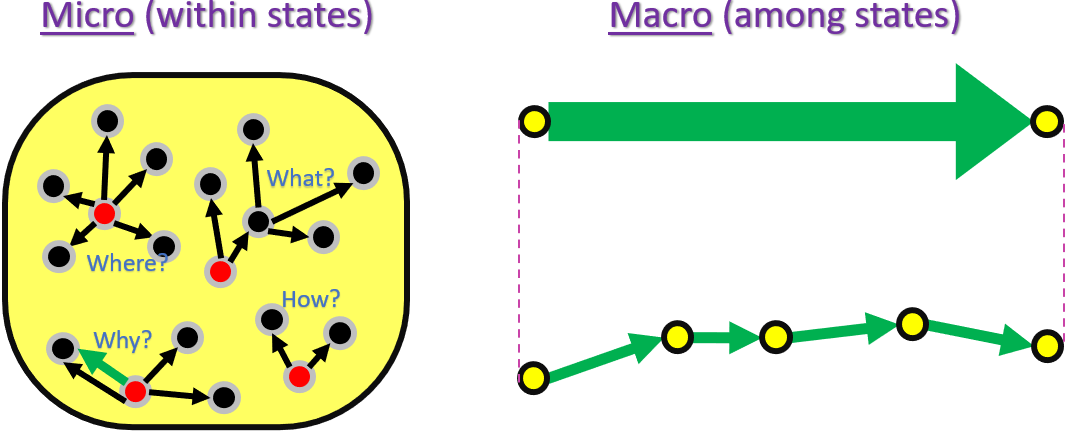}
	\caption{Every change (small or big) is accompanied with a will.}
	\label{fig:will_effect}
\end{figure}

Finally, the inference process, described in \ref{sec:basic_wills}, occurs not only in understanding, but also in planning, specifically to comprehend the problem and the goal. Therefore, we can view understanding/planning as the macro processes, with external will supplied to the agent from the outside, either in the form of some problem to solve or in the form of hidden will that needed to be extracted. On the other hand, inference is a micro process, derived from internal will (to make sense), and it occurs during the construction of the macro process described above. The micro process evolves within states, while the macro process evolves between states. See summary in Fig~\ref{fig:will_effect}.

In conclusion, every change, either in features or due to actions, is always guided by will, or as frequently as possible.

Also, note that hierarchy is what makes any partial model to be a full one. For example, we can describe a robot/person as: \textit{get out of apartment $\rightarrow$ went down an elevator $\rightarrow$ walked 100m straight $\rightarrow$ turned 90 degrees left $\rightarrow$ walked 200m straight $\rightarrow$ enter a door $\rightarrow$ walked ... $\rightarrow$ picked onions $\rightarrow$ ...}

This example is how any machine can grasp sequence of actions, as seen in the lower right part of Fig~\ref{fig:will_effect}. But for human it is clearly not telling much. We want to understand what was the purpose of this sequence. Hence, telling him \textit{"The robot went to buy onions in a nearby store"}, is what makes this model full, see the upper right part of Fig~\ref{fig:will_effect}.

\subsubsection{Will derivation from state}  \label{sec:Will_derivation}

As mentioned above, will is derived implicitly from state, i.e. from perception. From the most direct approach, as in the examples in \ref{sec:simple_examples}, with will derived from the first constructed state(s), to the most indirect one, where it is very difficult to reconstruct it, and it is usually derived from long process of aggregation\footnote{Due to the complexity of state to will derivation, it is also suggested to use flexible tool as DL.}.

Hence, we can derive the following equation:
\begin{equation}
	\text{state} \Longrightarrow \text{will} \Longrightarrow	\text{action}
\end{equation}

Our puzzle here is to figure out how the will is derived from state(s). As stated previously, will is learned (given only the states), in the evolution phase of \textit{MOM}. However, this claim is too general, while at the same time in many examples the will is specified explicitly in the state, i.e. the learning is very straightforward. See examples of the math problem, the robot request or the simple problems in \ref{sec:simple_examples}. This is true mostly for inanimate objects, see Fig~\ref{fig:will_types}.

\begin{figure}[H]
	\centering
	\includegraphics[width=0.99\textwidth]{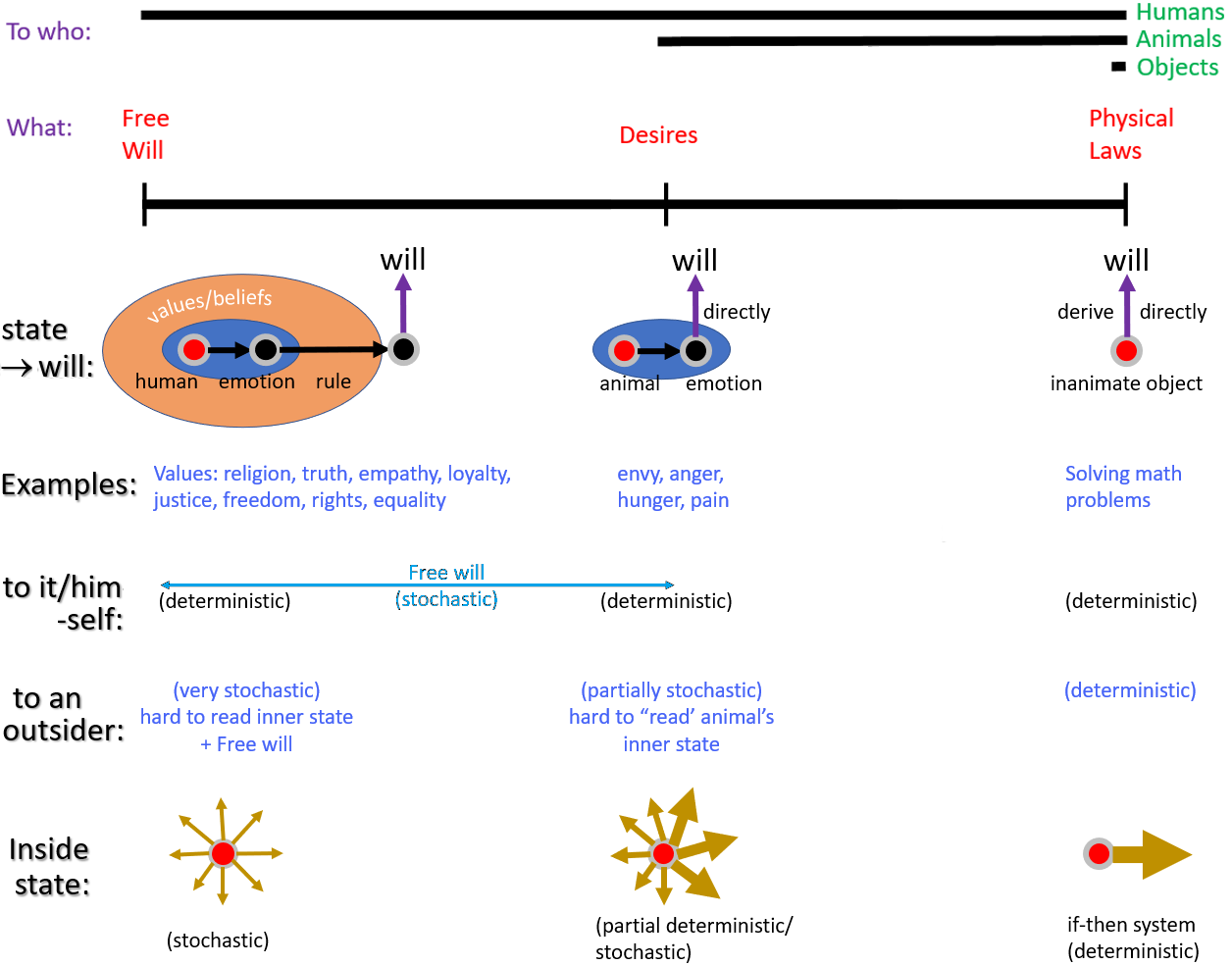}
	\caption{Will derivation from state.}
	\label{fig:Will_derivation_from_state}
\end{figure}

But there are many cases, that the will is implicit and hidden from us, such as for animals and humans.
See "state$\rightarrow$will" row in Fig~\ref{fig:Will_derivation_from_state}\footnote{Fig~\ref{fig:Will_derivation_from_state} is an extension of Fig~\ref{fig:will_types}.}, where animals are driven by hidden inner emotions.
Though humans can also act upon emotions\footnote{While Descartes’s philosophy of humans states that there are two modes of thinking: understanding and will, Hume’s theory on the other hand makes the rationality the slave of emotions, instead of the opposite common view where it is who's in control. This is our assumption also: the will dictate everything. It then use cognition as merely a tool to fulfill itself or to justify itself, no matter how selfish or horrible it could be.}, however some people acquire a second nature, that override emotional responses.
It is similar to the hierarchical mesh theory of free will \citep{frankfurt2018freedom}. It is usually derived from some set of values or beliefs. This set is represented by rules, and it is similar to language rules (see Fig.~\ref{fig:Language}), acting as an interface between two entities.
Also, just like language, this interface may be external tool set to the knowledge map, just like any additional constraints (time and other resources, and values for admissible actions to use) in problem-solving tasks. Hence, all of these may be placed in the top-down region as described in Fig.~\ref{fig:MOM_hierarchy}.

However, any set of rules is also deterministic, and so is the derivation of will from emotions in animals. See "to it/him-self" row in Fig~\ref{fig:Will_derivation_from_state}.
So how animals and humans models are still stochastic?

Indeed these three mechanisms are deterministic by definition.
However it is not the case for an outside observer or listener, see "to an outsider" row in Fig~\ref{fig:Will_derivation_from_state}.
For the animal it is usually hidden, while for the human there is also a second factor, the free will, struggling between these mechanisms, to choose to behave accordingly to values or not.

Finally, all these mechanisms as perceived by an outsider can be modeled in three forms of states and their admissible actions, ranging from totally deterministic as an "if-then" rule system, up to totally stochastic, see "inside state" row in Fig~\ref{fig:Will_derivation_from_state}. This represents admissible actions as either multiple or single.
Note, that rule-based system deals only with the deterministic rules, while ignoring the fuzzy rules, where more then one outcome exists.

For some examples of these three forms, see sub-section \ref{sec:special_examples}.

Note also, that studies show that including emotions in some event assist in memorization of it. It can be seen when monotonic type of talking reduce the motivation to remember or even to listen (may even cause drowsiness). This effect might just support the importance of will in any message.


\section{Practical MOM in Mature state} \label{sec:Examples}

In this section we start with some thorough examination of the proposed knowledge structure in \textit{MOM}'s mature intelligence state, especially grouping (either in space or in time or in both), following with different examples.

\subsection{Knowledge structure}  \label{sec:knowledge_structure}

Note that causality is a special case of modeling, a spatio-temporal one, where re-occurrence is consolidated. More generally, re-occurrence helps in learning both static objects and dynamic basic/composite events (equivalent to scenarios/scripts in \textit{OOP}). 
Usually learned events are only the basic ones, i.e. in micro scale, because compositions of these are too many, and it becomes a combinatiorical problem to remember all encountered compositions. Only few, very common composite events remain. Note that events can also represent common scripts, recipes and algorithms (can be used in problem-solving for example). Which is why in our opinion, basic algorithms, may use \textit{FP}, while time-sequences, event-related algorithms, that similar to planning, are of \textit{OOP} type.

\begin{figure}[H]
	\centering
	\includegraphics[width=0.75\textwidth]{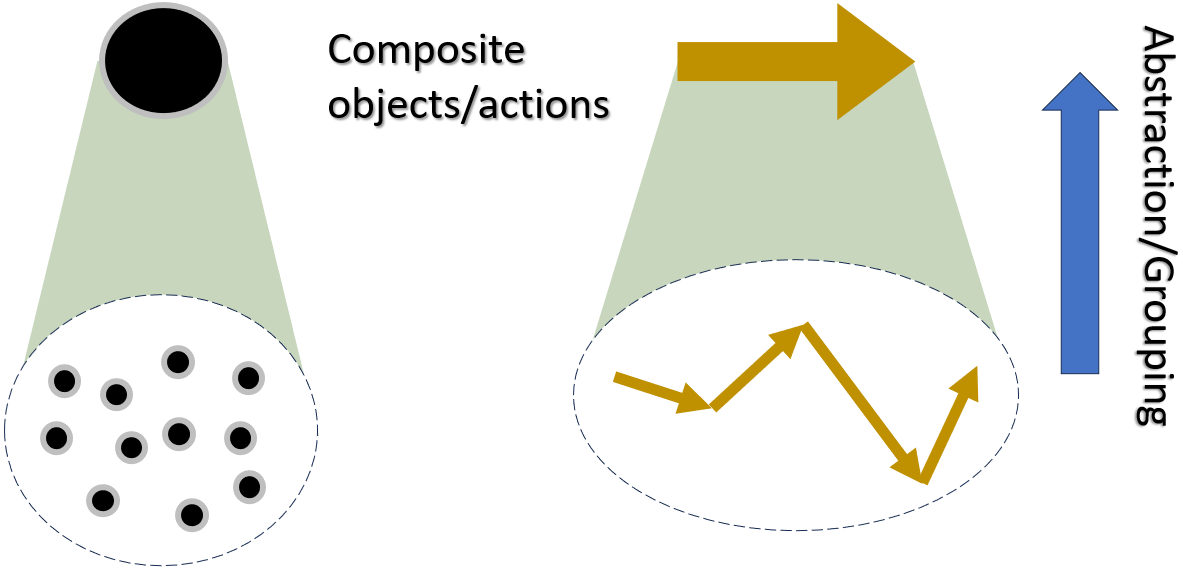}
	\caption{Aggregation of basic elements to composite ones.}
	\label{fig:Aggregation}
\end{figure}

See more in Fig~\ref{fig:Aggregation}.

Last sentence means, that aggregation in Fig~\ref{fig:Aggregation} is not only temporal for actions, as in Macro in Fig~\ref{fig:will_effect}, but can be also compositional, as shown in Fig.~\ref{fig:COMPOSITIONALLITY}, via \textit{FP} for example.

\begin{figure}[H]%
	\centering
	\subfigure[Example of full algorithm/plan/action]{%
		\label{fig:a}%
		\includegraphics[width=0.77\textwidth]{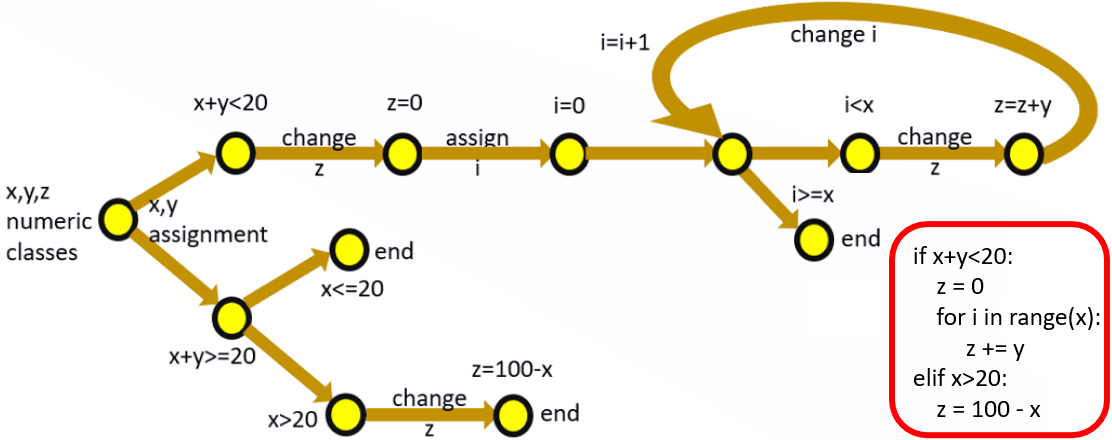}}%
	\hfill
	\subfigure[Composition of actions]{%
		\label{fig:b}%
		\includegraphics[width=0.87\textwidth]{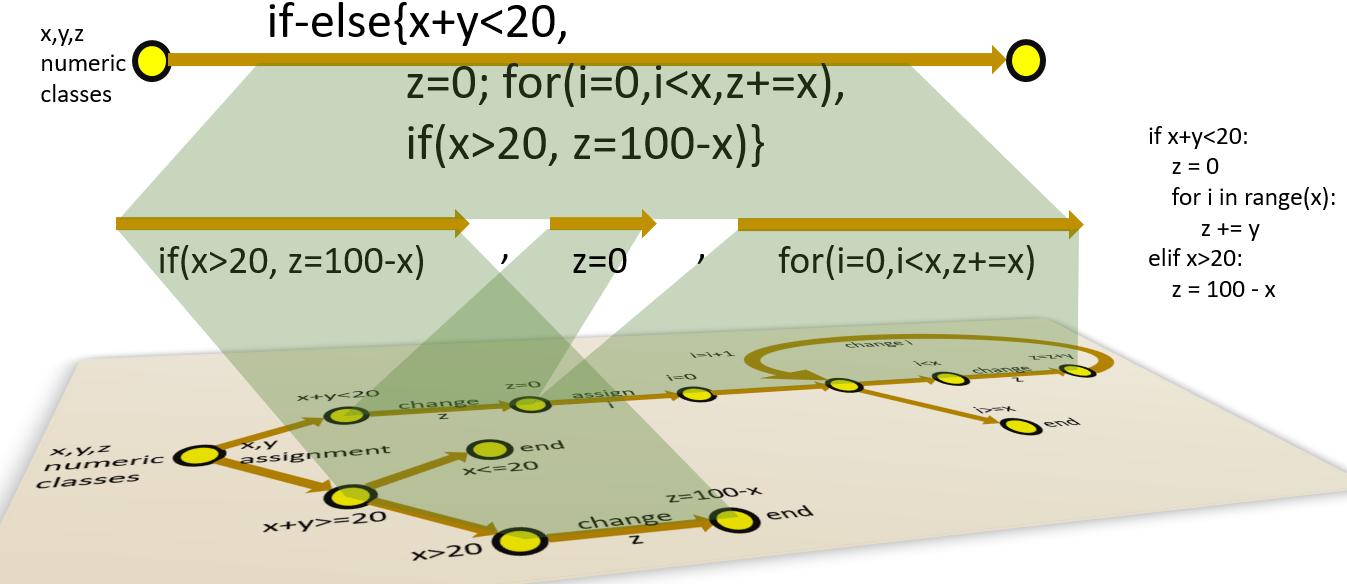}}%
	\caption{Hierarchy of composition in actions.}
	\label{fig:Aggregation2}
\end{figure}

We can see an example for action compositionality in Fig~\ref{fig:Aggregation2}. A given plan or composite action is illustrated in Fig~\ref{fig:Aggregation2}(a) as a directed graph, with the python code of the algorithm in red rectangle. The composition of actions is illustrated in Fig~\ref{fig:Aggregation2}(b).
More generally, as seen in Fig~\ref{fig:Aggregation2}(a), actions as algorithms are represented via flow charts, where there are many possible scenarios. Such abstract algorithm is seen also in the following demos, e.g. in math problem.
Also, though we have seen the flow only in the lower implementation level of actions, it is not limited to these levels, and can appear also in higher ones.

Note that temporal compositionality of actions is similar to the simple Chain-Of-Thoughts (CoT) in LLMs. But as we know this is only one branch in diverse thinking while planning for solution, see Fig.~\ref{fig:system1_and_2} and Fig~\ref{fig:system1&2_solution}(b). This is analogous to the Tree-of-Thoughts (ToT). But this is also limited to temporal or sequential type of thoughts/actions/algorithms. More generally, it can have flows in any direction besides the forward one. This is what introduces a more general scheme, called Graph-Of-Thoughts (GoT) \citep{besta2023graph}.
GoT models different thought processes, such as divergent-convergent thinking (generating ideas then converge into one final idea), looping over a specific thought to enhance it.
Note that all these processes in LLMs, describe cognitive flow operations, while we mostly concentrate on the final trajectories of thoughts, which can be memorized for later retrieval. We regard cognitive processes as in generally learned, to develop some intuition.
Finally, this complex process of generating multiple thoughts and evaluating them on-the-fly, is the macro process of entire thoughts. Similar process is exists in the micro in LLMs, e.g. beam search and alike - in producing single words.

Moreover, additional prior knowledge can be inserted in recording events: date/place markers for later recollection\footnote{It could be done for example by attaching the relevant spatial and temporal attributes to the particular event class}. Especially it is needed for temporal relativity between events, i.e. for after/before relation.

Some clarification so far: objects regarded as static, while their actions can be involving time, e.g. a dog object has a "running" action which when applied transforms its current state to a new state in the future\footnote{This running action is attached to the dog object as a feature, as long as the action is operating.}. Scripts are classes that contain a sequence of causally coherent event classes, i.e. each event causes the next one
\footnote{Philosophically, David Hume’s theory of causation states that cause and effect relationships are not a product of natural law or universal truth, but are instead based on the necessity that we associate events based on experience, i.e. it is yielded empirically, or: we infer it from past experiences. Hume argued that causation does not exist in the physical world, but rather is simply a construct created in our minds. On the other hand, causality by Immanuel Kant is a priori embedded in our mind, i.e. cause-and-effect is a prior assumption, even before data entered our senses. In our humble opinion, it does not matter that much, since causality indeed inferred (either a-priori or not) mostly unconsciously, e.g. at sleep. What matters is that causality is part of the will system, where it is a tool to gain control, or realize the will.}. But in order to be fully informative, it includes also all the background of these events: entry and exit conditions, objects involved in the events, and tracks - variations of the script, i.e. different options of event sequences. In other words, it is a flow program of "if-else"s made out of different events.
An event can be either basic, i.e. made out of primitive actions, e.g. via simple sentence \textit{"The dog ran home"}, or it could be composite (a scene), made out of basic events. A script can be part of some object, e.g. Eat-at-Restaurant Script is Event-Sequence slot in a Restaurant class.

Additionally, since scripts have many possible tracks to follow, it can be modeled usually via method, representing nested if-then structure (tree-like), to represent all possible common tracks. Actually, even the default sequence of events in a script is already a method, and it is belong/applies to the script object. Hence, knowledge should not be represented merely by nodes and edges as a simple graph, but as a hyper-graph \citep{bretto2013hypergraph}, or as nested graph \citep{angles2009nested}, 
where entities can also be groups. So we can abstract our original meaning of grouping (part-whole) not only on static info but also on a dynamic one, i.e. group of events.
This grouping introduce yet another type of hierarchy in our system, besides the abstraction.
Or we consider grouping as matching some general class to the given details.
Hence, there are two independent types of hierarchies that can evolve in the system. Both could produce temporal classes/instances, which do not need special memory, but the usual one, where their usefulness decay rapidly if not reused.

Moreover, gaining essence out of grouping (as it is in \textit{AKREM}) is a difficult challenge, but also important for summarization tasks, and for retrieval from memory. One way it could be done is by generating the meaning of the group, e.g. by formulating it as a problem-solving task, where we describe the problem/goal/trajectory that this group represents.


However, in \textit{MOM}, these formal definitions play out differently.
First, there are no conditions, but rather simple associative triggering of relative objects (either static or dynamic).
Also note, that beyond inner associations (attributes of an object) or specific types of relations discussed in Section~\ref{sec:Associations}, there could be also associations among concepts, either basic or complex/grouped, mainly derived from past memories (hierarchies).
Second, it is all based on abstraction principle. At the lowest level it is an accurate sequence of object and action instances, which is stored as episodic event. At an upper level it can abstract different parts of this or/and similar events, e.g. same sequence of actions but different objects participating them. A higher level of abstraction can add also abstraction of the specific actions/events that are involved in the previous episodic events. Even higher abstraction can involve only the major events that are shared among many episodic events, in a form of a group perhaps, and not necessarily with specified order.

Note, that this generalizes the recognition in the system, to be not only spatial (of static objects) but also temporal (of events), hence for example an action or a set/sequence of actions can trigger some event object/class. An event class like any class can have its own actions and properties, for example an action can contain some admissible actions that follow it or precedes it.


Also, since we can have hierarchical structure not only for classes and groups, but also for actions, then will field affects this structure also. It diffuses down the hierarchy and learns also the directionality of inner actions. That is, there is alignment not only on the actions in high levels, but also on sub-levels until the lowest level. Hence, we result with some type of dependence/alignment between the upper and lower levels actions' directions.

This hierarchy is what illustrated in Fig~\ref{fig:will_in_constraint_environment}(a), where coarse models are classes that can represent a group of classes in some structure (e.g. a group). As stated there, it is easier (less mental load) to plan in such hierarchy, since you can plan via small amount of large chunks (of sub-classes), then descend and plan also for small amount of classes, only smaller chunks, and so on.

\subsection{Examples of MOM in Mature state}



\subsubsection{Language as separated from knowledge}

Before presenting the following examples, we should note that language itself is a set of models, likewise are the translation between them. These models are most probably symbolic, i.e. included in the classes of our knowledge models. Although it might be partially also sub-symbolic (and perhaps be part of the cognitive models, see \ref{sec:ModelSeparation} and Fig.~\ref{fig:Modelling_machines3}). Either way, these models are utilized in every input/output language operations, such as reading or listening for input or talking/writing for output. It means that the knowledge itself is not represented in the form of language, since syntax and grammar are specialized for each language, while the knowledge itself should be invariant to the language we use to express it. Language is like our interface to the outside world.
Though thoughts themselves can be accommodated with language it is still not rooted in the knowledge itself, which is far more flexible. It, or "talking thoughts", is simply our way of simulating interaction with the outside world, similarly to any planning or imagination activities.

\begin{figure}[H]
	\centering
	\includegraphics[width=0.4\textwidth]{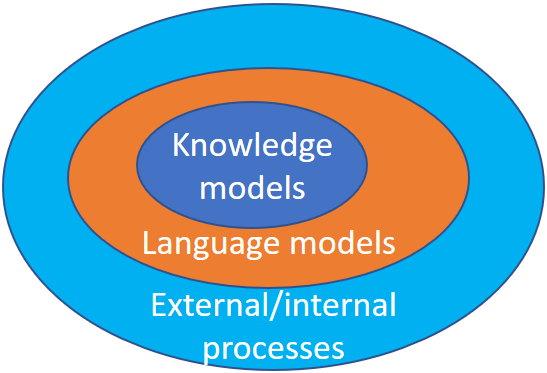}
	\caption{Language as separated from knowledge.}
	\label{fig:Language}
\end{figure}

In summary, Language is separated from knowledge or the actual cognitive operations, since these operations are more fundamental processes than language’s specific structure.
Language is merely a tool to communicate thoughts (to myself or others).
This interaction is visualized in Fig.~\ref{fig:Language}, where external process are e.g. senso-motoric activities and internal are e.g. planning, imagining, simulating, etc.

This side note purpose in relevance to the following examples, was to emphasize that we ignore the language to/from conversions with knowledge map.

Another note, is that knowledge represented in \textit{MOM}'s mature state (objects, features, actions) and wills – are all learned, in the evolution phase.
Hence the knowledge in the following examples is merely hypothetical demonstration, while in reality it could be different.

Finally, naming, or the inclusion of language to describe objects/actions/features is not necessary in model learning. The learning still occurs, even before its name is introduced, or if it is forgotten for some reason.

\subsubsection{Simple examples} \label{sec:simple_examples}

\begin{figure}[H]%
	\centering
	\subfigure[]{%
		\label{fig:a}%
		\includegraphics[width=0.97\textwidth]{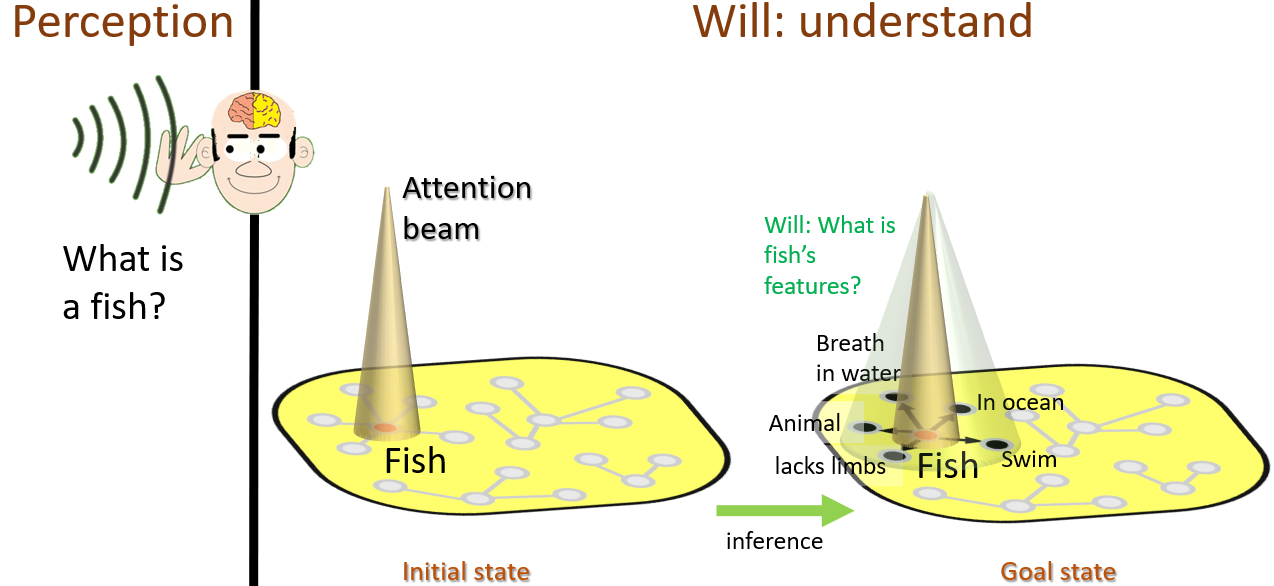}}%
	\hfill
	\subfigure[]{%
		\label{fig:b}%
		\includegraphics[width=0.97\textwidth]{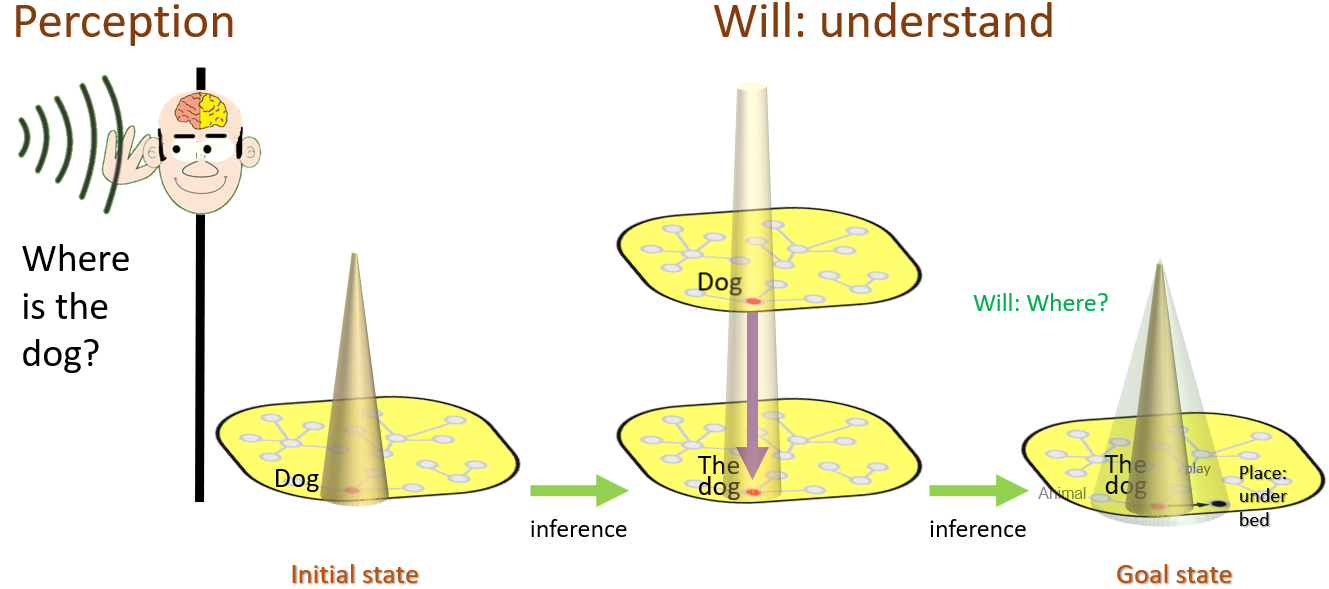}}%
	\caption{Simple examples.}
	\label{fig:simple_examples}
\end{figure}

The simple examples are represented in Fig.~\ref{fig:simple_examples}. Also we can see how each element of an object has its own measure of relevance, or more generally it could be other measures, all treated as features in the spoken class.

In Fig.~\ref{fig:simple_examples} examples, the cartoon of a head emphasizes that an attention is split: one part is perceiving incoming information, while the other part is trying to comprehend the total accumulated information received so far (by applying inference simultaneously with perceiving). Usually this part of attention is in the background (sub-conscious).

In Fig.~\ref{fig:simple_examples}(a) example, it first triggers the fish object, then guided by "what" question it extracts the relevant feature(s). Next it is ready to use language models to phrase the goal state in a correct grammatical form. This step is avoided as mentioned previously.

In Fig.~\ref{fig:simple_examples}(b) example, the dog class is triggered, and then an instance of it, assuming it to be the last discussed dog, is accessed.
Again, we use will to extract the relevant feature. This time "where" question directs us to a specific feature: Place, which is a class by itself, with the feature: "under the bed".

Next, we demonstrate high-order thinking, which is discussed in \ref{sec:Modeling}. 

\begin{figure}[H]
	\centering
	\includegraphics[width=0.95\textwidth]{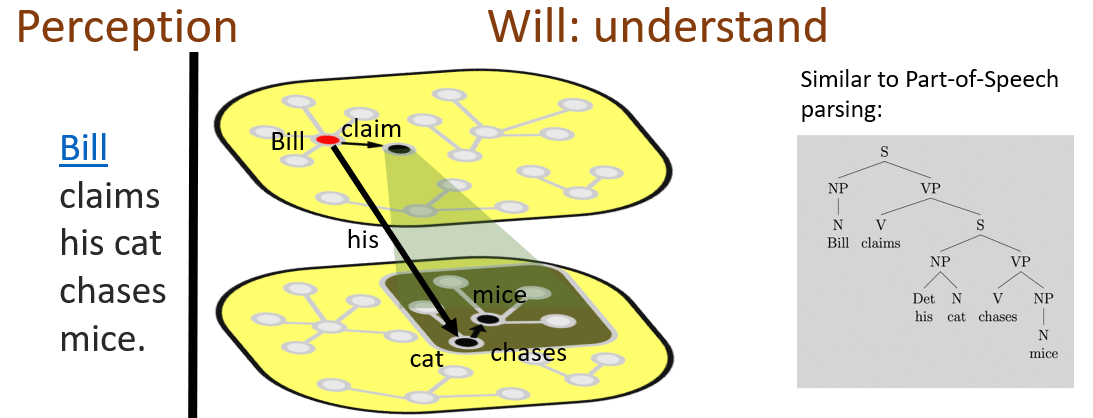}
	\caption{High-order thinking via grouping.}
	\label{fig:high_order_thinking}
\end{figure}

Fig.~\ref{fig:high_order_thinking} is an example of this case. To simplify the image we present only the final state of the given message. In the example we show that high-order thinking can be expressed as a hierarchy via grouping.
Actually, the group can be expressed in the same level.
Also, this hierarchy is similar to part-of-speech (POS) parsing.

\begin{figure}[H]
	\centering
	\includegraphics[width=0.95\textwidth]{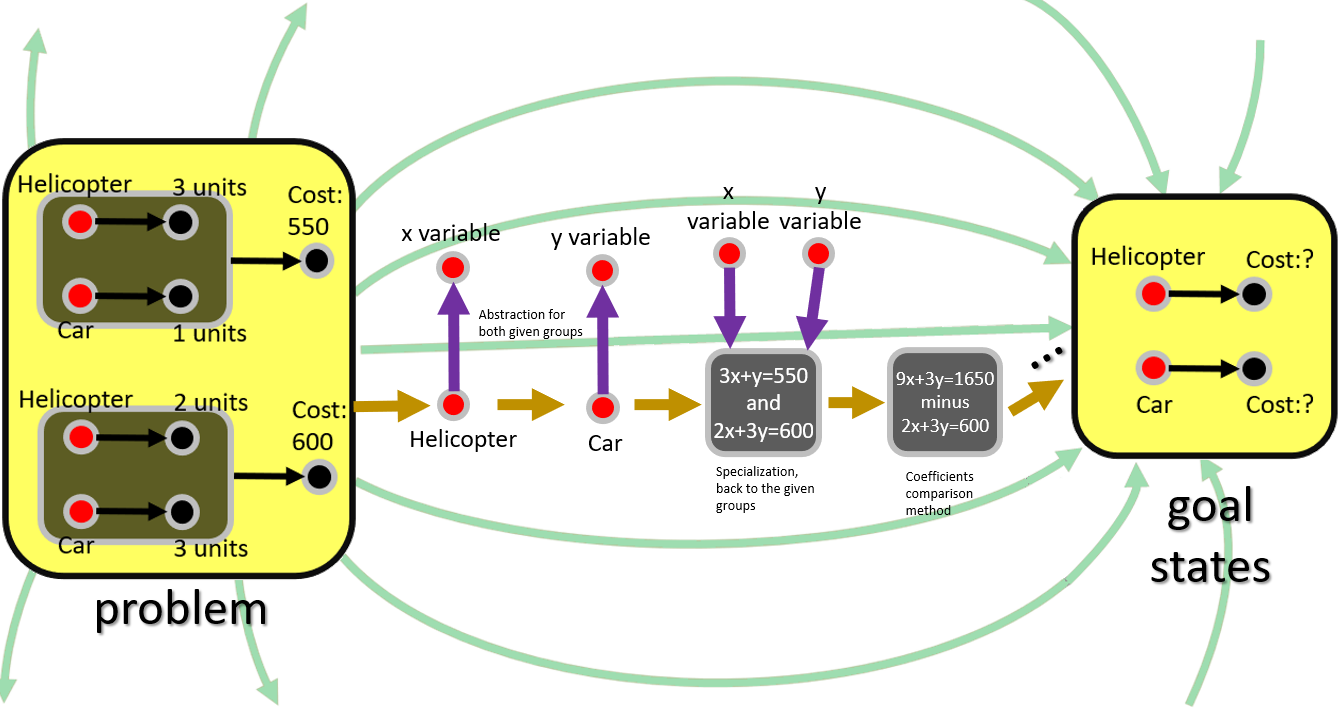}
	\caption{Math problem example.}
	\label{fig:math_problem}
\end{figure}

Next, an example of problem-solving task is presented in Fig.~\ref{fig:math_problem}. This example is a math problem, described as follows: \textit{"Math problem: given that 2 helicopter toys and 3 car toys cost 600. also given that 3 helicopter toys and 1 car toy cost 550. How much the helicopter and the car toys are cost each?".}

In Fig.~\ref{fig:math_problem} we see several things. First, we see a group object (of Helicopter and Car objects, colored in brown) in the problem state, and its cost feature. Next, we see the will induced for this specific problem. Next, we do not see it in the figure, but there is inheritance:  Helicopter and Car objects derived from Helicopter and Car classes, respectively, and these classes derive from the Toy class.
Finally, we see a short visualization of one proposed solution, displaying only the main elements in each intermediate state.

\begin{figure}[H]
	\centering
	\includegraphics[width=0.99\textwidth]{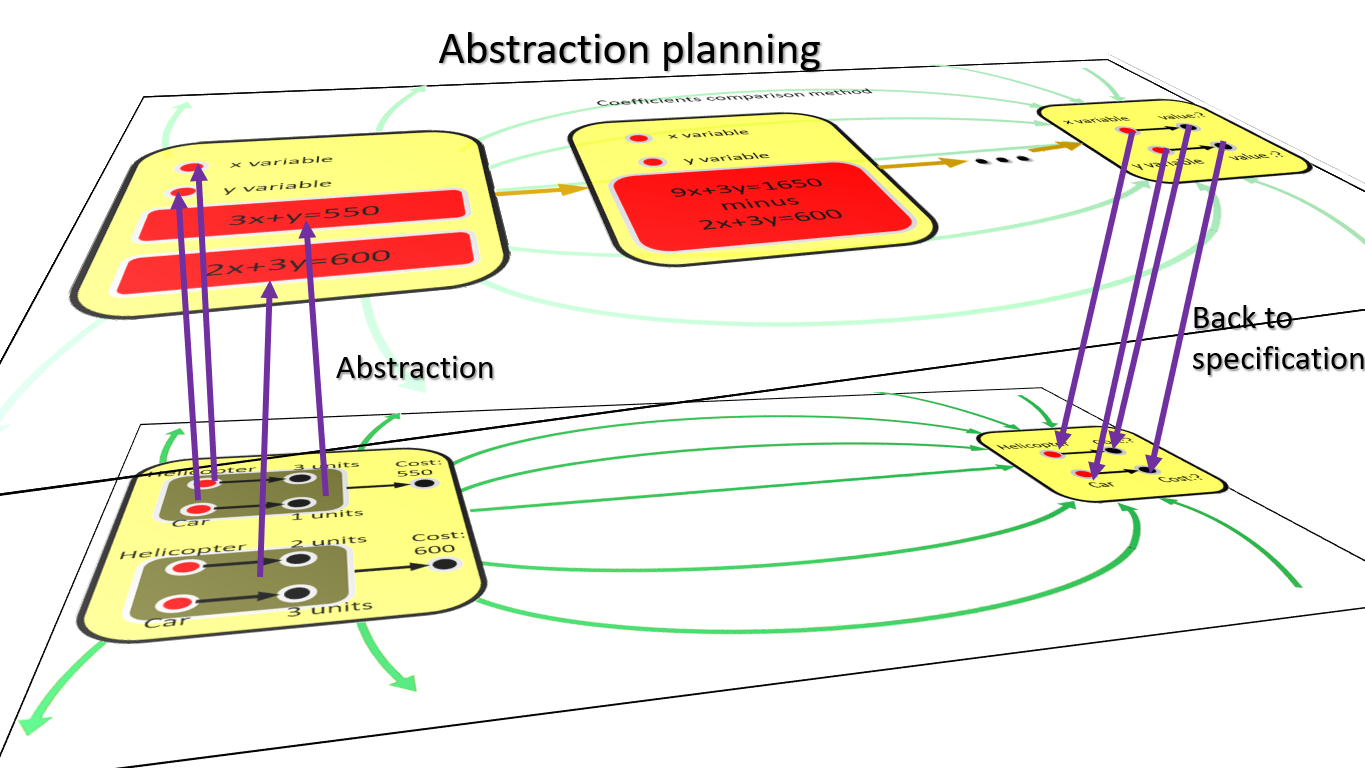}
	\caption{Math problem solution via the hierarchy of models.}
	\label{fig:math_problem2}
\end{figure}

Actually, this example is similar to analogy, i.e it works in both instance and in class levels. It first transform the instances described in the problem into abstract numerical classes, then it uses actions/algorithms to solve this problem in its abstract form, then after solving it, it returns to the instance level to reply the answer. See this example illustrated in Fig.~\ref{fig:math_problem2}.

Just a small note. As we mentioned in Section~\ref{sec:MOM_intro}, there are unidirectional features/actions, meaning not every action is easily can apply inverse action to retrieve the original concept. We can see in Fig.~\ref{fig:math_problem2} for example, the abstraction and specialization processes as such. Another example is differentiation and integration. Differentiation is taking an abstract function and calculate its instant tangent in a specific point, which is simply to do. This is abstract-to-special process. On the other hand, there is integration, the opposite process of differentiation. It is taking abstract function and produce abstract function, that calculates over unspecified range, unlike neighboring range of differentiation. This is in general hard or impossible to do. 

\begin{figure}[H]
	\centering
	\includegraphics[width=0.99\textwidth]{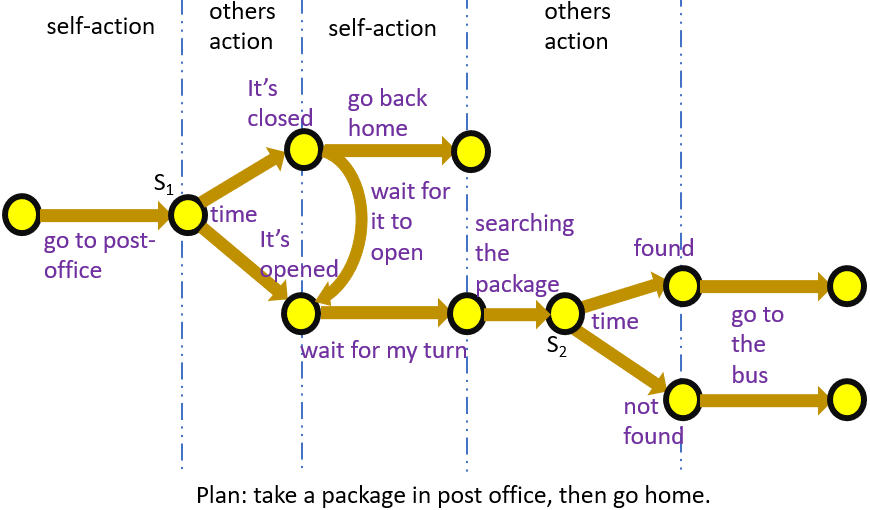}
	\caption{Uncertainty of actions.}
	\label{fig:abstract_planning}
\end{figure}

Another example of such abstract planning is presented in Fig.~\ref{fig:abstract_planning}.
Here we show that not all state transitions are deterministic. Many of them are stochastic even for the agent’s actions, and many others are stochastic due to dependence on other factors beside the agent. Like in any game, for example chess (other agents).
Stochasticity can be expressed for example via abstract states, having un-assigned class variables, i.e. unknown. For example, $S_1$ state in Fig.~\ref{fig:abstract_planning} contains a \textit{Post-Office} class which have un-assigned feature of either being open or closed, while $S_2$ contains a \textit{package} class which have un-assigned feature of either being found or not. See similar discussion about pre-assigned classes in \ref{sec:Operationability}.

In order not to make this planning too stochastic, we try to consider only most probable possibilities, and disregard planning the response to any of the possibilities.

\subsubsection{Complex example}

Next example is a complex one, hence it will be described lengthily. It is a mixed example, because it involves both the part of story comprehension, or understanding the problem (initial and goal states), and the planning part to solve the given problem.

\begin{figure}[H]%
	\centering
	\subfigure[]{%
		\label{fig:a}%
		\includegraphics[width=0.97\textwidth]{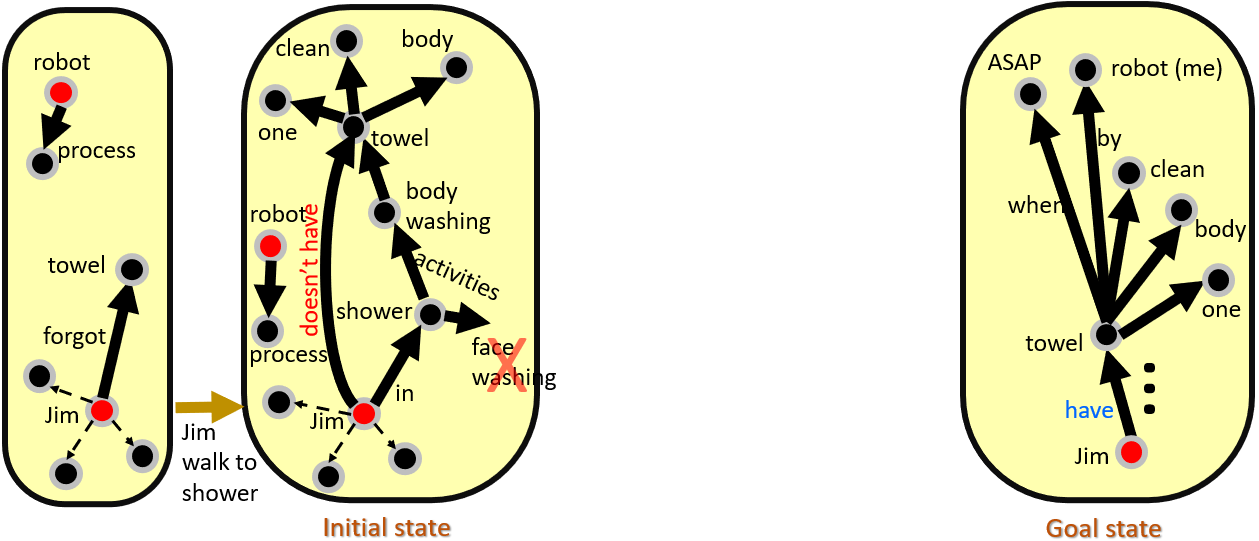}}%
	\hfill
	\subfigure[]{%
		\label{fig:b}%
		\includegraphics[width=0.97\textwidth]{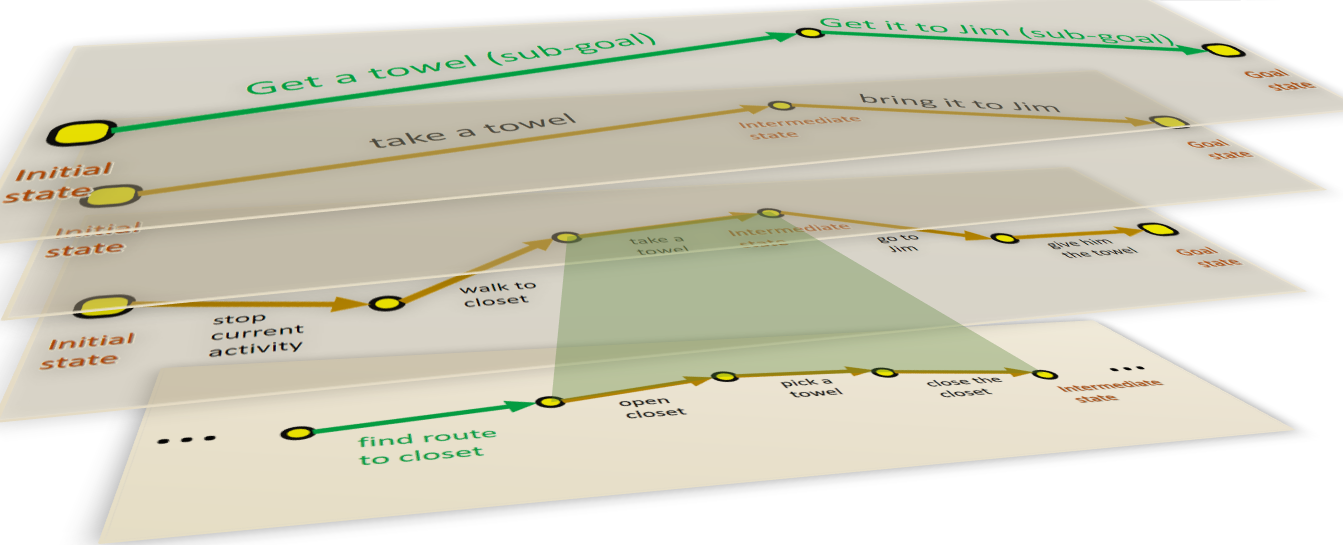}}%
	\caption{Mixed Example: Understanding and Planning.}
	\label{fig:mixed_example}
\end{figure}

Fig.~\ref{fig:mixed_example} is the result of the following description, revealed as an incoming stream to the robot: \textit{"Jim speaking to his robot: I’m in a shower. Forgot to take a towel. Can you bring me one?".}

Next, we develop our discussion about the example in the following sequence: description of prior background knowledge and settings prior to the event, then the process of perception and inference (Fig.~\ref{fig:mixed_example}(a)), and finally the process of planning (Fig.~\ref{fig:mixed_example}(b)).

There is naturally some pre-settings of this request: there is a serving robot, Jim is a human and he is its supervisor. When Jim talking to the robot, it needs to perform what Jim asks. Currently they are both in Jim's apartment.

Some background common sense is assumed to be learned by the robot. Such as facing uncertainty while receiving a request. Other constraints include: ethics - do not break stuff (e.g. wall) to get something, nor to ride over people or animals on the way; actions/states that are forbidden/required (e.g. embedded in cost feature); and general cognitive memory and time (resource) constraints: e.g. total number of steps or total time constraint (or order of it).

We can see the simultaneous inference process in the states. For example, in the initial state, face washing denial and towel's feature are not given in the input stream, but inferred. 
Also, note that the actual representation of this example is larger. Due to space limits, many details were omitted. For example, we do not show that “face washing” activity is a wrong branch, even though it requires also a towel. Since he did not specified this, the robot can either ask for confirmation/clarification, or assume which of the tasks is in more demand. Body washing means Jim is nude, and it requires from him to get dressed to find the towel himself, while face washing is less of a problem. Hence, body washing is the more challenging activity to retrieve the towel himself, hence Jim is in need for more assistance in this case.

As can be seen, the states presented in the figure are the act of understanding the problem, or the story so far, until the robot realizes the request from the story. This request is what turn this story as incomplete, which requires some resolution/completion to fill the missing steps of the story in the state space. This description is typical to most of our communication, since we regard the incoming messages as some mystery to be figured out, either by inner inference or outer inference. All for the purpose of constructing a legitimate full story.
Next we describe these two phases.

The process of state generation is as follows (Fig.~\ref{fig:mixed_example}(a)). It first tries to comprehend the situation: shower, towel. Next it tries to construct the goal state: by=robot, where=to Jim’s hand, when=as soon as possible (since shower is a matter of minutes).

We can also see, that state generation is not only performed to the future, but in some cases it can also be into the past, and not only that but it could be also not in consecutive steps, but in some skips, such as the skip from the initial state to the provided goal state. Hence, more generally, a story can be completed by different time skips, for example forward/backward constructing, or some random skips to different time lines or space-time settings.

Next, is the planning process (Fig.~\ref{fig:mixed_example}(b)), or solving the problem, or the “how”. It should start with the robot being stopping the current activity, or putting it on hold, or giving it a few moments if it is more important (these considerations are also important). Next is to start the new activity, of reaching the new goal. When: should start now. Where to go: to where towels that Jim wants are (these are constraints in the problem). How to get there: walk to where they are, take it and go to Jim. This is high-level model. Low-level is the walking procedure, velocity according to time constraint, while safety is assumed during walking. 

Note that we described the problem-solving as the "how" question. Relating to inference guiding basic wills, see \ref{sec:basic_wills}, we can claim that guiding will can also appear in the macro, for example as the type of will connecting the initial state to the goal state.

The planning process starts from general actions or wills, and descent to levels with more details, in order to make it an actual feasible plan. Or to reach as most detailed solution as possible, see more about it in Fig~\ref{fig:will_in_constraint_environment}(a). During descending the actions/wills are decomposed into smaller units for that purpose. For example, see the "take a towel" action in the 2nd level from the bottom decomposed into "open closet"$\rightarrow$"pick a towel"$\rightarrow$"close the closet" sequence of actions, in Fig.~\ref{fig:mixed_example}(b). Additionally, as can be seen, actions and wills are interchangeable. Sometimes we call it a will and sometimes it is an action.

As mentioned previously, in Fig.~\ref{fig:cognition}, planning is a top-down process, while understanding is the opposite. We can demonstrate it by using the same 
Fig.~\ref{fig:mixed_example}(b), only instead of looking at it as constructed top to bottom, we imagine it as constructed bottom-up. It can be done simply by telling what the robot did retrospectively: \textit{"Jim and its robot were home. Jim were taking a shower. The robot stopped what it was doing. It walked to a closet. It took a towel from the closet. It then went to Jim. And then it gave Jim the towel".}
In this case the process involve abstraction/grouping to aggregate the detailed actions into more general ones. For example, see the "open closet"$\rightarrow$"pick a towel"$\rightarrow$"close the closet" sequence of actions in the lowest level aggregated into "take a towel" action in the level above, in Fig.~\ref{fig:mixed_example}(b). Finally we (or any listener) reconstruct the original will from the detailed actions: \textit{"Bring a towel to Jim}.

Also note that at each state generation, weather it is planning or understanding, there is also local inference processes, which can be regarded as internal will.
As opposed to the external will, which we try to reconstruct from the story.
See more in Fig~\ref{fig:will_effect}.

Also, important to note is the surprising comfort of actions to be composed or decomposed, since the grouping is represented as simple sequence of actions, while in general the grouping of actions could have any form. This is not some lucky incident. It is appeared to be basic principle in programming, see the beginning of section~\ref{sec:FP} and its footnote.

Finally, this and other examples demonstrate that any successful solution is memorized for future use, in its full hierarchy, since any actions and sub-actions can be used as tools in future problems.

\subsubsection{More special examples} \label{sec:special_examples}

\begin{figure}[H]
	\centering
	\includegraphics[width=0.95\textwidth]{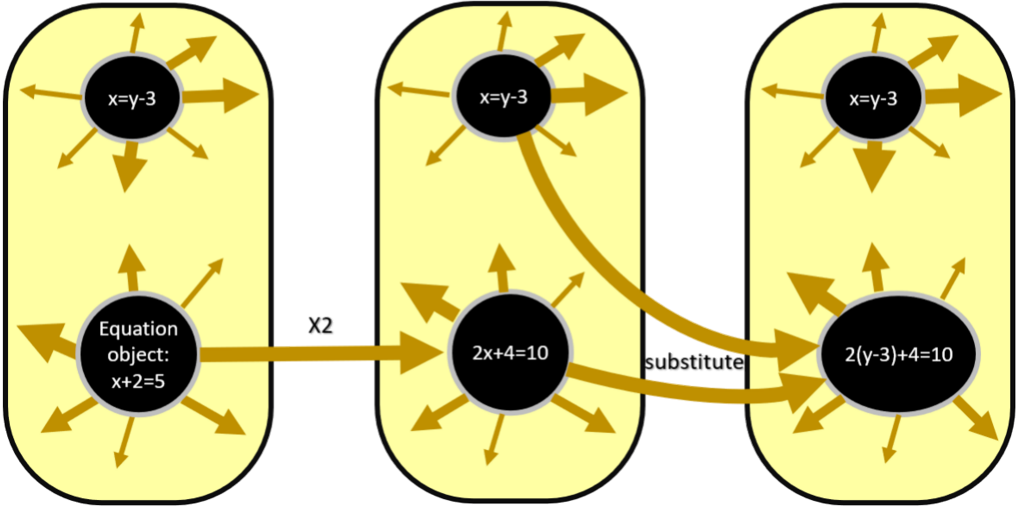}
	\caption{Developing equations example of actions applied on more than 1 object.}
	\label{fig:developing_equations}
\end{figure}

Here we discuss some interesting special cases.
First we show how actions, similar to functions, can operate on several arguments, or in our case on several objects. See the example of developing equations in Fig.~\ref{fig:developing_equations}.

\begin{figure}[H]
	\centering
	\includegraphics[width=0.99\textwidth]{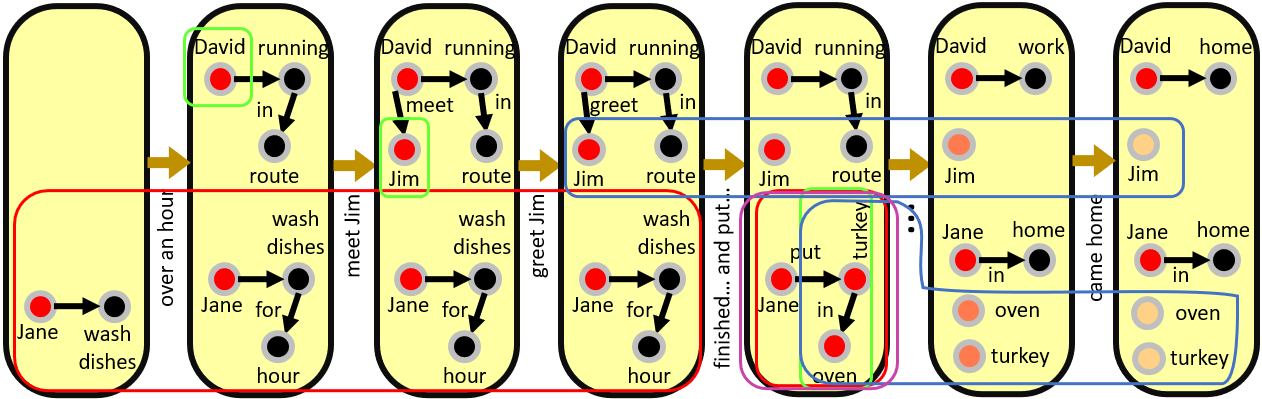}
	\caption{Example story with various effects in it.}
	\label{fig:varios_effects}
\end{figure}

In the next example, shown in Fig.~\ref{fig:varios_effects}, we can see various effects in a story. The story is: \textit{"David was running, while Jane has been doing the dishes for over an hour. In his route he met Jim and greet him. In the meantime, Jane finished to wash the dishes and put a turkey in the oven. ... David came back home from work to Jane."}

First, we see a process (changing object’s state after a while) in contrast to an instant type of action, in red color: the dish washing, which disappears in the presented next state (also in red).

Next, we see how objects are added over the states (in green) or removed due to to the pass of time or forgetfulness (in blue).

And again we can see actions that are applied on several objects (in purple).

\begin{figure}[H]%
	\centering
	\subfigure[An inanimate object]{%
		\label{fig:a}%
		\includegraphics[width=0.49\textwidth]{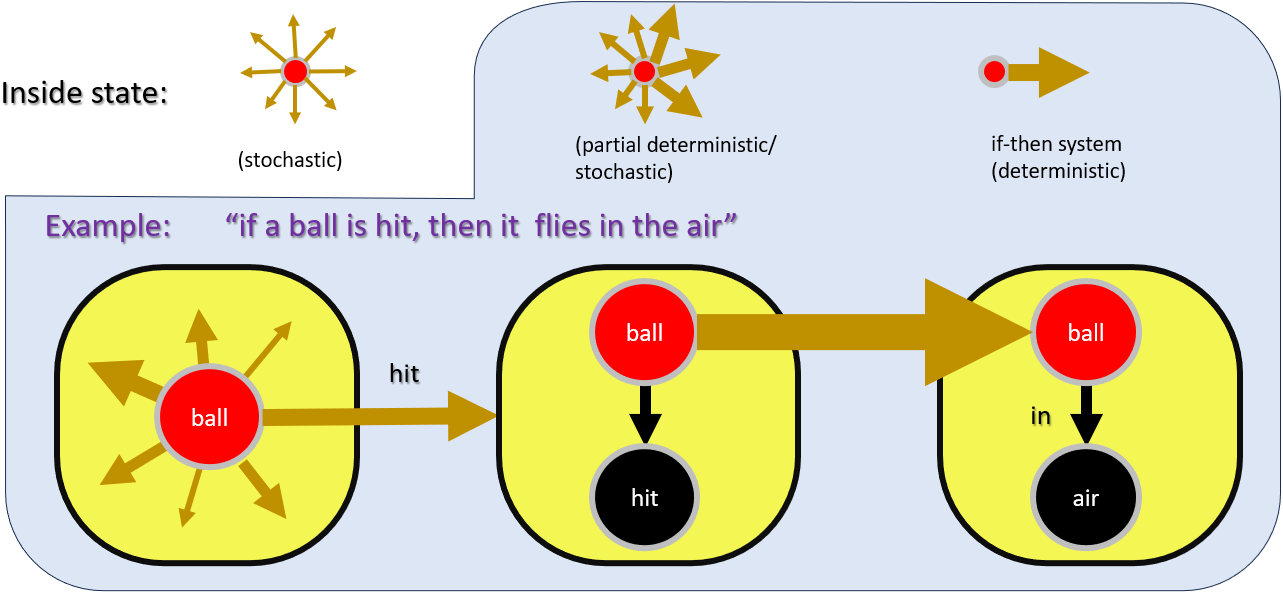}}%
	\hfill
	\subfigure[sub-symbolic to symbolic]{%
		\label{fig:b}%
		\includegraphics[width=0.49\textwidth]{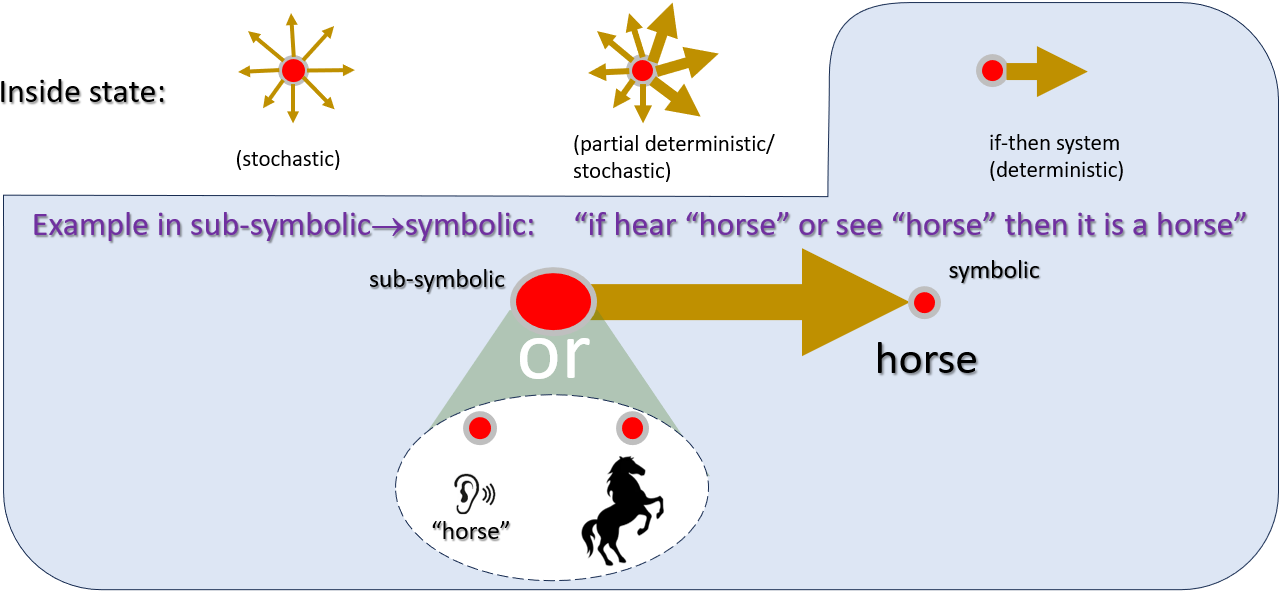}}%
	\hfill
	\subfigure[Multi-modal perception]{%
		\label{fig:c}%
		\includegraphics[width=0.49\textwidth]{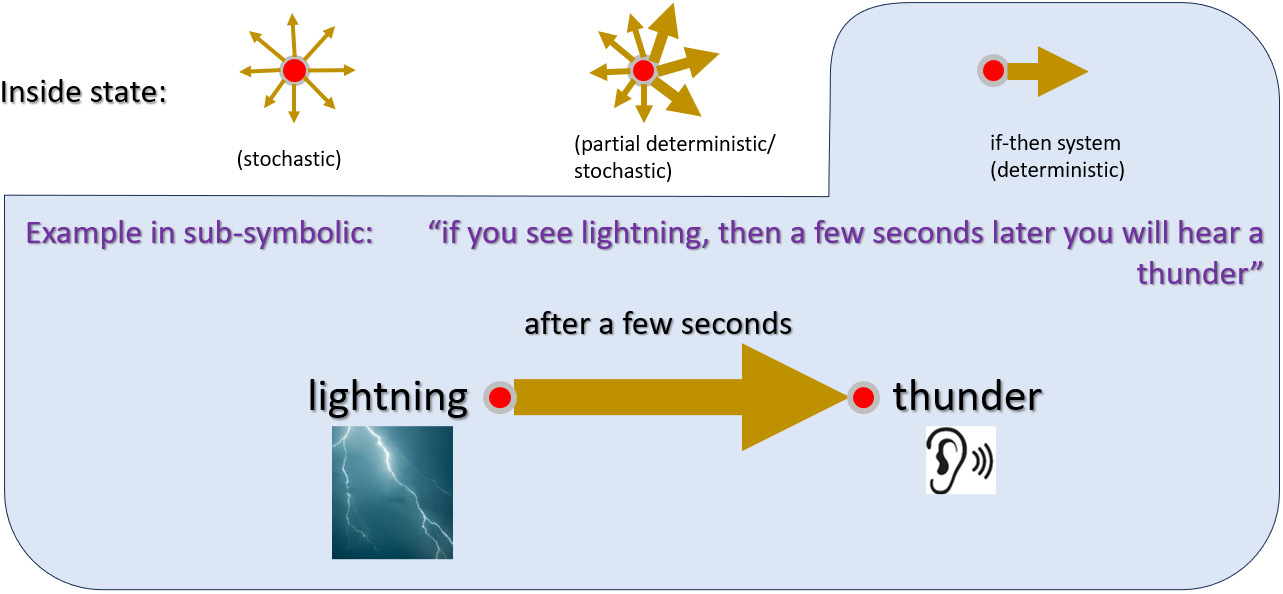}}%
	\hfill
	\subfigure[Human's free will]{%
		\label{fig:d}%
		\includegraphics[width=0.49\textwidth]{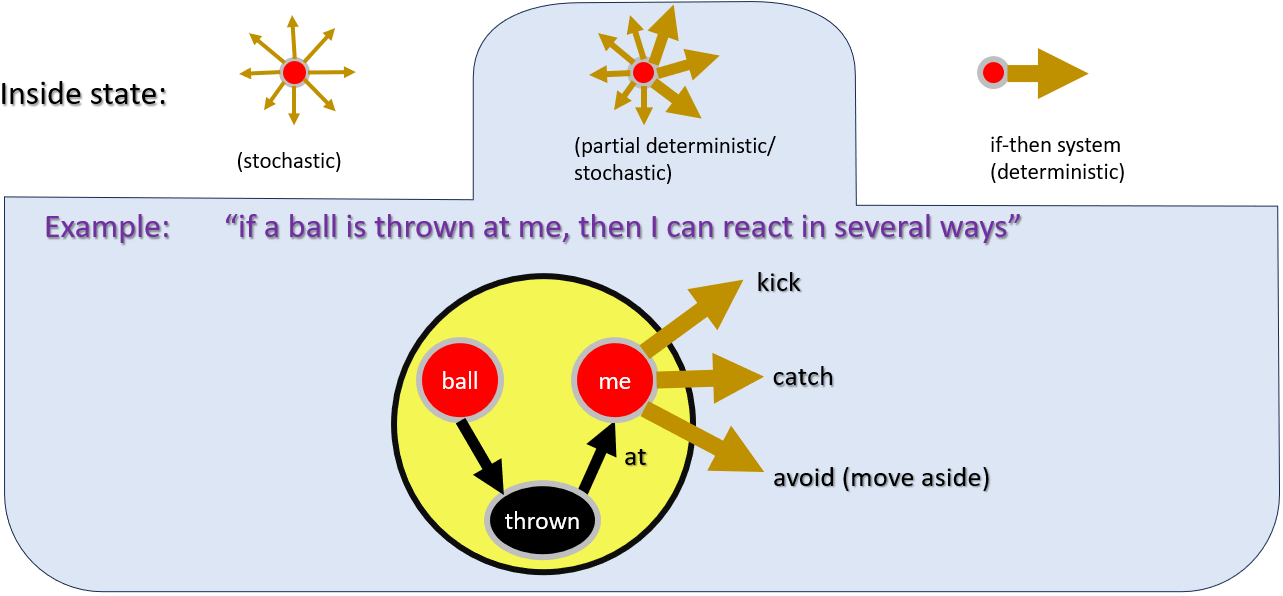}}%
	\hfill
	\subfigure[Pavlov’s experiment]{%
		\label{fig:e}%
		\includegraphics[width=0.87\textwidth]{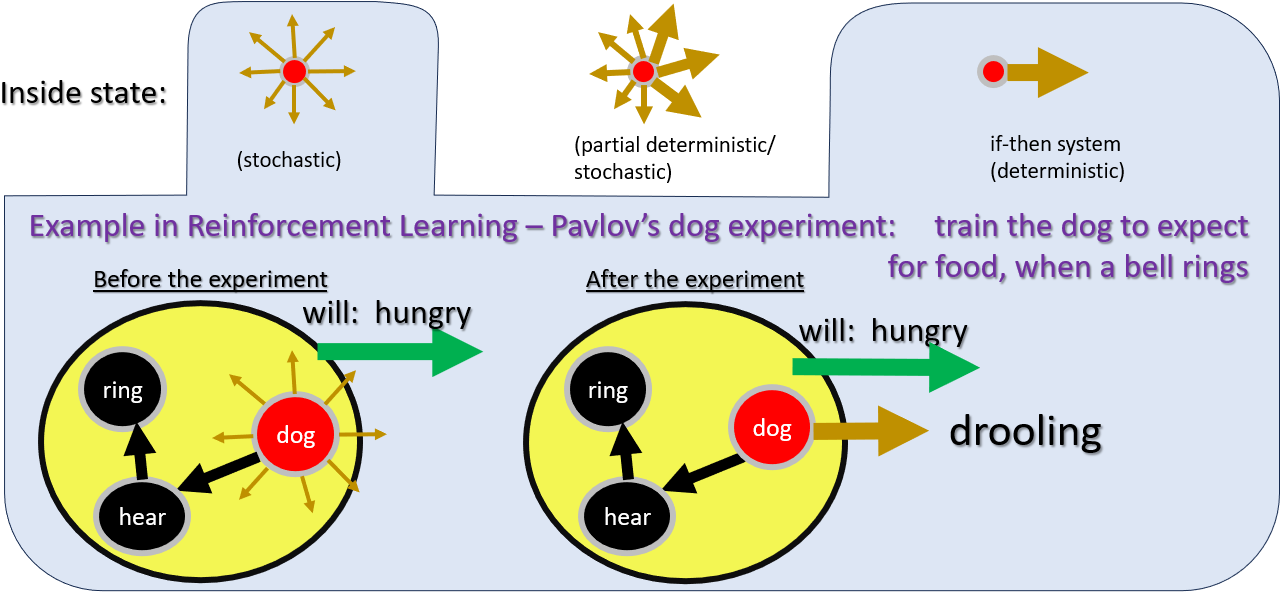}}%
	\caption{Examples of single/multiple admissible actions.}
	\label{fig:Will_derivation_examples}
\end{figure}

The next example demonstrates the principles described in \ref{sec:Will_derivation}. Specifically, in the following are some examples of the three forms of single/multiple admissible actions. See Fig.~\ref{fig:Will_derivation_examples}

Fig.~\ref{fig:Will_derivation_examples}(a) is about an inanimate object. Fig.~\ref{fig:Will_derivation_examples}(b) is about sensory perception transformed into our knowledge map.
Fig.~\ref{fig:Will_derivation_examples}(c) is about sensory perception, specifically from different modalities.
Fig.~\ref{fig:Will_derivation_examples}(d) illustrates human’s different options to act or react.
Fig.~\ref{fig:Will_derivation_examples}(e) presents a psychological experiment, that tried to teach a dog to associate a hearing ring from a bell, to being food served. Hence it is a reinforcement type of learning.
We see that before the experiment the dog has no special reaction to a ring (stochastic reaction), but after enough training we see he is developed a dominant response in the form of drooling (deterministic reaction), which can be directly translated to its will of expecting a food to be delivered.

Finally, we see here different types of interactions (sensory inputs and physical acts) with the environment. And we see that most of the interactions in the figure can be aggregated into a group object, e.g. event object, with the appropriate actions/features. For example, "hit ball" object with the action/feature of being in the air, "(audio) horse or (visual) horse" object with the feature of "horse" class in knowledge map, "a ball thrown at me" object (and its admissible actions), and a "ring-hearing dog" object with the drooling action associated to it.



\subsubsection{Another complex example}

\begin{figure}[H]
	\centering
	\includegraphics[width=0.95\textwidth]{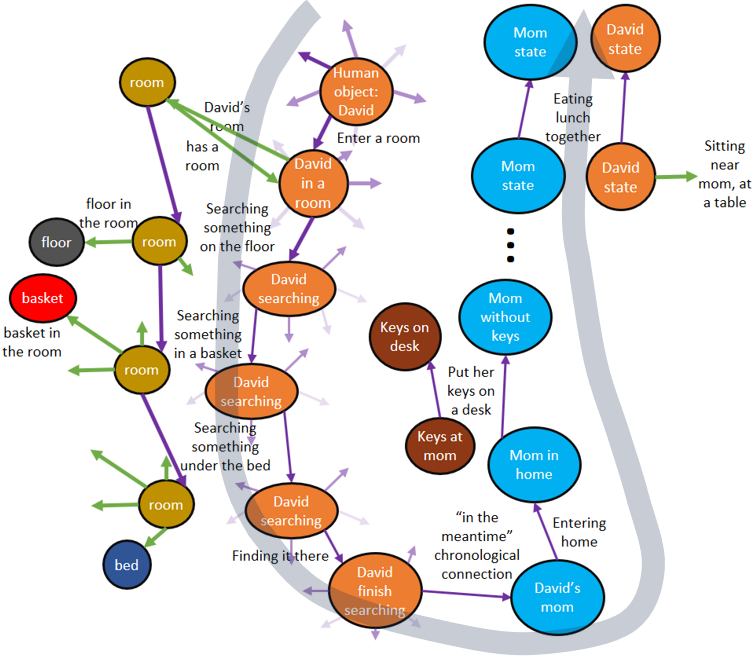}
	\caption{A part of a story.}
	\label{fig:examples_of_cognition_processes}
\end{figure}

In this example we will see how multiple wills can co-exists in a story.
In Fig.~\ref{fig:examples_of_cognition_processes}, for simplicity, actions denoted as arrows, usually perform changes in several objects to their different attributes (as mentioned in previous examples). Here high admissibility is expressed via salient color (while directionality as usually relates to will). 
Fig.~\ref{fig:examples_of_cognition_processes} is a \textit{MOM} representation of evolving state, representing a story. A state is consisting of several objects (joining/leaving as the story evolves), including their attributes and actions that define the state.

\textbf{The story}:
\textit{"David entered his room. He searched for something on the floor. Then he searched in the basket. Then he searched under his bed, and was thrilled to find the ball there. In the meanwhile, his mother entered home. She put her keys on the desk. Then she removed her shoes and put her sunglasses on the desk. Then she searched for David, found him, and they sat to eat lunch together"}.

Note how hierarchy is mostly around abstraction, and slightly about will. For example, in the story we could switch the focus in the middle of non-completed will, e.g. "while searching for something, mom entered the house". This example also demonstrates different levels of will. From the first sentence we understand the lowest form of will - he "wants" to move in the room. Later a higher will is introduced: searching for something. Only at the end a higher will is presented: he wanted his ball. There still could be a higher will serving these sub-wills, e.g. he wanted a ball to play with his best friend, whom he enjoys so much.
So at first it can be a sequence of actions connected by time, and as the story evolves, will is revealed, and these actions mapped in a clear field of wills, within different levels, similar to problem-solving scenario: problem state and goal state.

The same story is represented via \textit{AKREM}, without the evolution of details, see the video link in \citep{10.1007/978-3-031-19907-3_6}. 
Note, that if repeated often, the sequence of David or anybody searching in some place can be grouped/abstracted as an event class, also referred to as \textit{Trans-Frame} \citep{minsky1988society}.

Additionally, it is seen that this is a hyper-graph, since the actions, represented as edges in it, connect multiple sources to multiple targets. That is, actions, though connected to all objects by relevancy, can actually act mutually on several objects at once.
This is also supported in neuroscience, e.g. in a modified Hopfield network \citep{burns2023simplicial}, where set-wise connections instead of pairwise connections between neurons is proposed.
Altogether, the edges and the nodes (as mentioned previously they can be grouped to represent new group node) imply a representation of hyper-graph.

\begin{figure}[H]
	\centering
	\includegraphics[width=0.45\textwidth]{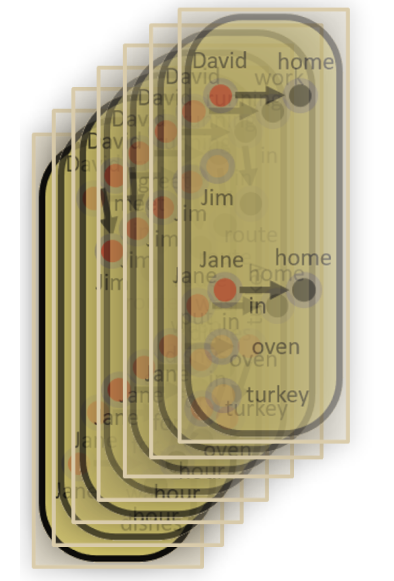}
	\caption{Dynamic knowledge representation of the story in Fig.~\ref{fig:varios_effects}.}
	\label{fig:dynamic_state_frames}
\end{figure}

Note that \textit{AKREM}'s examples, and the presented example imply a dynamic knowledge representation, i.e. watching the full knowledge state as it evolves in time, while a story is perceived. It is like frames in a movie, stacked one over the other. See Fig.~\ref{fig:dynamic_state_frames}. 

However, in all illustrations so far, we separated the instantaneous state from its evolution (state-by-state static representation), to emphasize the will main effect on how this state evolve.

Nevertheless, it spurs the discussion of which of the above representations is more appropriate. On the one hand, \textit{MOM}'s state evolution representation is supposedly virtual, and should not be grasped/accessed by the agent. On the other hand, this representation is presented in agent's mind, while planning, imagining, simulating, etc.

In memory, we suppose that there are two separated but interconnected memories: the usual LTM, and some uploaded items from it in the WM. This may explain the capacity limit in WM. Since if the planning are sequences of changed state copies, then there is a limit for how much can be held in terms of states and sequences of them. For example, imagine the size of a memory needed for a sequence of states in an evolving story, which includes multiple objects and agents.

Thus, to make the memory more efficient, we can use the dynamic representation, and at each "frame" store only the changes from the previous state. Also it supports our efficient representation of uploaded memories into WM, as being generated directly from triggered objects in LTM. Meaning, we maintain layered representation which is founded upon LTM knowledge map.

Likewise, grouping, as discussed in \textit{AKREM}, might just be for the purpose of reducing the amount of elements attended at each time instance.

\section{\textit{MOM} evolution}  \label{sec:MOM_evolution}

This section deals with how \textit{MOM} is evolve and develop through time, from "birth" of a human/agent till it reaches its adult phase, or mature intelligence state. A state we consider as stable. 

We do not consider this state as the goal state of cognition, since obviously the agent/human can still evolve and learn new things. However, we sometimes refer to it as goal intelligence state, merely to emphasize that this is the desired outcome of a thinking machine or an agent. A stage of its evolution that it is fully capable to assist and drive progress in different scientific faculties. It can do so by solving problems or designing new and creative artifacts.

Finally, this section concentrates mainly on the basic principle of consolidation in the cognitive evolution, and then it elaborates about the learning approaches that use this consolidation tool in their process.

\subsection{Consolidation}   \label{sec:consolidation}

So far, the \textit{MOM} cognitive model has been presented in its mature state. Now, the discussion is about how to reach it. This is a process in time, which is mainly based on consolidation. A more general view of cognitive evolution in \textit{AKREM} is discussed in Appendix~\ref{sec:AKREM_evolution}.

Consolidation is about transforming from chaos to some stable order of patterns, or from a continuous realm to a discrete one, as in quantum mechanics. An infinite amount of details is hard to handle (i.e. to understand and then to control), therefore consolidation to fewer patterns is required.
Consolidation also allows for fuzzy logic and categories \citep{wyler1995fuzzy}. 

It also allow for effective and minimal communication, with short codes, where there are few shared symbols.

Consolidation can be expressed in many forms, such as:
\begin{itemize}
	\item in the conversion of sub-symbolic to symbolic, for any type of \textit{element}
	\item in cognitive evolution: from flexible (at infancy) to less flexible (at adulthood)
	\item in problem solving: going back and forth between divergent thinking (open up for multiple solutions) to convergent thinking (consolidating into one or less potential solutions)
	\item in modeling, at program search, from huge hypothesis space for possible programs to a small set of hypotheses (as in \textit{DL}). It is both in the micro (within models) and in the macro (between models)
	\item in testing multiple versions of an unknown model, and finally converging into less/one version(s) that are/is consistent with evidence
	\item and in grouping/abstraction, where some separate elements become connected
\end{itemize}

Note that causality is a special case of modeling, a spatio-temporal one, where re-occurrence is consolidated. More generally, re-occurrence helps in learning both static objects and dynamic basic/composite events (equivalent to scenarios/scripts in \textit{OOP}). 
See more about this in Section~\ref{sec:knowledge_structure}.

Additionally, \textit{MOM} enables multiple parallel versions of the same thing, since any specific topic or subject can have multiple theories/models, sometimes in conflict.
Hence, like in the quantum superposition realm, multiple-version combinations could be tryout, and consolidation can help in collapsing them into fewer 
versions. Those versions should make the most sense, i.e. to be consistent on different occasions or supporting the majority of evidence.
Thus, it just maybe, that at infancy, a highly uncertain period, there are many versions created, and with time - only the most consistent ones survive (consolidate).\footnote{Note, that it has nothing to do with having multiple versions representing different perspectives or opinions, over specific controversial model/topic. In this case all versions are necessary, and not reduced with time. Similarly when multiple versions of a concept (with a same name), having multiple interpretations, depending on context, due to the fact that meaning derived from will. And since there are different wills for the same objects - different meanings/interpretations emerge}.

Interestingly, consolidation is about prioritization of models, i.e., it is about seniority. In other words, the more "proven" models are those that "call the shots", e.g. it is required from new untested models to be consistent with these old models. This prioritization is what allows for symmetry-breaking, i.e., the ability to go from totally symmetric/equal model proposals and converge slowly into a more confident and stable models.

Moreover, multi-version principle aids and fit also to the will field. As we discussed previously, it is dynamic resetting of the knowledge representation to fit the current will. In other words, the agent learns how according to a specific will, it should produce the relevant objects and actions to accomplish this will. It is similar to the idea of problem representation, where a problem is half-solved if we succeed in representing it in its most natural and efficient way. This fits perfectly with having multiple versions of "world representations", where the will acts as a switch to select among those. This is yet another multi-versionality, hidden in the additional degree-of-freedom: the will.
Similarly, settings or context, see \ref{sec:Modeling}, can determine the representation of knowledge.
Similar idea is advocated in \citep{schaffner2023sensory} where perception is preferred to have goal-oriented representation over "seen as it is" modeling. This is also embraced by attention mechanism, where the sensory information is always overwhelmed, hence must be filtered and handled efficiently.

Lastly, two operations help in producing consolidation.
On the one hand, to deal with a stochastic environment and ambiguous signals, \textbf{repetition} provides memory prioritized by relevancy. Hence the need for associative recalls, i.e. cues. Repetition is never exactly over the same thing, but rather over many different examples of a thing. Hence, it is useful also for generalization or abstraction out of diverse and noisy/blur/corrupted/partial examples. Therefore, the strict/rigid approach (e.g. symbolic or logic) is not suitable for learning in a real-world or world-simulated environment, but only in idealized well-constructed one.
Repetition is needed also with guided tutoring of an \textit{AGI} agent. 
Conversely, \textbf{sparsification} is about reducing irrelevant signals. It also support Occam's Razor principle, i.e. searching for the simplest algorithm among all possibilities.

In summary, these operations are also act as dualities.

\subsubsection{Reusability}

An additional form of consolidation is reusability \citep{chollet2019measure}, since the more learning progresses, the fewer new models are proposed in favor of using existing ones.
Hence, reusability is expressed via exploration (mostly at early stages) verse exploitation (in stable or mature stages), as in \textit{RL}.
In the beginning, many possible codes are generated for actions/models, but as time goes by, the process is less exploratory and more exploitative, i.e., there is more emphasis on retrieving known codes, while testing fewer new codes in parallel.
Similar effect was seen in learning abilities among children verse adults \citep{frank2022efficient}. 
This effect states that children can gain new memories easily, while adults exhibits interference with old memories in the process of gaining new memories, hence an inhibition process is stronger at adults.

In addition, reusability aligns perfectly with abstraction/grouping, in a constrained environment and limited resources. They are both needed to hold control of as much as possible, with minimum effort \citep{zipf2016human}, 
i.e. without generating many models of each thing.

In practice, reusability is about using less of the initial available tools, as the learning evolves.
Meaning, while regular \textit{DL} tools (if, sum,  
activation function) or the primitive tools, see Section~\ref{sec:Operationability}, can be used for program search of basic action methods or relation methods, the new methods apply reusability. In such methods, less primitive tools are used while the current methods are used more, thus encouraging more connectivity in the network, since the more existing methods are used, the more are associations are connected to them, thus avoiding isolation of new methods by inner program search.
It also encourages logical connectivity in the model, since \textit{DL}'s non-interpretable connectivity is replaced with more logical functions, as in algorithms.

\begin{figure}[H]
	\centering
	\includegraphics[width=0.85
	\textwidth]{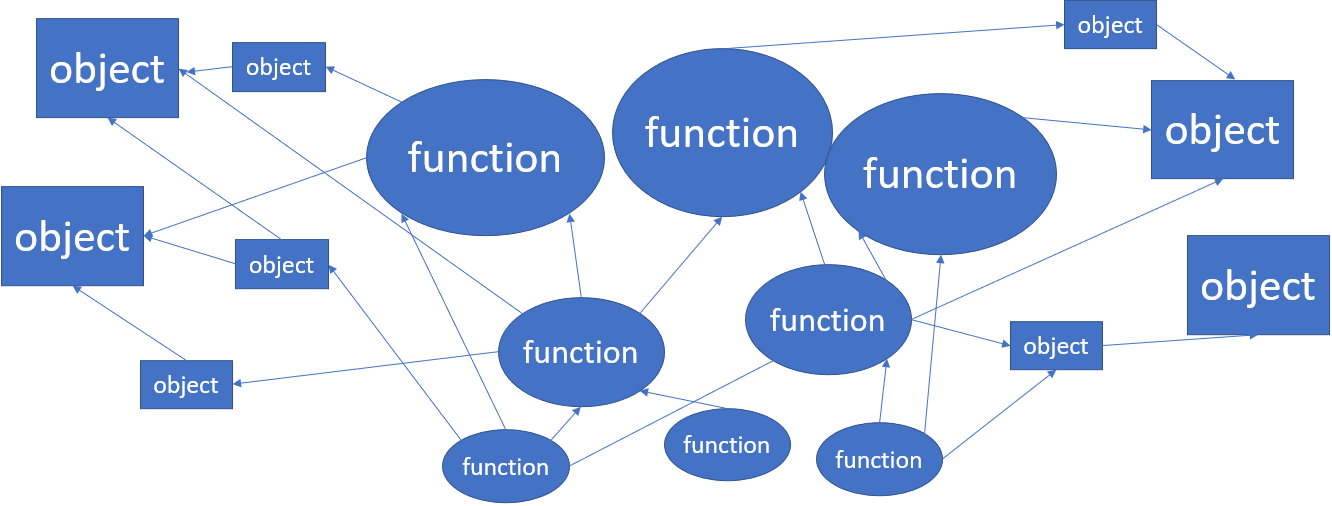}
	\caption{Actions, objects, compositionality, and reusability}
	\label{fig:COMPOSITIONALLITY}
\end{figure}

Moreover, Functional Programming can be applied to assist reusability. On the one hand, the general/outer structure is \textit{OOP}, i.e. \textit{elements} are grouped in an \textit{OOP} fashion. On the other hand, methods are kept in a pure operational immutable form \citep{Chen,van2009programming}. Meaning, having small and simple methods, which maximally reuse other functions, and without inner variables, due to objects-memory-only assumption (see Section~\ref{sec:Modeling}). Meaning, striving for methods that are comprised of other methods, as much as possible. This is compositionality/grouping applied in actions (similar to hierarchy of visual features in CNNs). 
See summary of reusability in Fig.~\ref{fig:COMPOSITIONALLITY}.

Final remark for this chapter: we encourage curriculum incremental learning over batch learning or even meta-learning, since our belief is that learning is gradual, in stages. Meaning, we cannot use shortcut to learning, by squeezing everything for an agent to learn. It must gradual process, most probably with failures and regressions.
Additionally, we advocate slower learning scales (days, months), with gradual improvements, in contrast to fast learning (hours, days) with final model as in \textit{DL}. Similarly, we are for larger and more dominant stage of exploration verse smaller stage of exploitation. It is because we humans use unguided or self-learning, just as supported by many important AI researchers, similarly we should allow for the AGI agent a long period of experimentation and exploration, before it converges into a solid/stiff world/knowledge model.
See more motivation in Appendix~\ref{sec:wake_sleep}.
Finally, \textit{DL} is very problematic with regards to data. Because of the phenomena of garbage-in garbage-out, the data scientist must clean carefully, tediously and exhaustively the data from outliers, duplicates, ambiguity and other factors. This both narrows the task, and do not let the model to learn exceptional cases, and it reduces the agent autonomy since the scientist is the one prepares him the data. Instead it should be able independently explore the environment and to cope with the data it perceives along with the ability to choose/control the data it attends or filters.
This factor should be also part of the AGI goal to organize its own gathered data in a consistent, comprehensive and usable way.

Also, it is for effective learning, e.g. knowing when to disregard information taught to you. An information that is way over your understanding. For example, learning a 12th grade material while you are in a 2nd grade level.


\subsection{Consolidation and language}

If we think over the idea in Fig.~\ref{fig:MOM_hierarchy}, we can derive to a broader theory about language as unique tool of humans. Think about all the animals or even pre-historic human races, that use very simple and small sets of symbols to communicate. We can imagine it as the perception module in Fig.~\ref{fig:MOM_hierarchy}, while the modeling module is very primitive and the top-down processing is merely for purposes of dangers or food gathering. This makes the models from top very simple and almost non-symbolic. I.e. it is like a "jungle" of neurons. Such humans or animals have very limited symbolic capacity both to communicate and think, or in other words to express themselves.

It means that attaching more words, in our opinion, makes this "jungle" of neurons, more and more ordered into clusters of concepts, via consolidation.

\subsection{Learning the modeling} \label{sec:Learning_the_modeling}

Here, model learning mechanism is proposed, where two contrary but completing learning approaches in AI are combined \citep{day2022knowledge}: empirical, i.e. from examples (induction), and expertise (rule-based). This is a learnable symbolic manipulation, or can also be referred to as a hybrid approach or Neuro-Symbolic, see \citep{marcus2003algebraic}.
Empirical is bottom-up (from examples to rules), e.g. via observation or passive interaction. Rule-based is straight from/in the top, via rules in abstract language, e.g. via conversation or observation. It can then descend to examples of these rules (deduction). Examples applying these approaches can be seen in Appendix~\ref{sec:Logic}. 
Similar ideas presented in neuroscience, as statistical learning and rule-based learning \citep{takacs2023neurophysiological}.

\begin{figure}[H]%
	\centering
	\subfigure[Bottom-Up]{%
		\label{fig:a}%
		\includegraphics[width=0.45\textwidth]{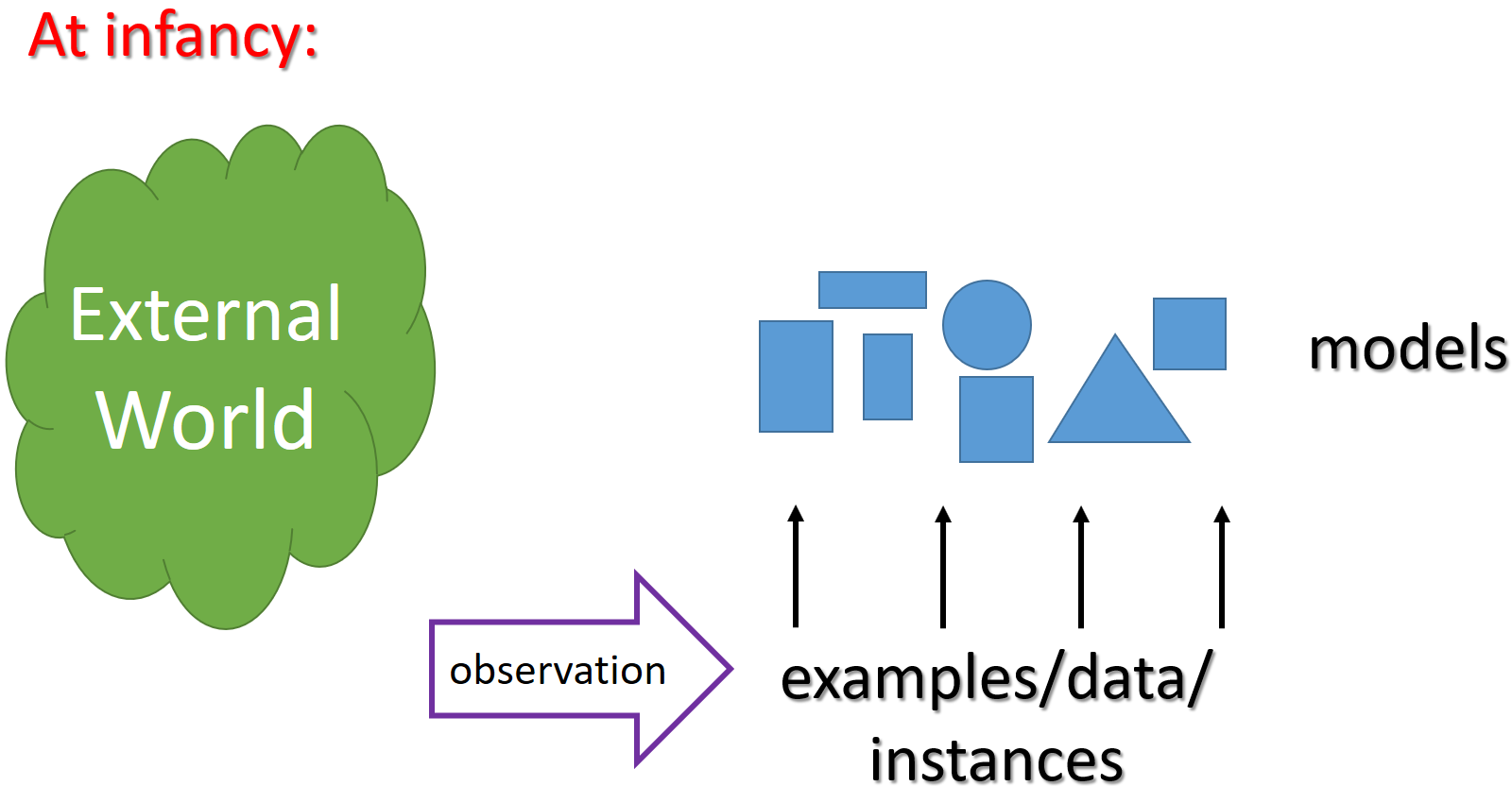}}%
	\hfill
	\subfigure[Top-Down]{%
		\label{fig:b}%
		\includegraphics[width=0.45\textwidth]{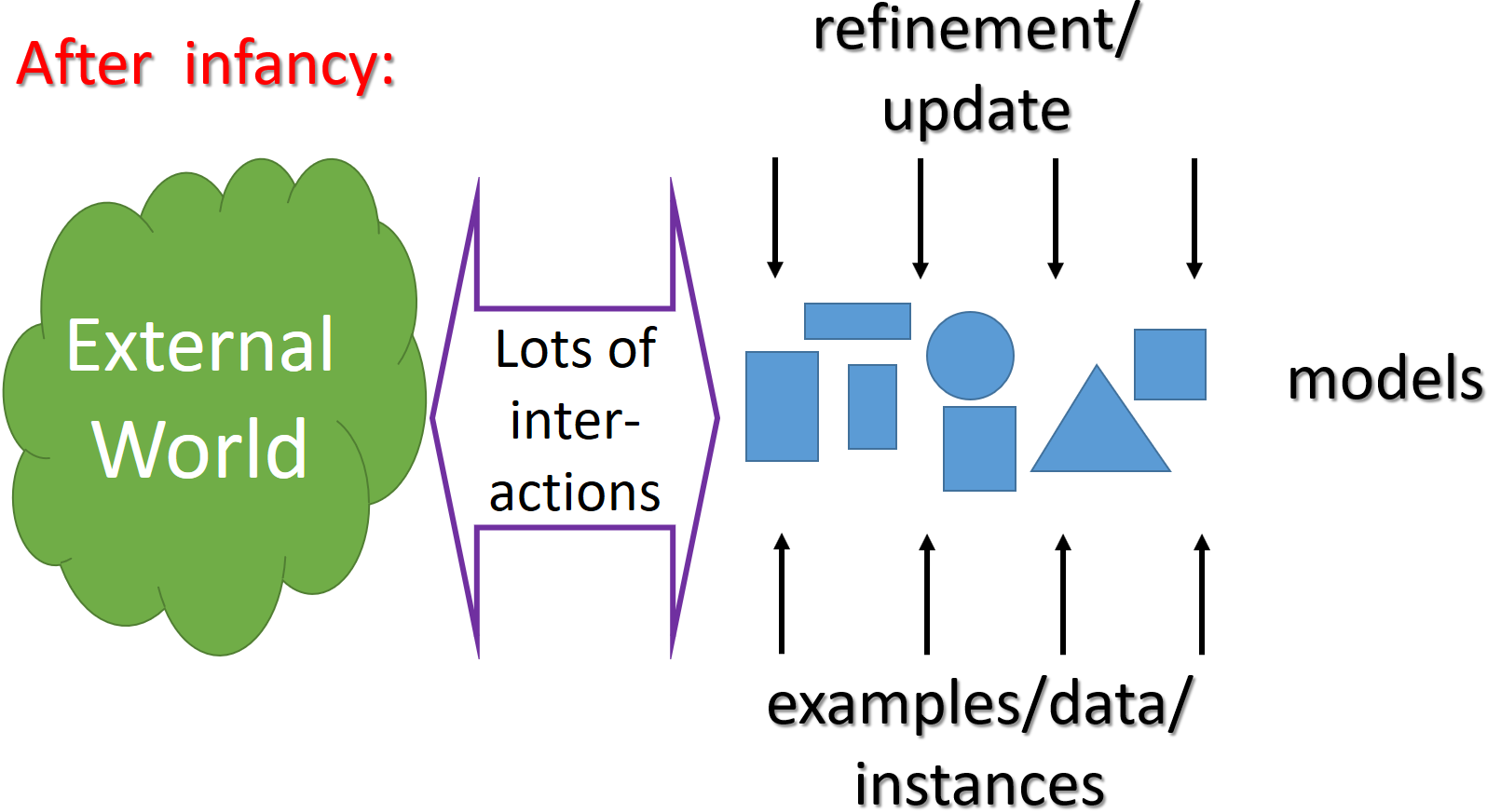}}%
	\caption{Bottom-Up and Top-down methods to learn models}
	\label{fig:top-down_and_bottom_up}
\end{figure}

These approaches can explain the development process of an infant.
We can assume that at infancy, the infant starts with observation (visual), then fuses visual objects with audio simultaneous stimuli, also via observations. This is the bottom-up learning process.
Physical interaction could be exercised too, but we will assume it is insignificant.
After infancy, he can use language to update and refine his preliminary and raw models, as it is done in childhood, and at school. This is the top(-down) learning process.
So we could say that just like gradual involvement of senses, so is the gradual involvement of interactions.
The idea of how models are developed is illustrated in Figure~\ref{fig:top-down_and_bottom_up}.

This explains the first step of infant's learning, i.e. observing or imitating, since he is yet to know the proper response to his perception. The first bottom-up step enables him the learning of models, and also the learning of implicit and explicit intentions embedded in the observed behaviors. Which can be the basis for his own intentions in the future, especially in conversation and other interactions. I.e. it might be very possible that different types of will or intention are first learned.

For example, in \citep{kimbeyond}
they talk about \textit{"the ultimate goal of this language is to align AI and our values"}, and \textit{"the essence of the alignment problem is the disconnect between intention (the purpose that we desire) and results (the purpose that we put into the machine)"}. I.e. we have no interpretability in current \textit{DL}, since it produces results without any intention behind it. Hence there is the need to supply or teach the AI with intentions, and not just any intentions, but those understandable to humans (hence the "alignment").

\begin{figure}[H]
\centering
\includegraphics[width=0.75\textwidth]{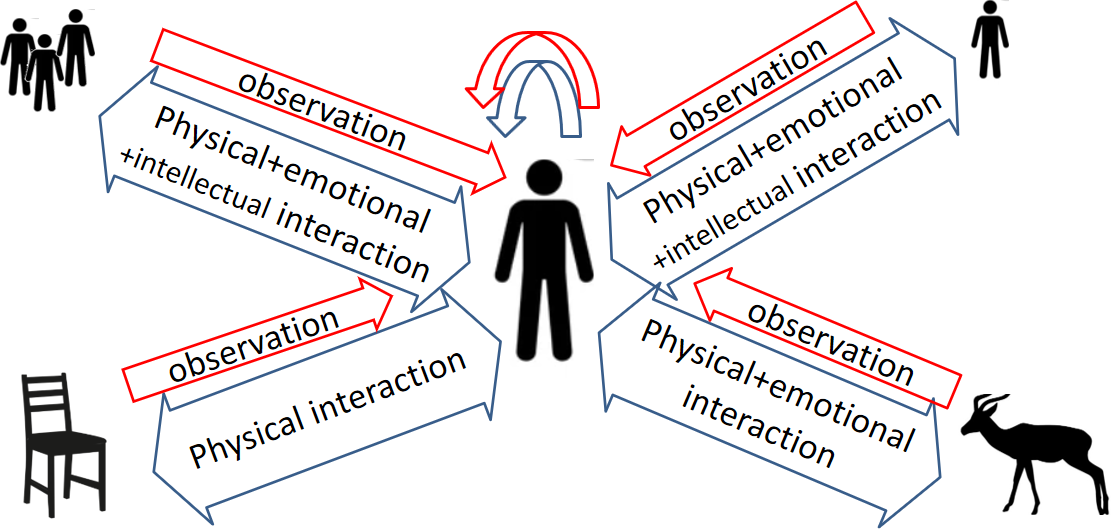}
\caption{Human create models from interaction, taken from \citep{EasyChair:7921}.}
\label{fig:Modelling_machines}
\end{figure}

Interactions we have can be passive (observing/mimicking) or active (e.g. RL), see Fig.~\ref{fig:Modelling_machines}. But it is not only for modeling external processes, but more generally it could model inner ones (meta-knowledge), such as introspection, or how to update models or create new ones, etc.
For example, there are general problem-solving approaches, such as forward/backward chaining, however, this is also consider to be model, and perhaps there should be different models (such as there are different heuristics) for different subjects. These models may be external as any other learned model, which can be recognized or associated when necessary.

These interactions, inner or outer, represent different models in the world. Next, we can wonder, why will is needed to be part of these models. Because most of classic AI studies assume that cognition is mainly about logic, which result with models represented merely by logic. But in our opinion, all these interactions presented above, are not represented merely by logic, but they also include very fundamental and crucial component - will. We assume that will is in every interaction an agent has with the world/environment. Either it poses its own will (animals and humans) or it has some function/role in someone else's will, e.g. an inanimate object has will embedded in its meaning or interpretation to the human (or AI agent). To its perspective or perception of the world. 

\begin{figure}[H]
\centering
\includegraphics[width=0.99\textwidth]{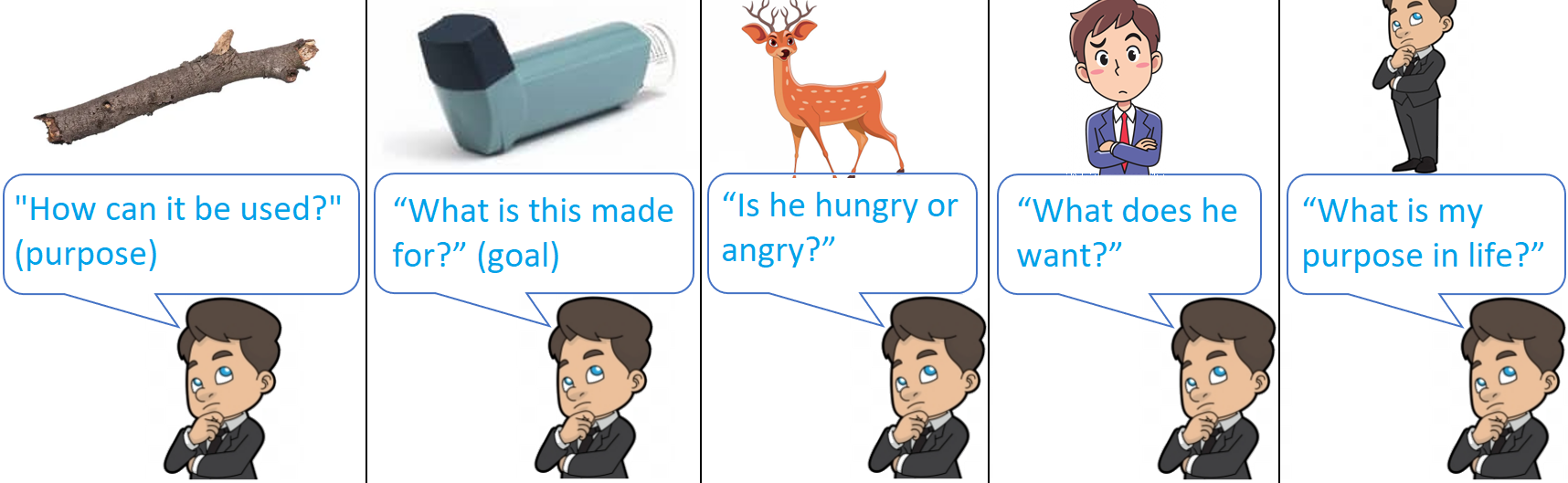}
\caption{Will is significant in modeling.}
\label{fig:will_significance_in_modeling}
\end{figure}

For example, as seen in Fig.~\ref{fig:will_significance_in_modeling}, when we encounter an object, we try to evaluate its function or its potential use for some purpose. While in encountering an animal, we try to figure out the animal inner state so that we would know how to anticipate and understand its behavior. Similarly, while encountering humans (single or a group), we try to extract their will from their decisions, actions or behavior, since we obviously assume that these outputs are derived from their inner will. Finally we externalize will to other realms, such as philosophical queries, where we wonder about the purpose of the world, our purpose in it, and so on.

From these examples and the discussion above, we can conclude that will is rooted deeply within us, perhaps it's our basic prior knowledge. This brings us to the belief, that a full model of any entity is consists both of knowledge about it and about some will associated to it. Or in a formula:
\begin{equation}
\text{Full model of anything = Knowledge + will.}
\end{equation}

This assumption can also demonstrate what a partial model is. One good example of a partial model can be seen in teacher-student interaction. For example in math classes, where the student do not comprehend the material he learns deeply. Though he can certainly absorb only the shallow view of it. Either it is a simple memorization, or a technical understanding of the material, such as not understanding the meaning of stuff but only their utilization, i.e. technical understanding. In other words, the student can end up with just knowing how a method or some formula can be used, without deep and full understanding of the concepts, such as determinant in algebra or Green's theorem in calculus.\footnote{These examples of misunderstandings in learning were the drive for the SciSoft idea: to make one detailed system of all theories and models, with all the necessary information presented to the student. More about it is discussed in \ref{sec:SciSoft}.}
These are examples of partial model.

Back to the bottom-up, top-down learning approaches.
These top-down and bottom-up approaches might contain models of concepts that do not belong to them both, but only to one of them. For example, concepts that are hard to define, like love, God, beauty, and tacit/unconscious knowledge like walking and breathing - all can be modeled simply by examples. Similarly are the sub-symbolic features, like audio and visual inputs - they do not have logic/linguistic/symbolic meaning, hence should be modeled by examples, as it is done in \textit{DL} nowadays.
\footnote{Also, as said by  John Von Neumann: \textit{"In mathematics you don’t understand things. You just get used to them"}, it may imply that children and adults perceive mathematics also by examples, without real comprehension of the concepts, i.e. perceive only their utilization and understanding their consistency.}
Hence, these non-symbolic concepts can be learned via usual non-interpretable \textit{DL}.
On the other hand, abstract concepts, like those in math and sciences, that appear less in the physical reality, can be learned solely in the top levels of LTM.

\begin{figure}[H]
\centering
\includegraphics[width=0.95\textwidth]{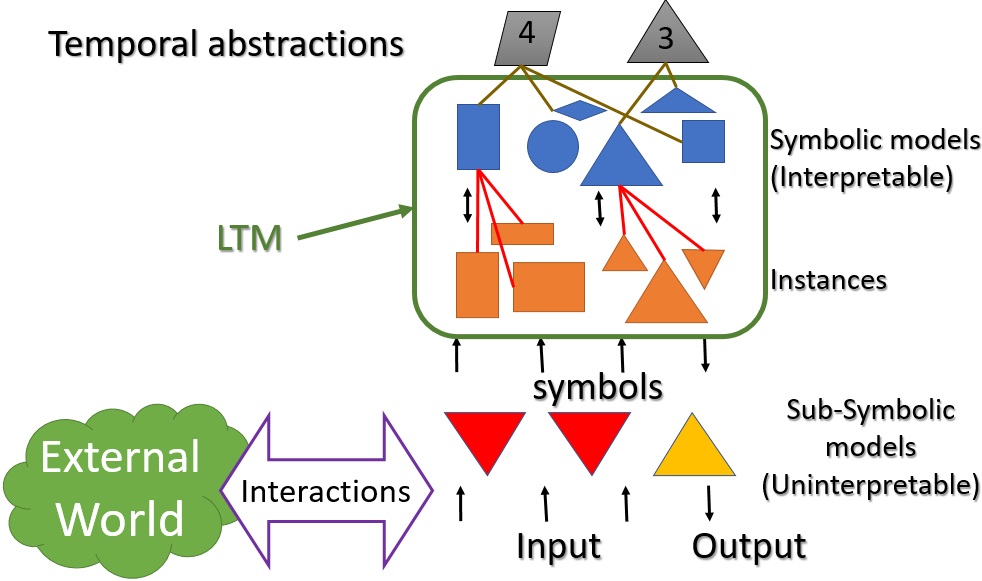}
\caption{Proposed cognitive model basic diagram}
\label{fig:Modelling_machines2}
\end{figure}

Moreover, in this hybrid approach, \textit{DL} is used twice: the extension of \textit{DL} and the use of \textit{DL} as the base level on top of which symbolic processing develops.
On the one hand, \textit{DL} is extended from its too constraint program-search to be much more flexible, if more operations are added as building blocks, see \ref{sec:Operationability}. Hence, symbolism is learned and adaptive just like in \textit{DL}, differently from most expert/rule-based AI. 
On the other hand, different input sensors are fused (e.g. visual stimuli with textual one) to represent specific symbols/concepts, i.e., the uninterpreted features in \textit{DL} become symbolic tokens (Fig.~\ref{fig:Modelling_machines2}).

This utilization above of \textit{DL} is hybrid because on the one hand classic AI is about enforcing our exact view of how the \textit{AGI} system should be (symbolic knowledge representation), while in contrary, \textit{DL} is about letting the \textit{AGI} system to find its own way with minimum bias from us. It is also the tension between heritage verse environmental influence. Same is here. What we do not know for now to put implicitly as inductive bias we put explicitly in a form of direct program ("heritage").

More about the reasons for advocating a hybrid approach can be found in Appendix~\ref{sec:hybrid}.

Additionally, the topmost level in Fig.~\ref{fig:Modelling_machines2} is actually temporal and used for creativity and problem-solving. In this level, temporal new abstractions are created, by stripping off attributes/actions/relations, thus connecting distant or different abstractions to perform for example analogy or transfer learning between different domains \citep{gentner1983structure}. For example, the abstraction is via the number of edges in the polygon classes (Fig.~\ref{fig:Modelling_machines2}). Or by abstracting the branching property of trees we could recognize "treeness" in a picture of a human lung. More generally, differently to structural similarity, i.e. in the relations, analogy could be based upon pragmatic similarity, such as in the goals/purposes. 
Note that abstraction can be applied to any element of knowledge, beyond objects, e.g. actions, and even complex objects/actions, i.e. any group of classes.

An example for using analogy, to transfer one problem to another problem, from a different domain, and abstract its solution, then specialize it for the target problem and eventually apply it, can be seen in \citep{bassok2003analogical}. 
As can be seen, it is about ascending models and descending them. It is actually about searching solution in the immediate neighborhood, by the available association and admissible actions. Then, if we get stuck and do not find a solution, we can rise to a more abstract level, to create some temporal (or existing) class(es), from which we can search for similar problems/solutions much farther then our current location.

Additionally, we see in the main AI fields that there is a single continuous aggregated hierarchy of knowledge, either in vision, such in CNNs, or in the classical processing levels in NLP. Although it is still a viable representation,  we propose another possibility: to separate this hierarchy into two parts. This can be inspired by the diagram in Fig.~\ref{fig:Modelling_machines2}. The first part is when we communicate with the world, or external type of processing, while the second part is when we process things internally. It can be also match our philosophy of min-body approach, where the external processes, like perception (input) or actuation (output) are merely the interfaces to communicate with the world which is not us, or the agent itself. However, the internal part represent our self, our interpretations, and so on. Such that, it is hypothesized that there could be two independent processes: top-down and bottom-up\footnote{This also represents the fact that we have almost separate approaches in AI: symbolic and connectionsm, that still barely can be fused. Similarly to the two major theories in physics: general relativity and quantum mechanics. As in physics, where we could have top-down large to small separate model and a bottom-up one, we could similarly propose here.}.

Actually, everything has a prior. Bottom-up processing include prior about external phenomena. Top-down modeling has also prior, even much more of them. Such as will, causality, physical assumptions, and more.
Moreover, as well known supervised learning, in which symbols are given as the end values of a NN, to learn all inter-mediate features between them and the subsymbolic input. The above idea can replace these symbols as true labels, with the top-down model prior knowledge.

One way to achieve this is by the known pre-training in DL. That is, the subsymbolic processing can be done in an unsupervised learning fashion, e.g. via Deep Belief Network (DBN). While the top-down processing makes the little tuning of the subsymbolic output to the appropriate classes (like the supervised learning small training phase).

\begin{figure}[H]
	\centering
	\includegraphics[width=0.95\textwidth]{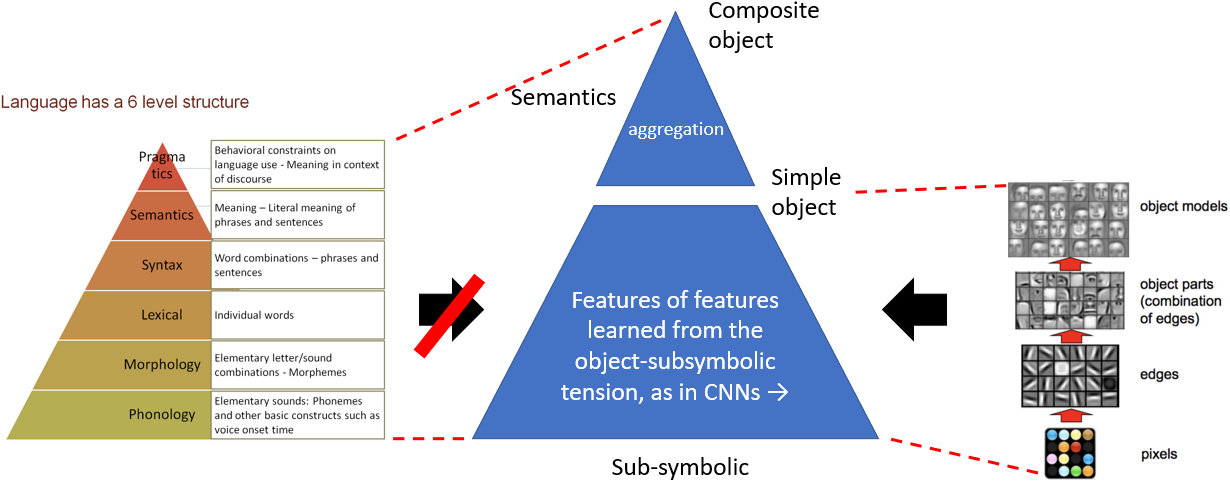}
	\caption{MOM hierarchy of external processing and internal processing}
	\label{fig:MOM_hierarchy}
\end{figure}

This is illustrated in Fig.~\ref{fig:MOM_hierarchy}. In it we see the NLP hierarchy, as one legitimate model, which allegedly captures both subsymbolic and symbolic processes in one hierarchy. We propose that semantics, the meaning of things perceived from outside and acted upon, are not propagate further, but rather stop at some level, and let top-down processes to dictate the meanings of things perceived from below. It means that we consider the automatic hierarchical feature generation by CNN, to be merely a tool for semantical recognition. A recognition of anything we consider to be meaningful to us, e.g. objects, their semantic features (not the sub-symbolic ones) and actions. This learning in CNN were producing visual features by the tension from (input, output) = (subsymbolic data, symbolic data) at the boundaries of the CNN, while trained via supervised learning. However, it can also be produced via unsupervised learning.
 
Even though we propose that the higher semantic hierarchy is driven by aggregation, it can be done by something else, but still separated from the subsymbolic processes.
Another option could be making the whole hierarchy learnable for example.

Additionally, learning models are not done in isolation. These learned models should be coordinated or consistent with each other, to avoid failures and confusions. Consistency-check is about learning model(s) in consistency with other existing models. It is beyond any type of transfer learning \citep{schunk2012learning}:
\begin{itemize}
\item Zero, positive and negative transfer 
measure how learning new task influences previous acquired tasks. Though positive is always preferred, negative is sometimes a consequence of previous incomplete or erroneous knowledge that should be revised. Still, if negative learning is not being regulated it may even result with forgetfulness. So the discussed consistency should be guaranteed within some range.
\item Forward and backward transfer \citep{wang2019forward}, is about transfer from old tasks to new ones (forward) or from new ones to old ones (backward).
\item Near and far transfer, is about the distance of new task to previous ones.
\end{itemize}


All these different transfer processes require active online cognitive effort (also referred to as "High road transfer") or offline reorganization of knowledge (also referred to as "Low road transfer"), to reveal new connections and analogies.


Finally, the operational modeling idea, the outcome of the learning approaches described so far, is not connectionistic like \textit{DL}, but made as classic AI, or logic-based. It is our goal intelligence. But in order to reach it, we propose \textit{DL} alongside with consolidation that described earlier. In neuroscience we can see evidence that higher cognitive animals use this kind of system, e.g. in \citep{piza2023hippocampus}, 
where marmosets (type of monkeys) use a “gaze-based” spatial navigation, contrary to the “place-based” navigation observed in rats. That is, to navigate they use more cognitive skills like understanding the functioning and the model of the objects they recognize, instead of merely physical orientation in the surroundings.

Similarly, while we recognize objects by similar techniques to \textit{DL}, i.e. CNNs, our models are still logic-based, e.g. in recognizing digits (by V1$\rightarrow$V2$\rightarrow$V3$\rightarrow$V4 brain perception) verse understanding their generation rules, e.g. 9 = counterclockwise small circle + clockwise half an arc.

\subsection{Evolution prior} \label{sec:Evolution_prior}

The main issue with learning, is to find the right guiding prior for the top-down processing. It is the core intelligence problem, in which the top-down guiding prior determine how to deal with the bottom-up perceived sensory data.

Up until know we have a lot of prior to embed in our AGI system. Prior like making sense, full modeling which includes also the will behind the actions, basic prediction, spatio-temporal grouping and abstraction, and more.

However, it is difficult to imagine how to imply all these priors at once. We must go back to psychological theories of development in stages, like Jean Piaget and Erik Erikson theories.

In our opinion, it is the development around the expressing or perceiving different types of will.
In brief, it starts from "what" question inquiries, to "how" questions, to "why" questions. Each of those with regards to our immediate social surroundings. 

More precisely, it starts first from detecting single objects and entities in visionary and auditory inputs. It includes prediction of tracking these objects and their transformations. This is the "what" inquiries, trying to align the will "what" with the perceived data.

Next, is the "how" will. It is about behavior of these objects. I.e., how they transform, how they act and react, and so on. Then, we also include the agent itself, not only external observation, i.e. interaction. The "how" also include "how" I can act in my environment. This means moving, throwing, braking, and so on.
As stated in Section~\ref{sec:will_types}, these objects do not posses their own will, hence it is easier and make more sense to start with. Meaning the prior evolve from simple to complex, in stages. So will is not needed in these stages.

Next, after gaining enough practice of objects and their behavior, and how the agent and other agents affect objects, themselves and others, we can turn to the "why" question. But the answers are basic: simple type of wills (originated from basic desires). This is the lowest level of social interaction the agent has.

A higher "why" questions that follow, are of more deepen sources of wills, both in the agents, and hence also searched also in other agents as well.

All the above means, that the prior is applied gradually or conditionally, i.e. only when the "what" is satisfied enough (how much is enough), is when we are prepared to the next stage of "how", and then "why". And then extend it further, as a function of our affect.

However, it could be that these stages are not roughly separated, but all exists in the same time, only have different dominance at different times, just like brain waves.

Eventually, we back to the main issue of learning described above. Or in other words, how human will or human intention can be materialized? for example in a software or algorithm. It is a very difficult question, since this concept is too abstract and amorphic. We have no direct connection to it, but only implicitly, through perception and our own will realization. But we have no idea what is it and how it is materialized\footnote{This is why top-down prior is learned, as anything in AI agents. What in humans is simply some sort of soul guiding the learning, the growth, and the adaptation to environment, in our opinion.}.

However, we can overcome this issue. For example, as proposed for \textit{AKREM} in Appendix~\ref{sec:AKREM_evolution}.
In \textit{MOM} on the other hand, it can be realized in the following steps.
The "what" questions are simply noticing many different features of different entities and objects, and assigning them to these objects. All this done without language or logic, in an unsupervised learning fashion.
Then, the "how" is attaching actions to these objects. Obviously this step requires the previous step: of having objects that these actions applied on.
Finally, the "why" is handled by adding directionality to these actions. Also this requires the previous step: having actions. This directionality is a feature that connects all different actions, so that they are not merely independent entities, but entities that relative to each other. These actions are also context-dependent. 

\begin{figure}[H]%
	\centering
	\subfigure[What-step: objects+features]{%
		\label{fig:a}%
		\includegraphics[width=0.48\textwidth]{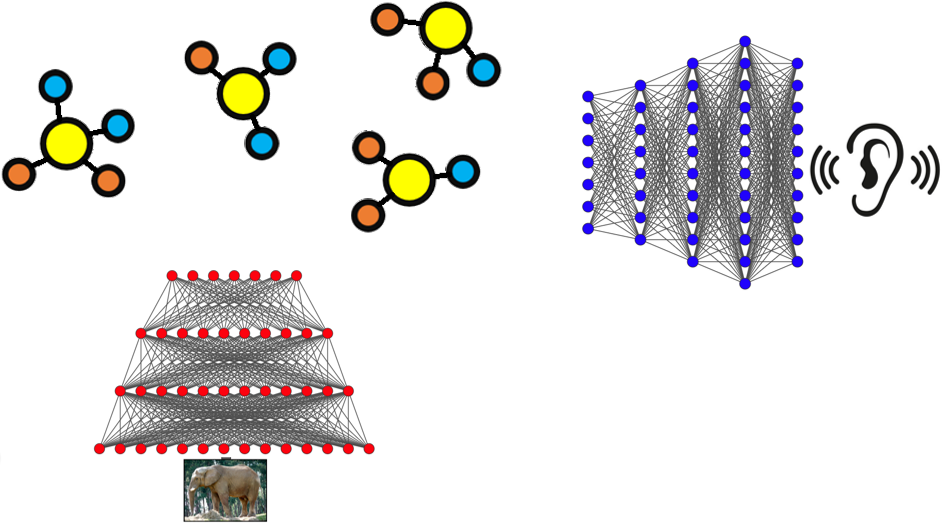}}%
	\hfill
	\subfigure[Why-step: objects+features+directional actions]{%
		\label{fig:b}%
		\includegraphics[width=0.48\textwidth]{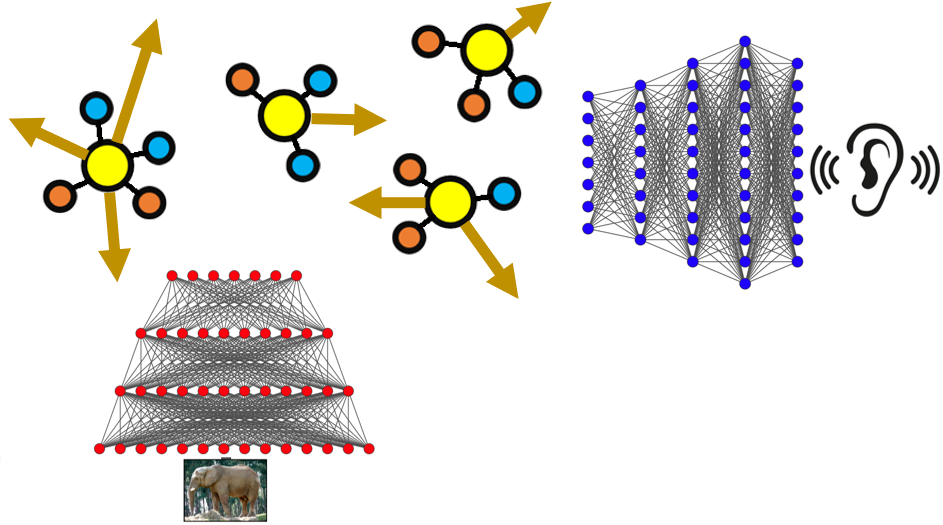}}%
	\caption{Evolutionary steps, neglecting the how-step}
	\label{fig:evolution_prior}
\end{figure}

See these evolutionary steps in Fig.~\ref{fig:evolution_prior}, where the small blue circles represent audio semantic features, the small orange circles represent visual semantic features, yellow big circles represent objects, and the orange arrows represent actions with directionality. Note that due to redundant repetition, we omitted the "how" step, as the intermediate step between Fig.~\ref{fig:evolution_prior}(a) and Fig.~\ref{fig:evolution_prior}(b), which is exactly like Fig.~\ref{fig:evolution_prior}(b) only without the arrows at the end of the orange lines.

In summary, this overall process is cyclic, i.e. whenever new features learned, new associations to objects can occur. New actions can always be added, and directions can be updated.
Also, note that basic objects and their features are already abstraction. It means that even the most basic instances we know from \textit{OOP} are already abstraction or generalization of different features that describe it. In other words, the most basic semantic objects are already generalizations.


\section{Implementation directions} \label{sec:Iplementation}


\subsection{Model separation} \label{sec:ModelSeparation}

Another issue that precedes learning, is how to obtain separate models at all. One way, is like the \textit{DENN} (Dynamic and evolving NN) idea \citep{10.1007/978-3-031-19907-3_5}
, i.e. always learning the "model of everything" while refining it more and more with every new experience, such that new sub-models are produced. 
The idea here, is like Jeff's hierarchy \citep{hawkins2007intelligence}, where the top model is always reached, at perception from senses, and it decides which lower model will handle the situation. E.g. model for problem-solving, for learning, and for story message (where it connects separate events sequentially). And it can go on further. For example, problem-solving model chooses the most appropriate model for solving a particular problem. Or, a learning model selects the most appropriate model for assimilating new knowledge. 
Or, conversational model being about taking turns, waiting till me/other side is finished, recognizing self models\footnote{We can adapt here the philosophical perspective of identifying myself as an observer, i.e. an observer of thoughts, actions, etc. Meaning, treating my self model as equivalent to others self model. This way we also remove the ego (and perhaps any selfish will) from the agent. Also this way embodiment is a way to recognize its body as a tool to fulfill its will.}, deciding whether the sender's message imply the updating of some models due to conflicts for example, or its simply was not interpreted correctly or something wrong with the message hence should be clarified or repeated\footnote{Hence, the need for prior knowledge inserted via some inner information embedded in the models, such as consolidation measure, or in this case uncertainty parameter and/or the rate of recent model updates.}, etc. 
Or in perceiving: perceiving fictional information is treated differently than factual information, and so on.
Or in "act like" missions: act like some of the known models of other agents/people, or act more creatively (creative model).

The idea from \citep{EasyChair:7922} can present similar duality here also. The above refinement can be regarded as top-down model evolving, while there could be also bottom-up one, where existing models can be grouped into a new or existing model, similar to the opposite operations in \citep{EasyChair:7922}: merging verse splitting.
Hence, the modeling itself can be refined to convergence as the famous AI's generalization and specialization operations, using positive and negative examples.

Furthermore, there is this common idea that an infant holds already about 100 billion neurons, and what change along the development is only the connections between neurons. This enforce our view of separation, since we could have a static NN, and the learning will only change the separation into models via different connections.

\begin{figure}[!htb]
	\centering
	\includegraphics[width=0.99\textwidth]{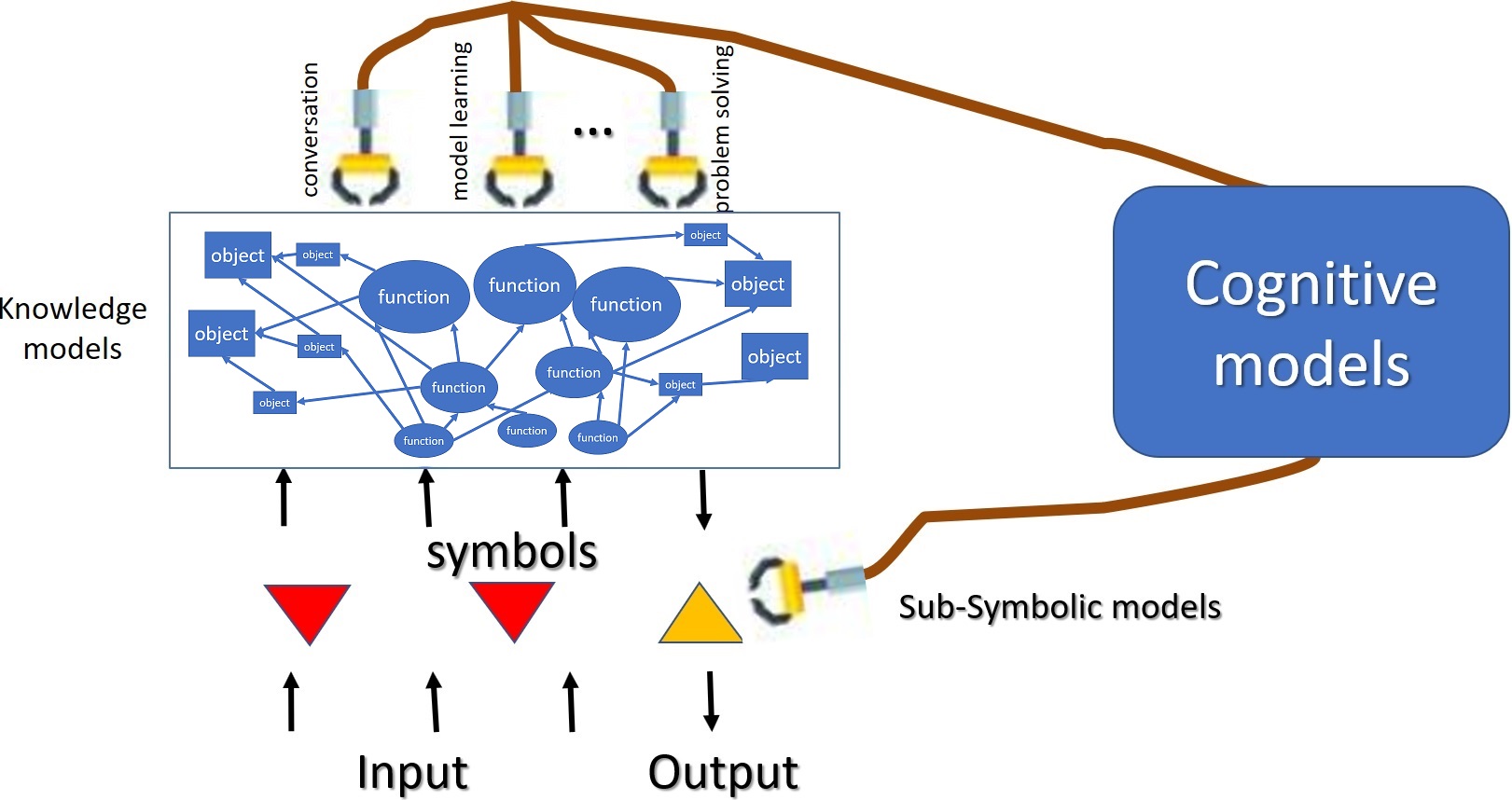}
	\caption{Separating cognitive and knowledge models}
	\label{fig:Modelling_machines3}
\end{figure}

In conclusion, this option was produced since problem-solving and alike are very complex models, which is why the proposal to separate them from the knowledge models. But it extends further - perhaps there is separation of model representation. May be some models can be represented as operational classes, but others cannot\footnote{See similar idea in Godel's ideas, where truths are more generally are not algorithmic. Or in \url{https://medium.com/paul-austin-murphys-essays-on-philosophy/roger-penrose-on-kurt-g\%C3\%B6del-and-g\%C3\%B6delian-truth-7cc6e8f79069},
where it is stated that there is separation of mathematical truth verse sensing, and that we have more access to mathematical objects than to perception, of a rose for example.}. These other models could be not interpretable nor can be explained by the agent, e.g. models that are placed in the background of thinking itself, thus they are “hidden” or implicit. 
It can also represented as declarative type of memory (knowledge models) verse implicit memory (other models).
See Fig.~\ref{fig:Modelling_machines3}.

One way to implement model separation is proposed in \url{https://matt-rickard.com/mixture-of-experts-is-gpt-4-just-eight-smaller-models}, where they propose switch transformers, to generate sparse distribution among experts.

\begin{figure}[!htb]%
	\centering
	\subfigure[Nested DNN structure]{%
		\label{fig:a}%
		\includegraphics[width=0.9\textwidth]{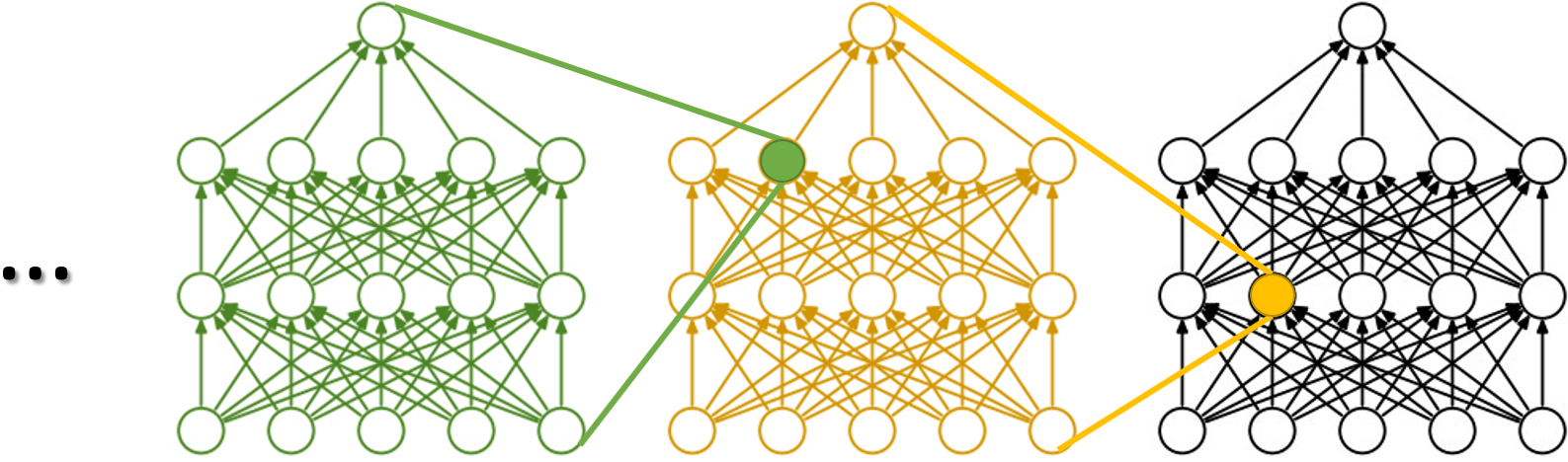}}%
	\hfill
	\subfigure[Initial Nested DNN]{%
		\label{fig:b}%
		\includegraphics[width=0.44\textwidth]{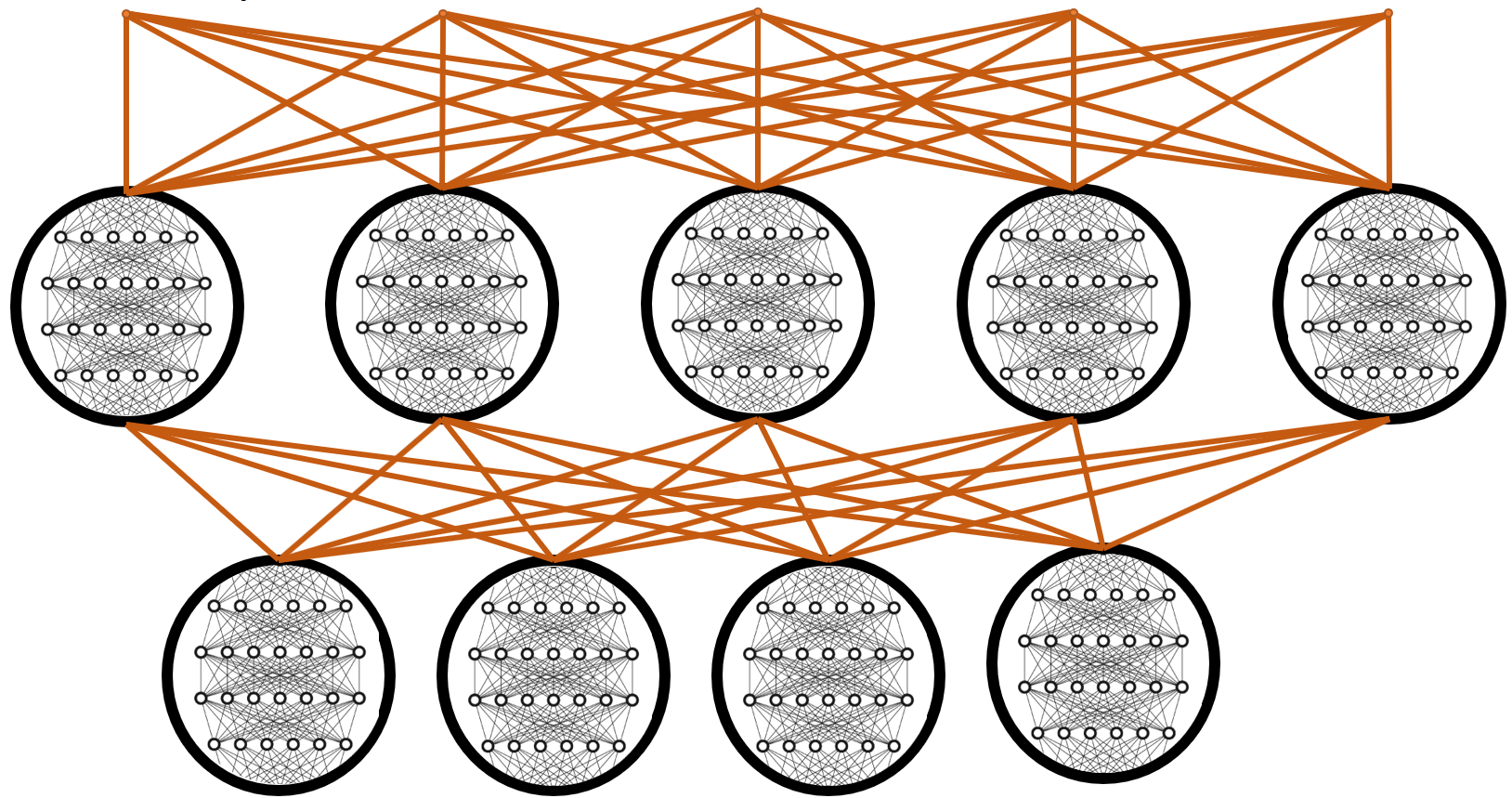}}%
	\hfill
	\subfigure[Learned Nested DNN]{%
		\label{fig:b}%
		\includegraphics[width=0.44\textwidth]{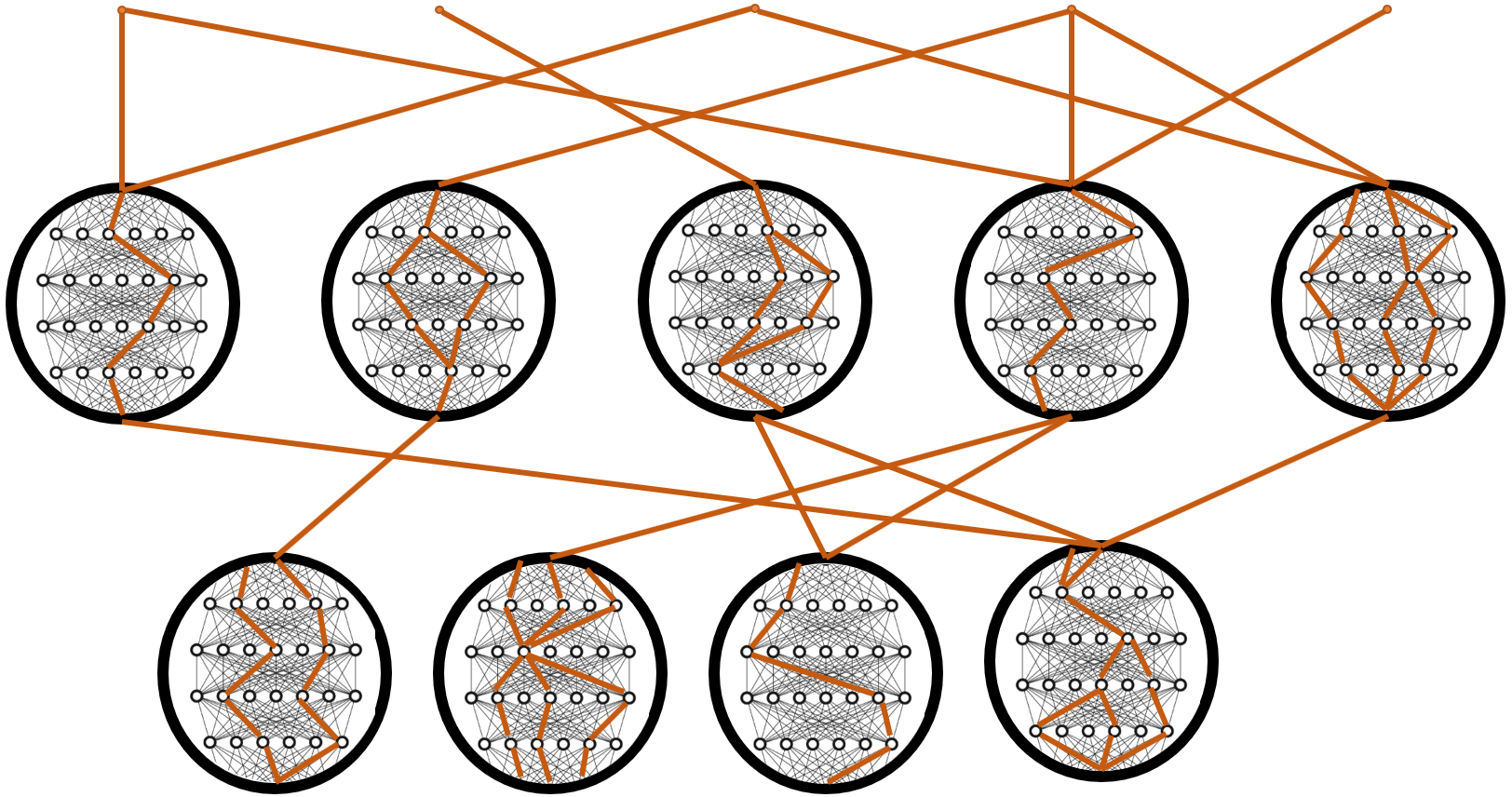}}%
	\caption{Nested DNNs for model learning}
	\label{fig:Multiple_Consolidation}
\end{figure}

Another way to implement model separation is by nested NNs (or NN of NNs) \citep{beniaguev2021single, gidon2020dendritic}, see Fig.~\ref{fig:Multiple_Consolidation}(a), where consolidation occurs in multiple scales. It can be encountered through many phenomena in nature, e.g. in the universe (consolidation into stars/solar-systems and galaxies), in fractals (such as snow-flakes), and in other recursive structures. It can be also seen in the transition from some initial network of unlearned models Fig.~\ref{fig:Multiple_Consolidation}(b) to Fig.~\ref{fig:Multiple_Consolidation}(c), as a consolidation in multiple levels, both in micro (within models), and in macro (between models). 
The modeling or the reorganization of inner elements, is occurring at many levels of models, i.e. from the basic models to the most complex ones.

Recently, a neuroscience article \citep{gidon2020dendritic} discovered evidence in the brain, that supports our system twice:
\begin{enumerate}
	\item First, it extends the normal binary operation of all-or-none message passage between neurons, i.e. if the signal overcome some threshold. Now they found additional logical functions, that are quite basic and necessary in our system: AND, OR, and even XOR.
	\item Second, it admits that these complex action functions considered so far to require multi-layer NN to be implemented: \textit{"These action potentials allow single neurons to solve two long-standing computational problems in neuroscience that were considered to require multi-layer neural networks."}.
\end{enumerate}

Moreover, as it is well known \citep{minsky1961steps}, DNNs can implement logic gates (AND, OR) with 1 layer and more complex gates like XOR with several layers. Hence, combining with the article above \citep{gidon2020dendritic}, it strengthens our hypothesis that brain neurons are DNNs themselves.

\subsection{Attention} \label{sec:attention}

After assuming a set of unlearned models, Fig.~\ref{fig:Multiple_Consolidation}(b), it can be assumed further, that will or more probably consciousness, acting like a flashlight or a beacon, produces consistent attention (over time) to learn/attend each model (or several of them) individually.

\begin{figure}[!htb]%
	\centering
	\subfigure[in general 1]{%
		\label{fig:a}%
		\includegraphics[width=0.47\textwidth]{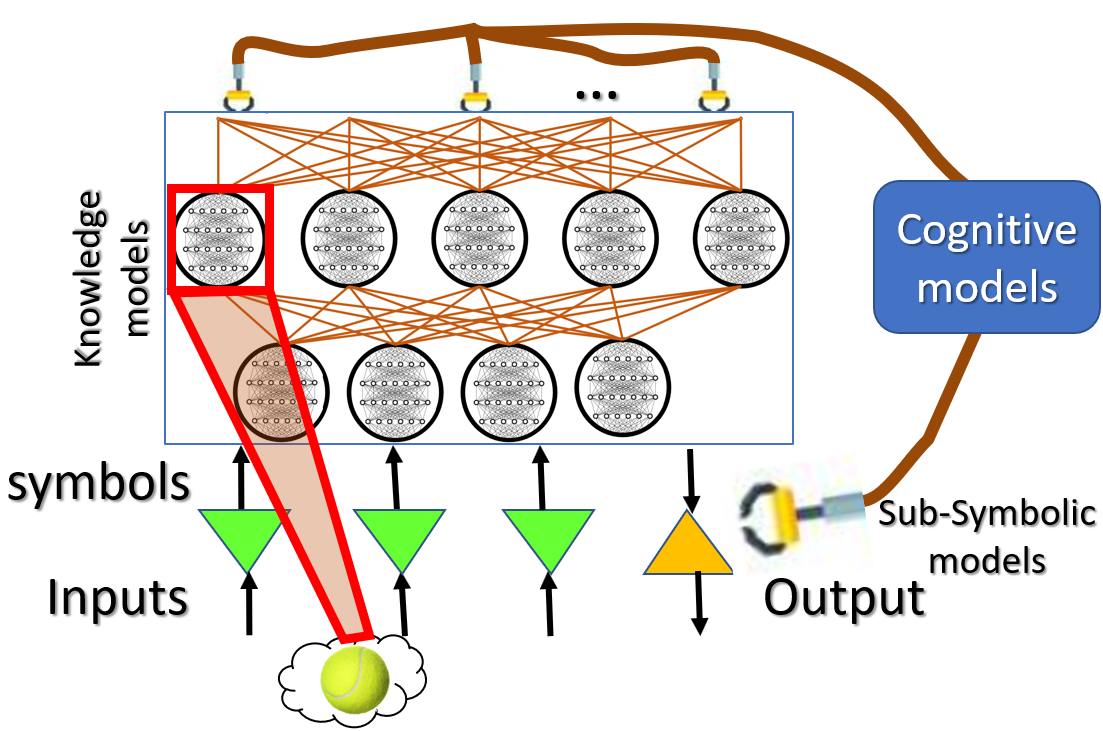}}%
	\hfill
	\subfigure[in general 2]{%
		\label{fig:b}%
		\includegraphics[width=0.47\textwidth]{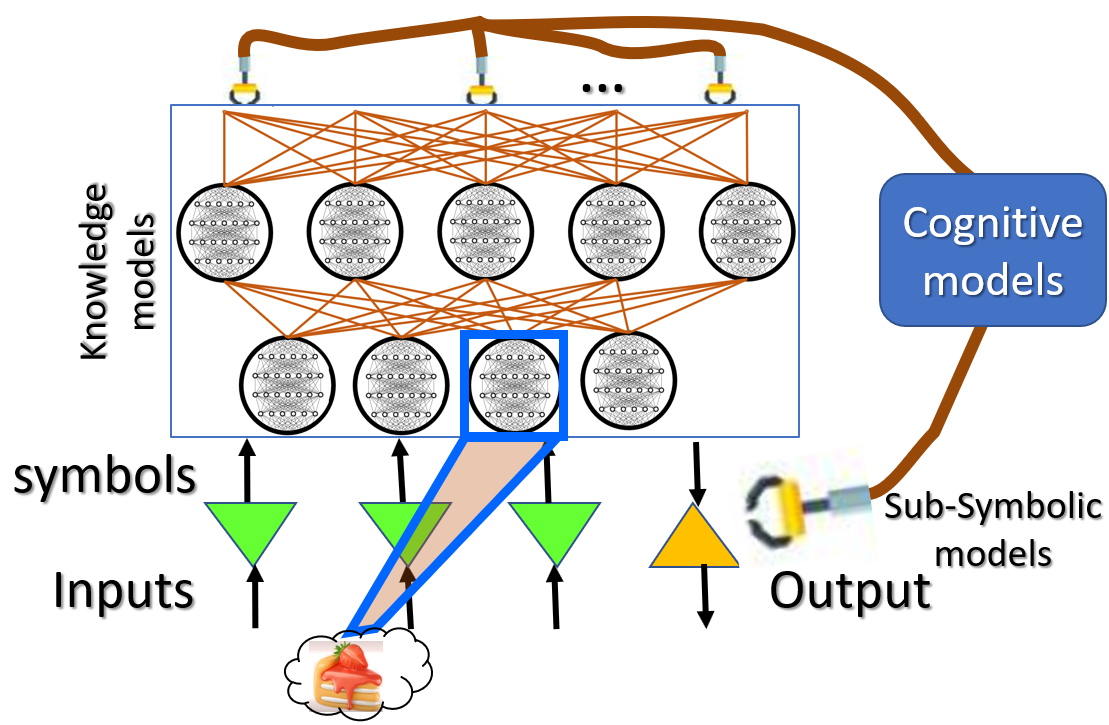}}%
	\hfill
	\subfigure[in visual perception]{%
		\label{fig:c}%
		\includegraphics[width=0.8\textwidth]{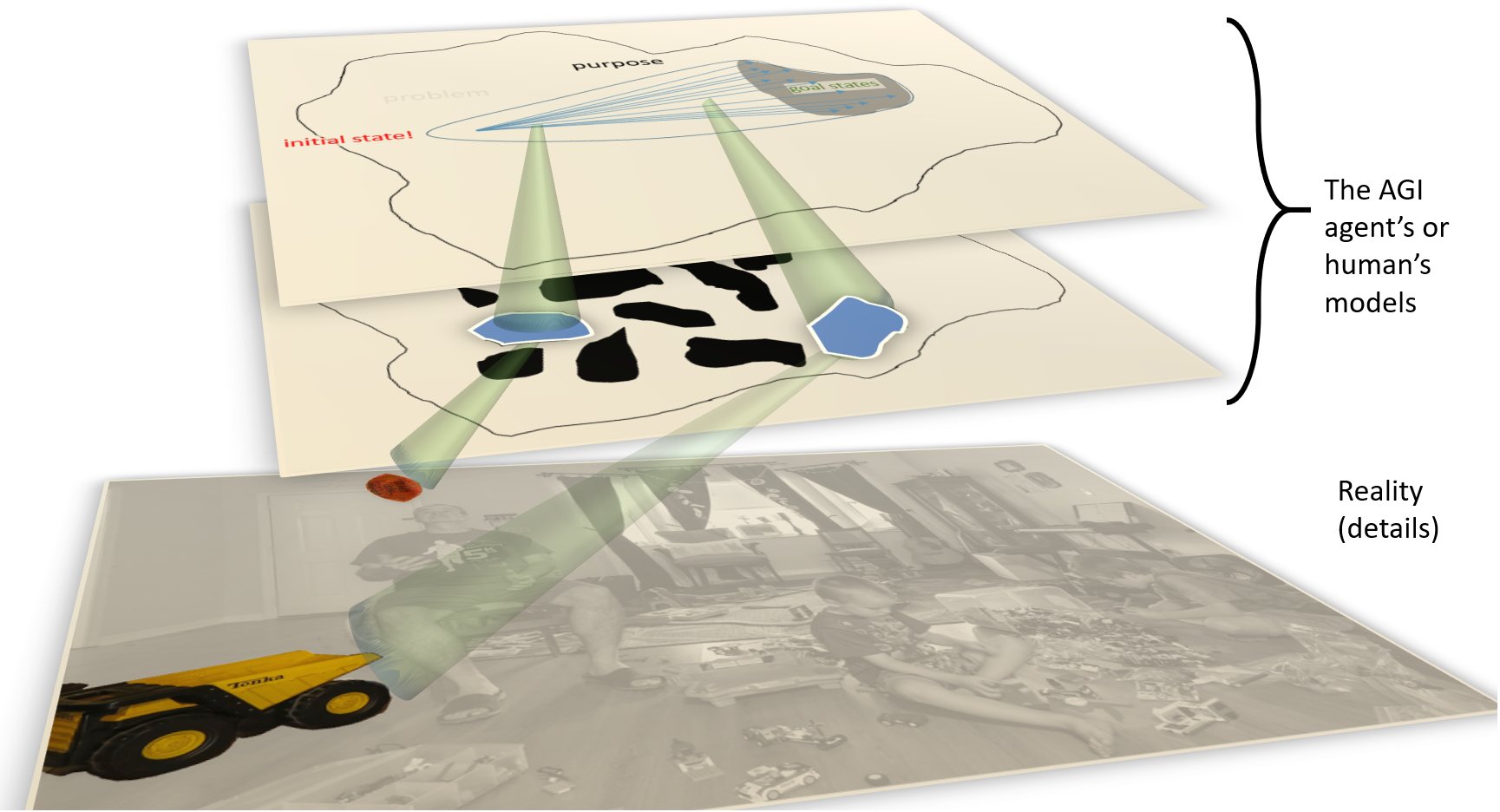}}
	\caption{Attention in models}
	\label{fig:attnetive_model_learning}
\end{figure}

This concept explains why an infant is usually very focused over his toys (e.g. a ball), and tracking them is essential for this process. This effect sticks till adulthood, also in the process of using the cognitive model (i.e. after the learning stage). It is the need to be attentive only to a limited set of models ($7\pm 2$ items in WM) \citep{plass2010cognitive}.
See examples in Fig.~\ref{fig:attnetive_model_learning}. Fig.~\ref{fig:attnetive_model_learning}(c) demonstrates split attention concentrating on two specific models: a football and a truck toy.

In other words, will is coming from the top (or from the side of an hierarchy, if it does not interested in top-down focusing but on specific models at any level), shining like a projector, focusing on one/few models, while tracking it/them in the real world.

Therefore, the utilization of attention can be illustrated in an example of an infant learning a "ball" model. During waking hours, an infant is gathering instances of its current models, e.g. a ball, and at sleeping, he uses these instances to train his models, for the purpose of making sense. First, he tries to figure out different models, then he tries to model them also in time, thus he is able eventually to track them, which is a validation of the correctness of his "ball" model \citep{Ganguli}. Because the final test of his model is prediction, hence temporal modeling is what enables prediction, or more specifically forecasting (prediction in time) \citep{ganguli2018intertwined}.

Also, attention is needed to construct a coherent picture of visual perception, see \citep{roelfsema2023solving}, 
which explains it by attending each object in a given scene separately.

Note, that attention to a few models also implies that just as humans, \textit{AGI} agent need not to understand and model everything, but only what it is focused on or interested with.
Also, there is the idea of bidirectional attention, which is bottom-up (external) verse top-down (its own will), and describes the competition between having a (strong) will and being highly influenced by the outside. In \textit{AGI}'s case, it should be mostly navigated by external guidance, if will is not engineered into it.

In addition, attention can have different "focal length", like the theory of vision, having small pinhole perception at lower levels (e.g V1), and a bigger one at higher ones (e.g. V4). Meaning, the ability to sometimes see small details and sometimes see the big picture. In model attention it is the same: we can both have low-level more detailed attention on smaller models, upto a high-level attention for more general or composite models. In comparison to classical object detection in computer vision, high-level classes/concepts use only the higher-level features for the classification task, but more generally there is no reason not to be attentive to low-level features whenever is needed (hence the use of residual connections as in ResNet and Transformer).

Moreover, in regarding to focus (or continuous attention), see \url{https://medium.com/publishous/lack-of-focus-heres-what-you-need-to-know-from-a-neuroscience-point-of-view-90b6f33942d7}, there is the problem of associative behavior against focusing on purposeful thought, e.g. at problem solving. It means, there is on the one hand the process of recalling relative past knowledge, verse on the other hand the process of current situation handling. It is also a duality (between attending the past and the present/future).

In addition, it is well known that sub-conscious processes are like seeds, working in the background, where suddenly an idea can pop-up, what's called "aha" moments \citep{sadler2015wallas}. 
This phenomena can encourage our idea of splitting attention between the main consciousness and the hidden one, which works like a detective, trying to fill up missing information, deduct new conclusions, and sometimes even solve larger problems. See more about attention splittings in Appendix~\ref{sec:Creativity}.

These and other examples of attention splitting are examples of how attention is limited and flexible, and it is distributed from total concentration to even but low attention. It resembles the attenuation theory \citep{treisman1964monitoring}, which states that while perceiving input from senses, e.g. audio, we simply lower the volume of the background and turning it up to some desired inputs, i.e. tuning the attention. This explains the ability to hear our name called or call for help in a crowd while talking with someone.

Additionally, there is this battle between senses influencing models (bottom-up attention), verse models influence how senses are interpreted (top-down attention) or whether they are being distorted to fit the models better. See more in \citep{auguste2022oligodendrocyte} 
and in \citep{franke2022state} 
about how our inner state influences sensory attention, i.e. to specific mode of perception.
Similarly, in optical illusion \citep{laeng2022eye}, 
they showed that anticipation to darkness can affect the eye dilation.

Finally, attention in our perspective is very similar to the attention in \textit{DL}, only without regulation. Meaning, without considering the ideal state of consolidating into symbolic reasoning of models and operations and most importantly allowing for dynamic abstraction. However, \textit{DL}'s attention is similar by also allowing for multiple implicit functions in a given learning NN, since it reacts differently depending on the input. In other words, the DNN can be regarded as a group of undeclared models/functions, generated by attention units, thus implicitly implement compositionality and reusability.

Moreover, \textit{DL} has an issue with unspecified prior knowledge, which can be perhaps resolved by attention. One of the difficulties in early \textit{DL} studies, is the decision upon the inner connections in NN or its structure. It could be for example the decision about sharing or grouping features \citep{schlichtkrull2018modeling} or instead separating features. E.g., when traffic data is set apart from weather data \citep{koesdwiady2016improving}, or roads from stations \citep{huang2014deep} and are being fused only later (how later is also a prior knowledge to be decided upon). Or when tasks are separated to groups \citep{huang2014deep}. Or when NN structure is sparsifyed/pruned \citep{ioannou2018structural}. 
All this prior knowledge is difficult to recognize apriori to different tasks, hence we could use \textit{NAS} or attention which deal with this dynamically.


Also, there is the paradigm that the brain works on causality of chemical and electrical processes. On the other hand, in contrary to this deterministic view, there is a non-materialistic paradigm \citep{bentovish2022g}, which allow the free-will, where the mind is not causal, but instead governed by a soul.

If we dig further, in \url{https://bigthink.com/the-well/eastern-philosophy-neuroscience-no-self/}, eastern philosophy and neuroscience research advocate the idea that there is no "self", and that it is only illusion in our mind. It is actually similar to relativity theory, where there is no space without time. In analogy, it means that the brain is materialistic and includes "self" as part of its models. Therefore, it may mean that the soul, separated from this physical brain, is merely a passive, non-participating observer, while thoughts are appearing externally by some unknown source. That is, the observer soul, simply observes external stream of thoughts, and falsely attribute some of them to a "self". In the article they argue this, since the thoughts are not controlled by some "self", although being attributed to some fictitious "self", one of the many models we have of different objects and humans in the world. Hence, our idea of self-model is similar. It is treating ourselves as another "thing" to model.
This observing soul is one possible source for attention. Similarly, the sources can be awareness or consciousness. 

Lastly, attention mechanism also suggest that memories not under attention are still there, accessible, ready to be addressed. Moreover, examples like attending a movie/story while simultaneously trying to comprehend it and fill up missing information imply subconscious processing in the background (see attention beams in the inference in simple examples \ref{sec:simple_examples}).

\subsection{Implementation using Separating models\\and Attention} \label{sec:attn_seperate_models_impl}

If we use the two principles of separating models via attention mechanism, we can try to construct a more specific implementation of learning.

First, we assume that the nested NN structure, as seen in Fig.~\ref{fig:Multiple_Consolidation}. Hence, each neuron is some NN by itself, and can represent all relevant models of some specific concept.

Second, we can imagine how many fundamental concepts are there, to evaluate how many complex neurons we need, assuming a concept per neuron. If we look at an average person, or at a dictionary, or a glossary, or even encyclopedia, for the index of used terms, there are not so much of them. In the most exaggerate case, it is an order of thousands. An average person holds much less concepts in his daily life. Even if we include specific jargon of experts in different field, it is still not exceed thousands of concepts. Hence this is terribly small in comparison to human or artificial NNs.

Third, after we decided to use nested NNs, and agreed upon the scale of concepts to be learned, we can continue to what actually will be learned. We represent each concept (object or action or feature), i.e. class in \textit{OOP}, along with its instances.

Next, we can assume that if the lowest level in our nested NN, is of basic concepts, then the upper layers represent different aggregation of these basic concepts, e.g. different groupings or abstractions. Next level is similarly an aggregation of these aggregations.


\subsection{Multi-version model} \label{sec:Multi-version}

Side note: this section is still under construction, hence should be skipped.

It is very like the Pathways Language Model (PaLM) idea \citep{chowdhery2022palm}, encouraging sparsity, multi-modality and continual learning of tasks (unlike multi-tasking where tasks are pre-defined).
Also it resembles the multi-head attention in Transformers \citep{vaswani2017attention}, where several versions of attention are learned.

One suggestion to implement multi-version is by using 3d NN instead of the usual 2D one, where the paths are not limited to the same 2D network, but also have the freedom to move to other direction in 3D space. One simple option of this idea is a stack of 2D NNs, let say $N$ of such NNs, where the paths can go from any NN to another. This way we can implement cyclic learning, where at each interaction with the environment, the current version of the NN is 

\begin{figure}[H]
	\centering
	\includegraphics[width=0.7\textwidth]{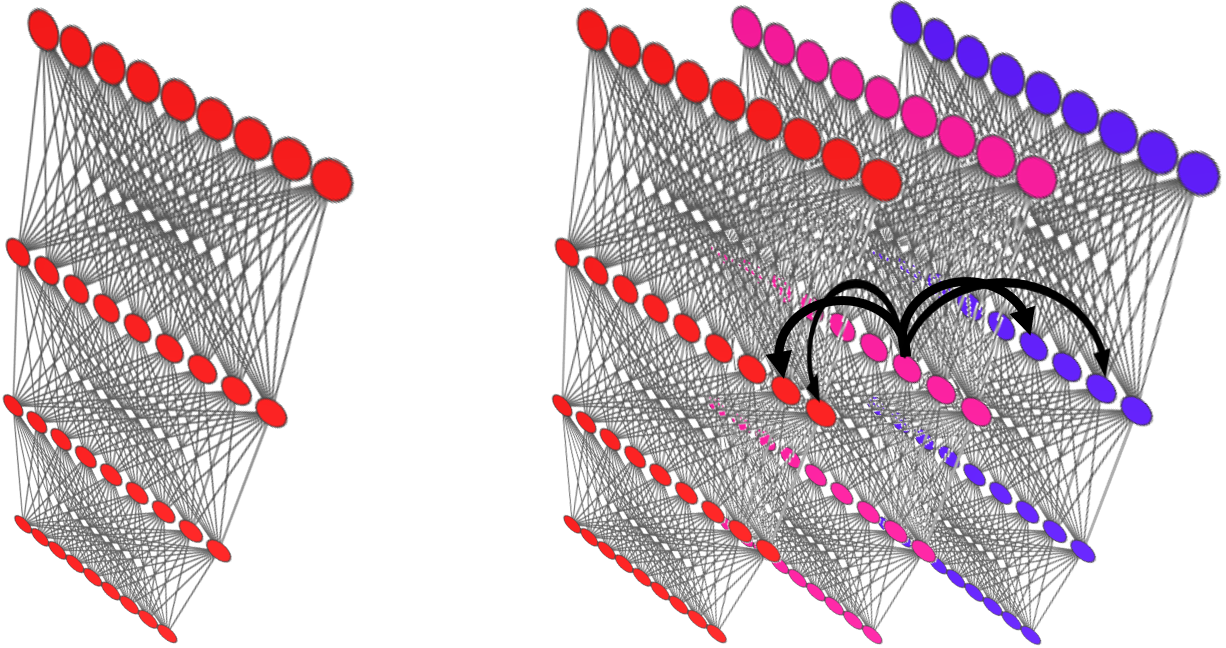}
	\caption{Single-version verse Multi-version implementation suggestion.}
	\label{fig:multi_version}
\end{figure}

See Fig.~\ref{fig:multi_version}.

\subsubsection{Functional programming search} \label{sec:FP}

It is well known that the \textit{DL} representing DNNs is functional, i.e. in order to allow for back-propagation (BP) or differentiation from target to source, it must be a cascade of functions\footnote{This idea, or how assembly programming language works, or any simple machine language, or Turing machine - all might suggest that any algorithm, even if it includes (condition/loop) flows, is a sequence of actions, hence \textit{MOM} propose that any such algorithm representing a method in \textit{OOP} is an aggregation of other actions in a sequence.}, i.e.:
\begin{equation}
	\begin{split}
		Y & = g \circ f_K \circ f_{K-1} \circ ... f_2 \circ f_1(X) = \\
		& = g(a(...a(W_2 \cdot a(W_1 \cdot X+B_1)+B_2)...+B_K))
	\end{split}
\end{equation}

Where $X,Y$ are input and output respectively; $W_i$ and $B_i$ are weight and bias matrices of layer $i$ in the DNN, respectively; $a$ is some activation function (e.g. sigmoid or ReLU); and $g$ is special function for the specific task in the DNN's last layer; and finally $f_i(X) = a(W_i \cdot X+B_i), \quad \forall i=1..K$.

Similarly to the DNN described above, the learnable symbolic manipulation should be a cascade of functions.
Hence, here we transform common programming language tools, the basic building block of any function, into a form of operators or functions. 

First we introduce some common symbolic/logical tools that are used in programming:

\begin{itemize}
	\item Arithmetic operators (+,-,*,/,modulus, power/square-root), 
	\item Mathematical operators (min/max, absolute value, rounding, norm, factorial, sinusoidal/exponential/logarithmic/polynomial functions) and their constants ($e, \pi, \tau$, ...), 
	\item Assignment operators, 
	\item Comparison operators (equality and non-equality), 
	\item Logical operators (and, or, not, all, each, (in)equalities, exists, count), 
	\item Flow operations (while, for, if, if-else, if-elif, ...)
	\item Identity operators, 
	\item Membership operators (if a sequence is presented in an object), 
	\item and Bitwise operators (shifting, AND, OR, NOT, XOR, etc).
\end{itemize}

Then we present some examples of transferring them into functions, in the form: $\text{tool} \Longrightarrow \text{function}$:

\begin{equation}
	\begin{split}
		\text{if x then y} & \Longrightarrow if(x,y) \\
		\text{if x then y else z} & \Longrightarrow ife(x,y,z)
	\end{split}
\end{equation}
So for example a composition could be:
\begin{equation}
	\begin{split}
		& \text{if x then y elif x2 then y2 elif x3 then y3 else y4} \\ & \Longrightarrow ife(x,y,ife(x2,y2,ife(x3,y3,y4)))
	\end{split}
\end{equation}

\begin{equation}
	\begin{split}
		\text{for x do y} & \Longrightarrow for(x,y) \\
		\text{while x do y} & \Longrightarrow while(x,y)	
	\end{split}
\end{equation}

Also expressions:
\begin{equation}
	\begin{split}
		\text{x and y} & \Longrightarrow and(x,y) \\
		\text{x and y and z} & \Longrightarrow and(x,y,z) \\
		\text{x or y} & \Longrightarrow or(x,y) \\
		\text{not x} & \Longrightarrow not(x) \\
		\text{x + y} & \Longrightarrow plus(x,y) \\
		\text{min(x,y,z)} & \Longrightarrow min(x,y,z) \\
		\text{x} \leftarrow \text{y} & \Longrightarrow assign(x,y)
	\end{split}
\end{equation}

As seen, functions can have specific arguments, or unknown list/dictionary of them, as in Python for example. Some of them can be set with default value at function entrance. Also, they allow recursion.
Also, Lambda function, taken from HOL, makes it easy to produce functions on-the-fly. It is an anonymous function, hence it is not needed to be pre-defined. This idea can be utilized similarly to temporal abstractions, i.e., the addition of temporal actions when it is required.

All these temporal items can be stored in some storage unit, for cases when some of them are reoccurred. In such cases, they are transferred to LTM. See more about temporal abstractions in Section~\ref{sec:Learning_the_modeling} and consolidation in Section~\ref{sec:consolidation}. 

Finally, 
how different NN structures are equivalent to a rule-based algorithms (or usual programming)
is explained in Appendix~\ref{sec:DNN_equivalence}.

After the proposal of the basic building blocks, the next issue is what learning approach is should be utilized to consolidate the learning process into the correct program. Two approaches proposed:
\begin{enumerate}
	\item \textbf{Generative}: The program-search is the opposite of \textit{DL}. In \textit{DL} we start from large number of hypotheses and as more examples are used (either more epochs or more examples), the number of possible hypotheses is reduced. But this is due to fixed/static architecture to apply BP in it.
	Here, on the other hand, is the opposite, we start from no hypotheses and generate multiple hypotheses dynamically. Meaning, we propose each time a new composition of basic functions. Since we do not have fixed structure we cannot use BP. However, this method is computationally heavy, since there is never guarantee that a good hypotheses will be found, especially with consistency with other models.
	\item \textbf{Eliminating}: similar to \textit{DL} we transfer the set of functions into a DNN. It will be described further in the following.
	\item Additionally, perhaps there is some combination of these methods. For example, \textit{DL} is applied first to narrow the number of hypotheses, then the generative approach is applied somehow.
\end{enumerate}

\textbf{Related Work}

In literature, there are many code representation methods, for tasks like classification (e.g. naming the code or its purpose, decide if it is a malware, etc.) or producing a sequence similar to translation, e.g. code completion or code captioning \citep{alon2018code2seq}. Here presented some code representations and their model that is used for accomplishing some task:
\begin{itemize}
	\item syntax-based methods \citep{bielik2016phog} represent program as a bag of words/tokens. It can be processed for some task via RNN/LSTM for example or a Transformer, e.g. via BERT \citep{feng2020codebert}.
	\item code2vec method \citep{alon2019code2vec}, represents a program as a bag of all possible paths between any pair of nodes in a tree representing the program. It uses a Transformer model in order to discriminate different paths with different importance. Thus they use attention mechanism.
	\item Another more natural method to represent a program tree is via graph neural network (GNN) \citep{fernandes2018structured}.
\end{itemize}

The methods above concentrate on code representation, to accomplish some task given the code. This is discriminating AI. While our problem is generative, i.e. we need to generate code for the specific task of model predictiveness, for example as any LM: via forecasting next sensory inputs.

One approach handling this is Genetic Programming (GP) \citep{koza1994genetic, lehman2022evolution} based an evolutionary approach. It works similarly to GA. For example the population is a vector of numbers, where each number is interpreted as a code to select from a list of production rules to constructing a function. These production rules represent correct syntax or feasible expansion of a code tree. 
Here is an example:
\begin{enumerate}
	\item $\text{S} \rightarrow \text{bool}$		
	\item $\text{S} \rightarrow \text{exp}$		
	\item $\text{bool} \rightarrow \text{and bool bool}$
	\item $\text{bool} \rightarrow \text{not bool}$		
	\item $\text{exp} \rightarrow \text{ife bool exp exp}$	
	\item $\text{arg} \rightarrow \text{while bool exp}$
	\item $\text{exp} \rightarrow \text{mul exp exp}$		
	\item $\text{arg} \rightarrow \text{input}$
\end{enumerate}

where $\text{arg}$ represents any expression or argument, starting at the bottom as the input to the program. Then the right side of each rule (the terminal) is the function and the number of its arguments. We could produce initial population of vectors step-by-step, till the program is complete. For example, assuming maximal depth of a tree to be 3, and a Depth-First-Search method:
\begin{enumerate}
	\item 1: $and(,)$, depth=1
	\item 4: $and(while(,),)$, depth=2
	\item 2: $and(while(not(input),),)$, depth=3
	\item 3: $and(while(not(input),),)$, depth=3
	\item $\text{arg} \rightarrow \text{input 0}$
\end{enumerate}

\section{Related/Prior Work} \label{sec:PriorWork}


\subsection{Comparison to other cognitive architectures} \label{sec:Comparison}

NOTE: This sub-section is also under construction, hence can be skipped.

\begin{figure}[H]%
	\centering
	\subfigure[Abstraction capabilities]{ 
		\label{fig:a}%
		\includegraphics[width=0.3\textwidth]{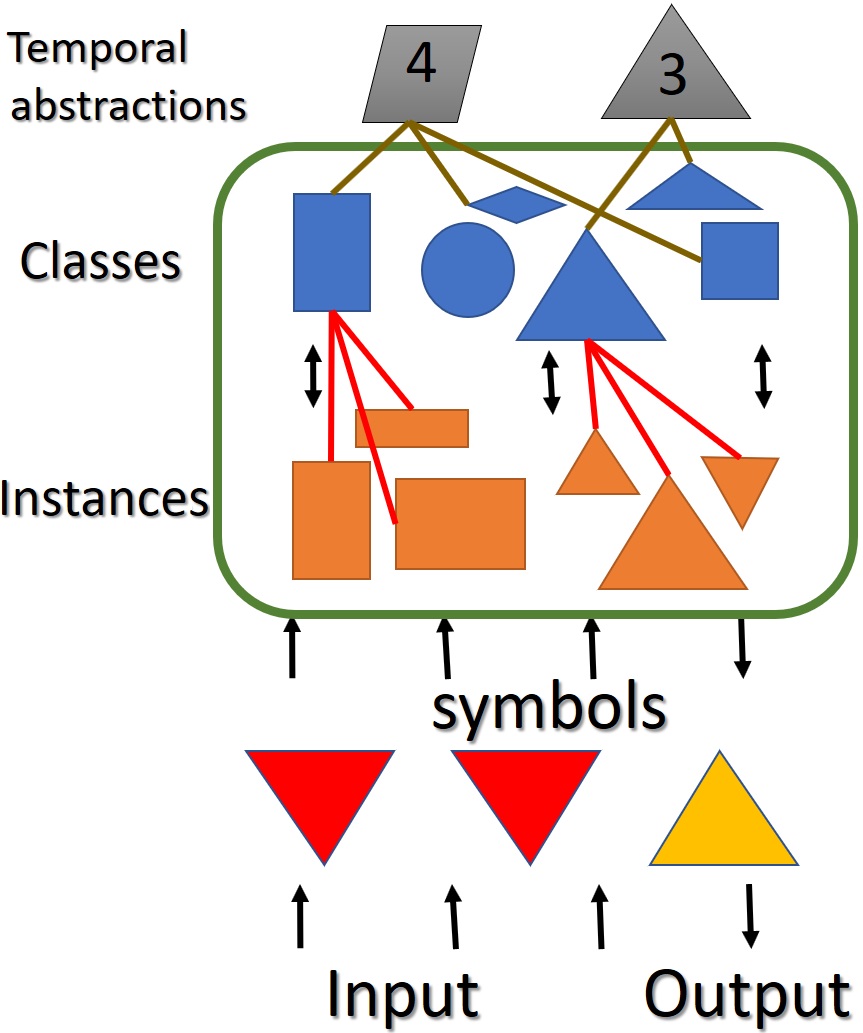}}%
	\hfill
	\subfigure[Main components in \textit{MOM}]{%
		\label{fig:b}%
		\includegraphics[width=0.65\textwidth]{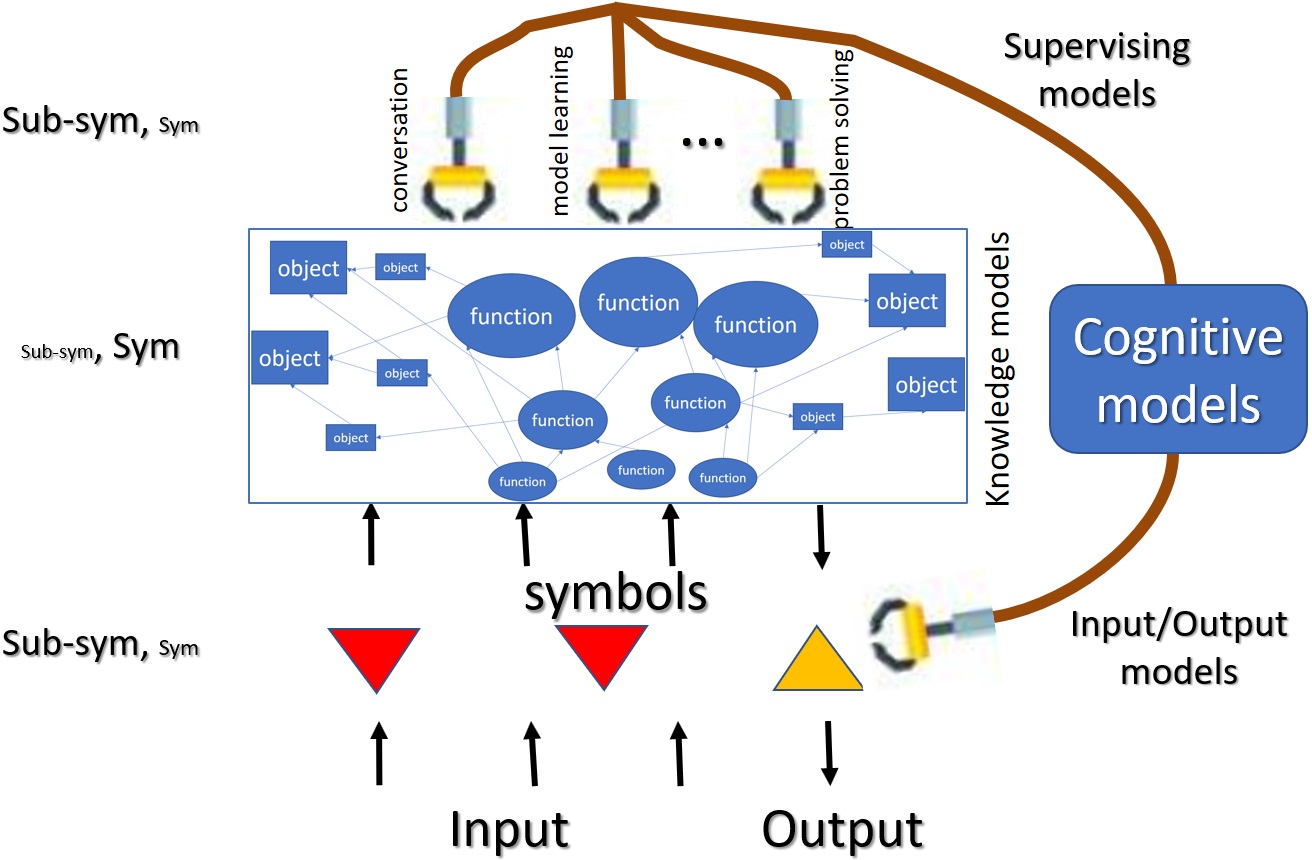}}%
	\caption{Overview of \textit{MOM} in different aspects}
	\label{fig:MOM_Overview}
\end{figure}

The first important difference between \textit{MOM} and other CAs, is that \textit{MOM} does not have a strict block diagram separating different modules (modularity) of different functionalities. Instead we embrace holistic approach, setting all different functionalities in one place. For example, the knowledge models include different memories and cognitive operations such as abstraction. One example is that we do not separate semantic and episodic memories. We include both events and the concepts they are built upon in one knowledge map\footnote{Note that implicit memory is also partially represented in the knowledge map. For example, tying shoes or driving a bicycle is a physical activity that its meaning is stored symbolically, while the physical memory of its implementation is stored in the nervous system and in muscle memory.}. The only separation for now is that of knowledge models, supervising models and I/O models. It can be regarded as layered structure, where the I/O layer is the most external, then the knowledge layer, and finally the most internal layer is the supervising one. It can be viewed in Fig.~\ref{fig:MOM_Overview}. We can see that potentially, any model can be either of symbolic type (interpretable) or of sub-symbolic type (uninterpretable). However, each layer has some dominant type.

Also, mention about logic general explainability, verse narrow specific type (structured) of explainability tools in current \textit{DL}: SHAP values, LIME, gradient heat-maps, etc. Most of which simply represent feature relative significance. This is more of an analyzing tool than a real explainability. Some say similarly about attention mechanism in Transformers \citep{jain2019attention}.

Real explainability is logical, where like a step-by-step reasoning, there is a model-based narrative guiding the listener with the relevant elements, that eventually construct a full-proof model to explain the action/decision of an AI agent, as it is in humans.
Hence, \textit{MOM} strives to deal with this issue generally.

Will is expressed both internally, to handle a given situation, such as how to react? E.g. how to solve some given situation/problem. Either immediately, e.g. via instinct, or in a delay, as a plan.
This reactive mind state is both for passive will and for active one, i.e. in either way it’s either react or simply act.
And it is expressed externally, to understand a given situation, such as comprehending a message/story/phenomena..
In both cases (my/others will), which is about control, what precedes it, is making-sense/understanding will. I.e., internally organizing models from past experiences, and externally – understanding a given state using these models.

Traditional behaviorists speak of controlling variables (i.e., variables that control the animals behavior), meaning input$\rightarrow$behavior$\rightarrow$output, or stimulus$\rightarrow$response. 
whereas control theorists speak of controlled variables (i.e., variables that the animal attempts to control). That is, animal behavioral output functions to reduce the discrepancy between the perceived actual state and the reference desired state. Thus, the function of the behavior is the control of perception, rather than the stimulus causing the behavior.
In Traditional humans are mechanistic, like billiard balls. In the control, they are purposeful, agentic creatures. 
I partly disagree: Traditional=open-loop, Control=closed-loop or feedback. Still in both we learn models. And in both we have will. It’s simply a matter of direct or indirect interaction with the environment.
In the diagram you see opposite causality of behavior. Including will can implement both cases, since we deal with all sorts of interaction, hence also various learning regimes, as we showed in RL case. The left diagram is more general, while the right one is only in problem-solving case. As we’ve seen – there’s also designing, etc.
There’s also emotion as indicator of alignment with will. E.g. in learning: as RL – after a good/bad result/reward which comes after an action, and at perception: when the state is recognized as good/bad – the optimal/appropriate action is selected. Example: “when a child sees a lollipop and has a positive emotional response based on past experiences eating a lollipop, the anticipated pleasure is serving an incentive function. When the child is struggling to open the wrapper and finally rips it with his teeth, shifting his experience from frustration to pleasure, then the emotional response has reinforced the behavior and he is more likely to rip it with his teeth in the future.

\begin{figure}[H]%
	\centering
	\subfigure[]{ 
		\label{fig:a}%
		\includegraphics[width=0.2\textwidth]{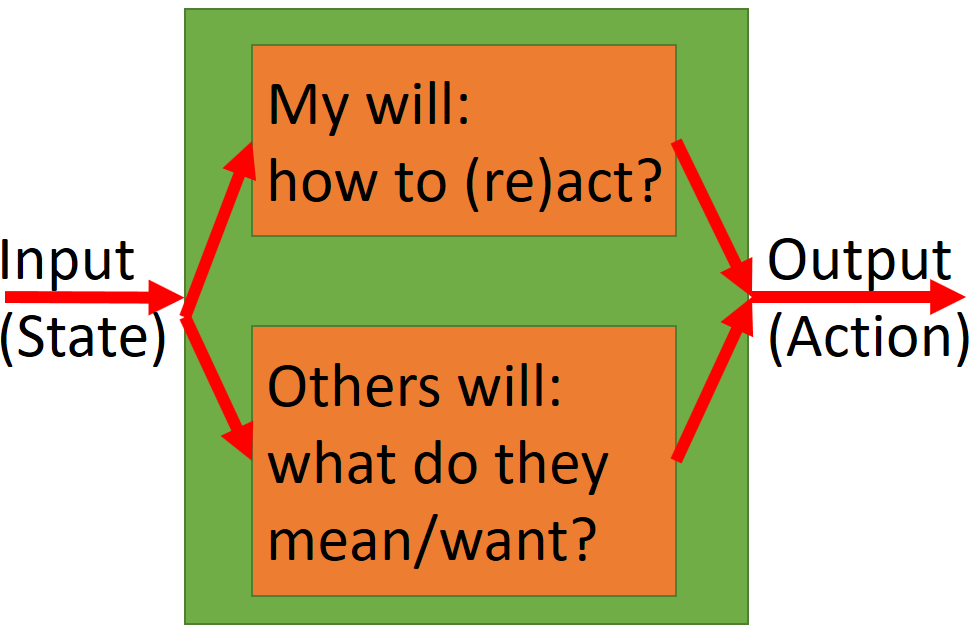}}%
	\hfill
	\subfigure[Powers model of animal and human behavior]{%
		\label{fig:b}%
		\includegraphics[width=0.75\textwidth]{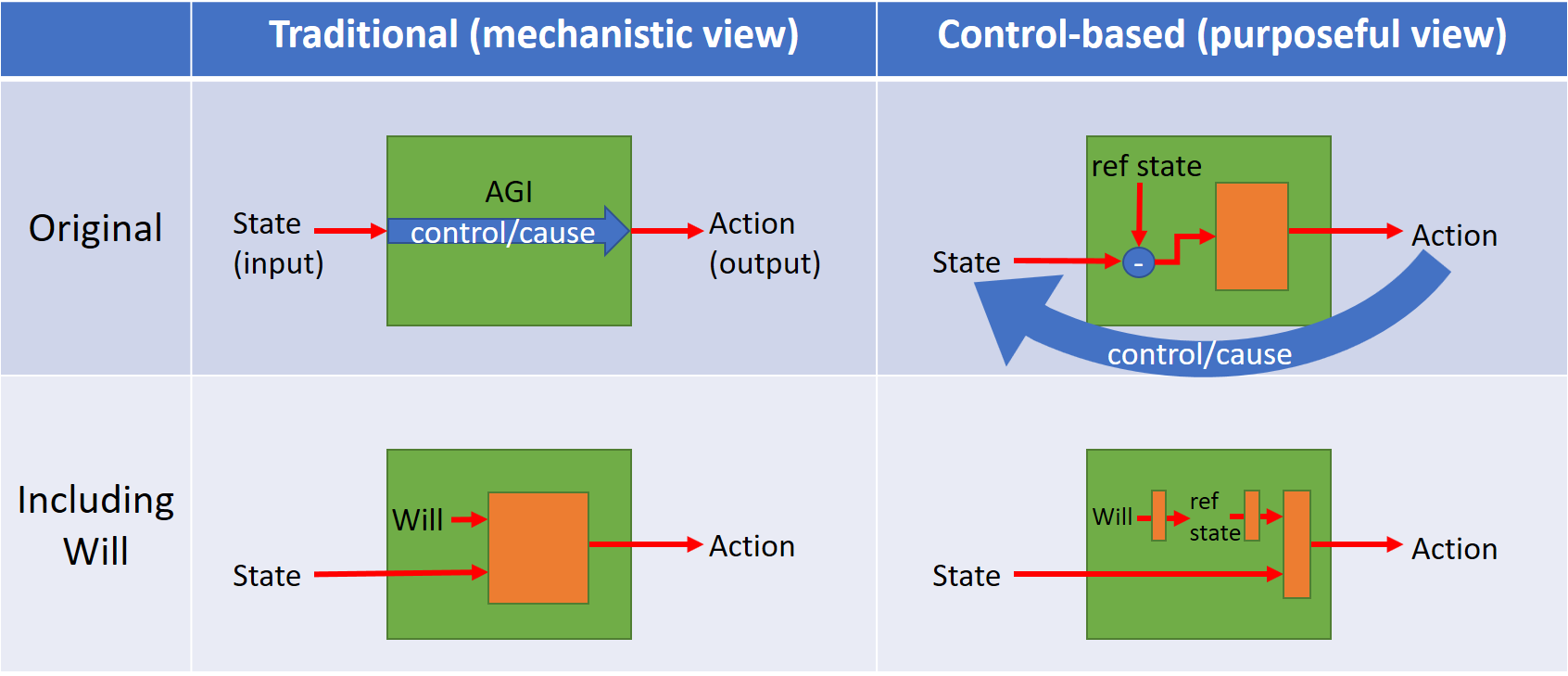}}%
	\caption{}
	\label{fig:wills}
\end{figure}

Fig.~\ref{fig:wills}.

CAs fall into four top-level categories: symbolic, emergentist, hybrid and universalist.
For example, symbolic CAs include: ACT-R, Cyc, EPIC, ICARUS, SNePS, SOAR. Emergentist CA: Hierarchical Temporal Memory (HTM) (Hawkins..2007), and DeSTIN (Arel .. 2009) is a hierarchical temporal pattern recognition architecture, with some similarities to HTM but featuring more complex learning mechanisms.

Cyc is a symbolic CA that represent knowledge via formal logic, specifically HOL. It is in contrast to knowledge graphs such as semantic web, since they represent simple binary relations, such as given a triple of SOV (subject-object-verb). It cannot represent uncertainty, and reason over distributions, though we can represent it within additional continuous features. It uses simple forward and backward inference, given a set of rules, to perform reasoning. This is highly inefficient and cause computational explosion, hence they can implement it on very simple problems.

On the other hand, Cyc is especially concerned about common sense, or the filling missing information, as \textit{MOM}. It understood that consistency is not global strict rule of knowledge, rather human knowledge can contain inconsistencies sometimes, especially moving from one context to another. Meaning they advocate a local consistency, for a specific context. 
This is what we refer to as multiple versions of the same concepts.
One important observation in Cyc, is that common sense knowledge is often not found in knowledge sources and databases, such as Wikipedia, since it is obvious. For example, that water flows downhill, that humans are mortal and cannot be dead and alive at the same time, and so on.

NARS is also store rules for everything. It also has forgetting system due to limited resources. \textit{MOM} uses both forgetting due to low use and other measures, and by encouraging abstraction to remember more general rules/objects and less specific ones. It is mostly concerned with inference from classes to new classes, i.e. performing logic learning (induction, deduction, abduction). For example induction: learning classes relationships from given data/evidence. These and other types of learning should be considered in our evolution phase. Also as in our \textit{MOM}, the inference is in runtime. In NARS the objective of learning is to organize knowledge, which is similar to our \textit{MOM}.

ACT-R uses declarative memory to store objects and their features, while it uses production rules, i.e. if-then rules, to store actions. We do not confine ourselves merely to straightforward deterministic rules, but allow for various actions, while consolidation allows the flexibility to be in the range of more options or less.
However, we do not use probabilistic representation/programming, which assign probability for each admissible action, although it is a viable option. For now it is left to the process of creativity, which dismiss the most admissible option in favor of others. Still, it is certainly a possibility, and very simple change in the current model.
Also, ACT-R is limited to a single declarative unit of knowledge while reasoning, e.g. a single perceived object. This complies with our concentrated attention. Meaning each object is perceived solely, and afterwards they can be proceed to inference.

LIDA is a hybrid CA. Like \textit{MOM}, it also emphasize the understanding and making sense goal. Then it uses GWT to compete over attention.

SOAR for example has action selection unit, that prioritize among some triggered production rules, based on the accumulated knowledge so far. This is a fast response system, to make sure it reacts in human time scale. However, as any CA, this selection is not dependent on the will, only some fraction of it, perhaps the goal, but mostly on the current state.

SOAR, like \textit{MOM}, is also emphasizes highly the limitation of resources, especially processing time over optimality of the response. One issue with rule-based system like this, is retrieval time as a function of knowledge size. Meaning, that as the knowledge increases, the time it takes to find something in it get larger, this is highly inefficient. Hence, we propose fast class recognition via usual \textit{DL}. Although the next step of inference is based on classes, which can be like rule-based also inefficient.

Additionally, SOAR like \textit{MOM}, uses emotions as rewards when applying RL, while interacting with the environment, i.e. without specified reward.
SOAR like CRAM \citep{flanagan2006control} 
and other CAs also uses the breaking-down problems to sub-problems principle.

SPAUN is an example of Neuro-Symbolic platform, that utilizes VSAs, where the sub-symbolic data is transferred into distributed vector representation, which can represent symbolic data and can be manipulated as cognitive operations.

OpenCogPrime (OCP) integrates multiple learning algorithms associated with different memory types, using a weighted labeled hyper-graph knowledge representation and making heavy use of probabilistic semantics. It is based among other things, on Cognitive Synergy Theory (CST), which includes different types of memory: declarative, procedural, sensory, episodic, attentional and intentional. In our case, attentional is perhaps separated from intentional, though they are both supposingly originated from will.
Also, as knowledge represent as hyper-graph constructed from nodes and edges, they are weighted with truth values such as probability and confidence. This is similar to NARS (frequency and confidence) and to our measures, which can be additional features of simple or complex objects. However,as previous CAs it uses production rules.




CLARION has similar structure of triggering, i.e. the neurons triggered in the sub-symbolic to symbolic section are correspond to the object/class they detect, while CLARION also trigger in opposite: from triggered symbolic class to its corresponding sub-symbolic triggered features.
Also they have bottom-up and top-down learning, however, \textit{MOM} implements both learning or recognition only unidirectionally, i.e. bottom-up, similarly in learning - the top stays top, it does not affect the bottom.
Unlike \textit{MOM}, CLARION is implementing self-will, and hence attached to meta-cognition to monitor and regulate these motives (wills). Hence, it is also explains human unique psychological phenomena. We on the other hand, dismiss self will, and do not interested in simulating human-like agent psychologically, but rather only rationally, only by including will as important component for effective communication. Subsequently, we include will as external component to be included in modeling other things outside the agent.
Action selection is according to goal in both levels, which is similar to \textit{MOM}. However, we do not separate motivational representation to explicit goals at the top and drive strengths/activations at the bottom. We strive to make things continuous and gradual. This is again, the same motive in most CAs: moduling or separating of functions, which in our opinion makes the system more rigid.
CLARION uses label nodes, node for each concept (in top), as \textit{MOM}, but their features are represented via distributed representation (at bottom) to be retrieved only by relevance to a given scenario. Again, this representation might be problematic if we to include also abstractions and assignment/specialization/instanciation.



In summary, ...

\subsection{Knowledge Graph Reasoning (KGR)} \label{sec:KGR}

There are different types of reasoning approaches in knowledge graphs \citep{tian2022knowledge}. Mainly logic rules-based, representation-based, and NN-based methods.
KG reasoning is the task of complete missing elements in KG by inference. However, the original KG has other components as well: KR learning (KRL), knowledge storage (KS), KG construction
(KGC), knowledge updating (KU), and knowledge reasoning. 
KGC includes knowledge extraction (KE), knowledge fusion (KF), and knowledge processing (KP).

KS proposes storing mostly via relational or graph data structures.
KE is about extracting objects, relations or both in the same time, mostly by utilizing pre-trained DNNs.
KF usually discover entities that represent the same semantics in different KGs, and then eliminates the ambiguity of entities from different sources.
KP mostly concentrates on relations between objects. KU is about how to update objects, relations, triples of these, and more generally how to combine new knowledge with old one.

\begin{figure}[H]
	\centering
	\includegraphics[width=0.99\textwidth]{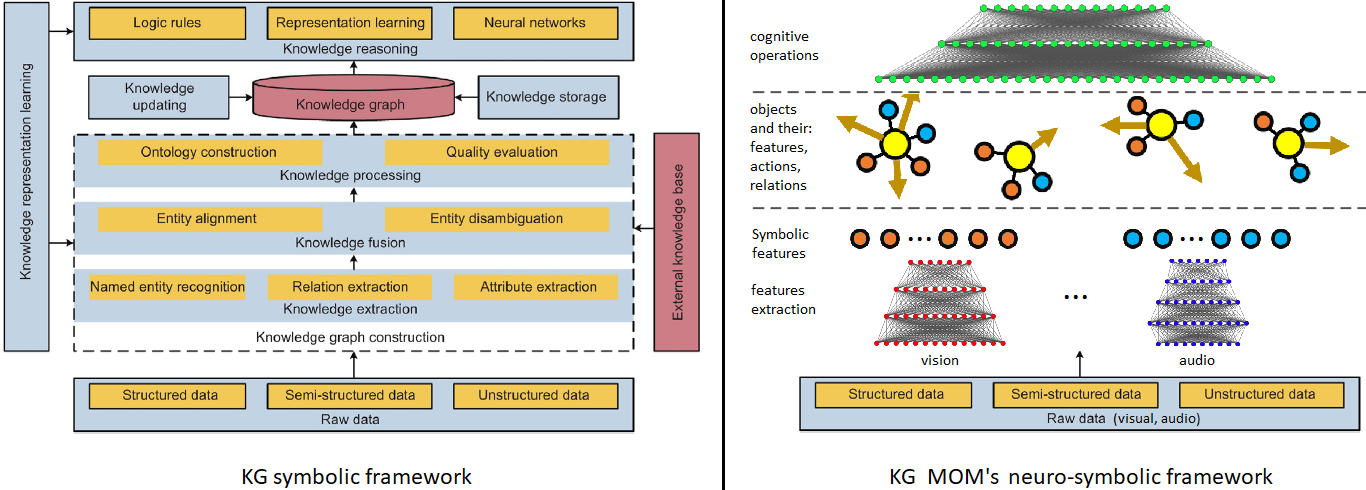}
	\caption{Comparison KG taken from \citep{tian2022knowledge} and our \textit{MOM}.}
	\label{fig:KG_and_MOM}
\end{figure}

However, \textit{MOM} is hybrid neuro-symbolic model, that learns mostly as NNs and some other principles, such as consolidation and top-down learning. Moreover, most of the KG construction processes are neuronal and not via logical or symbolic processing, especially not from a pre-designed KB, but directly from raw data. Also, although most of KE processed via DNNs, it is based on supervised learning from given datasets. See the comparison of these approaches in Fig.~\ref{fig:KG_and_MOM}.

Logic rules-based KGR use rules and KG features to infer new facts, e.g via first-order logic (FOL) pre-defined rules, or via ML extracted rules, and via paths between entities in a graph structure.

KR representation-based method compute semantic relationships among entities through projecting the semantic information (such as triples) into a dense low-dimensional vector space. There are three main approaches in this method. Tensor decomposition-based approach decompose a relation tensor of KG into multiple matrices, which are used to construct a low-dimensional embedding of KG. The distance approach learns distributed representations of all entities and relation types by vector addition of all relation triplets, see TransE \citep{bordes2013translating}. And semantic matching model measures scores of relation triples by comparing them with other entities and relation types in the low-dimensional space. These comparisons evaluate the similarity among the relations in KG.

KR NN-based methods use different DNNs (CNN, RNN, graph neural network (GNN), and deep reinforcement learning (DRL)) to produce new facts by processing the graph structure of KG.


\subsection{Comparison to previous versions of \textit{MOM}} \label{sec:Cognitive_model_comparison}

\begin{figure}[H]
	\centering
	\includegraphics[width=0.95\textwidth]{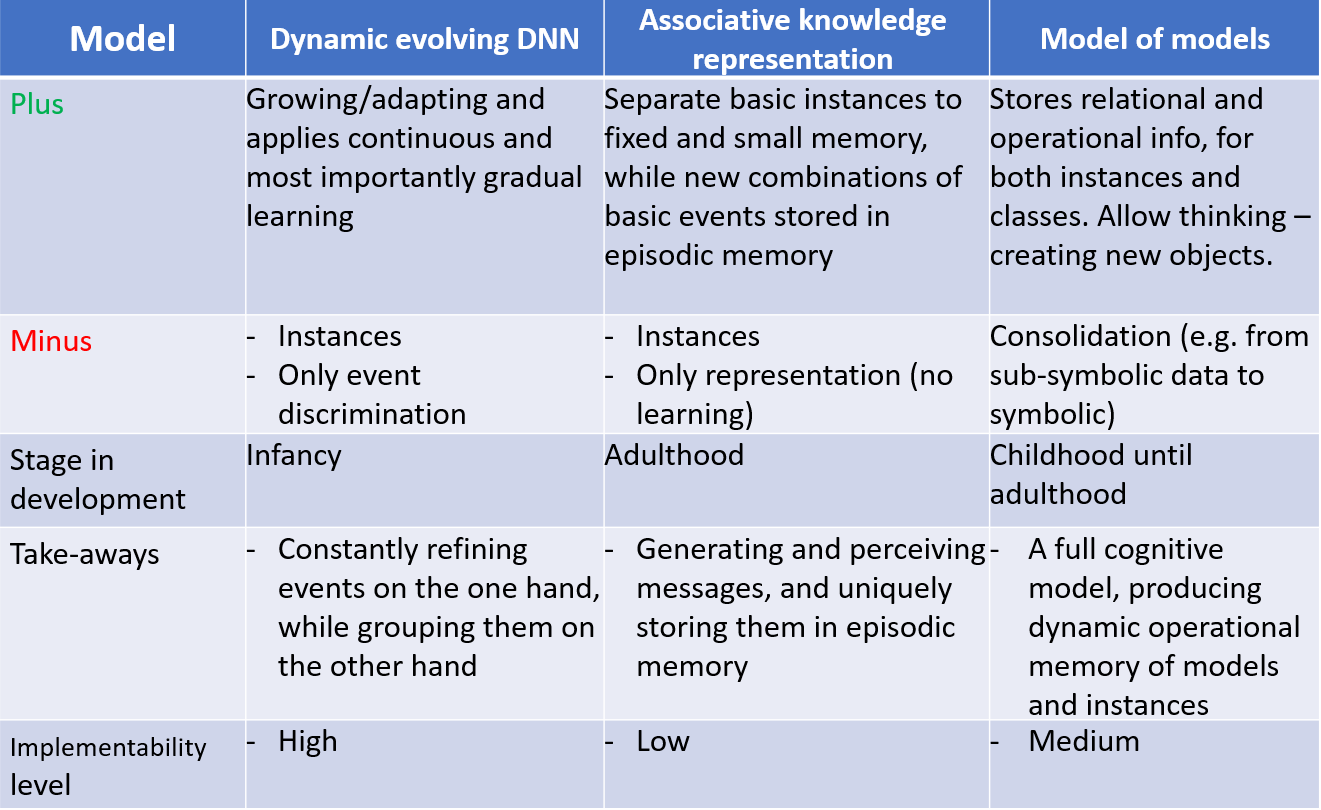}
	\caption{Comparison table of models discussed so far}
	\label{fig:comparison_table}
\end{figure}

Here all models developed so far are compared, in Fig.~\ref{fig:comparison_table}.

In summary, \textit{DENN} stores each new combination of events, while \textit{AKREM} stores dynamically in episodic memory any newly encountered combination of basic events. Basic events are stored separately in two types of memory: concepts and actions. \textit{MOM} on the other hand, unites those separate memories into one dynamic operational memory, consisting of concepts, actions, relations and any instance of those.

\section{Conclusion} \label{sec:Conclusion}

The following is the summary of the paper including some key takeaways.

First, a new cognitive model is introduced to join the existing CAs, in the form of dynamic operational memory of models and instances. In it, a holistic approach is embraced, assuming that intelligence should be highly versatile and diverse. As opposed to "picking one side", which makes it not fully representative and unbalanced, since it is inclined to one or few directions only.

Next, our model includes will as an essential part of the modeling process. Thus, in case of its absence, it turns most of the learned models to be very partial.
In the \textit{OOP} formulation, the will is an additional variable/attribute, and it is mostly significant in the top level while it is least significant in the lowest one. Moreover, our assumption is that will is derived from states or the representation of the world, via different channels (emotions, rule systems), and it affects the actions of our accumulated knowledge, by learning to align with them.

Next, operationability turns static knowledge representation into a dynamic one, thus enabling cognitive processes. The actions are learned via regular program search, via either \textit{DL} or other tools, in a self-supervised manner.

Next, one way to ensure that the continual learning is consistent, is by implementing local learning, i.e., concentrating on updating only one/few model(s), while confirming compatibility with other models.
A model can be learned either from examples or directly by using existing operations (logically).
Another way to ensure that learning is consistent, is by a slow process of consolidation. It is ensured by maintaining a high level of flexibility over a long period, while pursuing more and more consistency within and between models.

Next, reusability is utilized to enhance connectivity between models, instead of learning them as separate entities.

Finally, the cognitive model is designed via inverse engineering. Meaning, starting from our highly aware and mature cognitive state of mind, and then tracking back in time to study its evolution.

\subsection{Future Work} \label{sec:future_work}
The main problem is how to implement a cognitive system, that produces the appropriate models, i.e. how grouping/clustering occur, to generate the right models. Also, how these models produce new ones by the correct compositions.

In addition, there is the issue of how sub-symbolic become symbolic. Perhaps the models produce objects and actions directly upon sub-symbolic data. Furthermore, if continuing this line of thought, then models may be removed at all, which converts this problem to pure \textit{DL}-based approach.

Additionally, relevant to the last issue, is model evolving. Is it model refinement of some main model to sub-models, which controls how models of knowledge are used? Or is it that all the models are separate? And if it is by refinement, is it one large \textit{DL}-based model, and all the rest are knowledge models? Or if we continue this line of thought - again, end up with pure \textit{DL}-based one huge model, containing implicitly all the different models, their actions and attributes. But then how \textit{elements} and abstraction are implemented in such a \textit{DL} model? More about the suggestion above see Section~\ref{sec:ModelSeparation}.

These two examples ending up with \textit{DL} pure approach (without logic representation), may imply that this theory can be utilized as a prior knowledge in \textit{DL} models, without the inclusion of explicit logic in the final representation. This is certainly viable/legitimate idea to pursue.

Also, though hierarchy by abstraction can be implemented, but then how can it be implemented by will, 
i.e. how to decide when and what to group in such a hierarchy?


Other \textit{MOM}'s challenges include:
\begin{enumerate}
	\item How to learn concepts and their features, or more generally classes.
	\item How to learn different wills, to be utilized for action alignment.
\end{enumerate}

The second issue also about how to extract will from the perceived state. Or it is more known as Symbol Grounding Problem: how the symbolic representation acquire meaning or more generally attach will to it.

These are all open questions to deal with.

P.S.: Neuro-Symbolic AI for us is merely taking inspiration of class-based structure, to act as the final stage of learning, while \textit{DL} is the main tool to reach it.
So it is all about flexibility of DNNs, only that we use consolidation to finally reach symbols.
Another implication of this, is memory. First, since it is vague and mostly reconstructed during replay (recalling). It means it is not recorded accurately. And second, since it is probably used in sleeping periods similarly in a vague/foggy form.

\appendix
\section{Appendix} \label{sec:appendix}

\subsection{Hybrid approach}  \label{sec:hybrid}

It is our opinion, that DNN represents a very basic computational model, comprised either from a sequence of (non-linear) operations or from a sequence of "if" operations. This limitation might be the reason for the counter-intuitive relations found when analyzing DNNs' interpretability. I.e., the effect of DNN finding wrong rules, is probably due to:
\begin{itemize}
	\item the set of restricted primitives (ifs only), though DNNs are general enough to implement if-else structure, since to produce specific output - some of the neurons must be not triggered (inhibited).
	\item no relation to other relevant models, to assist the DNN in choosing most appropriate rules consistent with the other models.\footnote{Actually, it is not entirely correct, since DNNs use shared parameters which can be considered as "shared functions" that are utilized by higher layers. However, on the other hand these are local functions, temporal ones relevant only to the trained task(s).}
\end{itemize}

One advantage of DNN's is their layered structure, which is equivalent to chain-of-thought (CoT) \citep{wei2022chain} or dynamic programming (DP) \citep{bellman2015applied}. I.e. it is a step-by-step process, which may imply perhaps why the more layers (deeper) the DNN has, the better it is, since it allows for more steps of computation.

But this is also its disadvantage: if it is not dynamic in structure, it always has the same number of cognitive steps, while in reality, each model should have different steps, and more precisely - each model can have different number of steps depending on the input (just like in an algorithm).

\begin{figure}[H]
		\centering
		\includegraphics[width=0.8\textwidth]{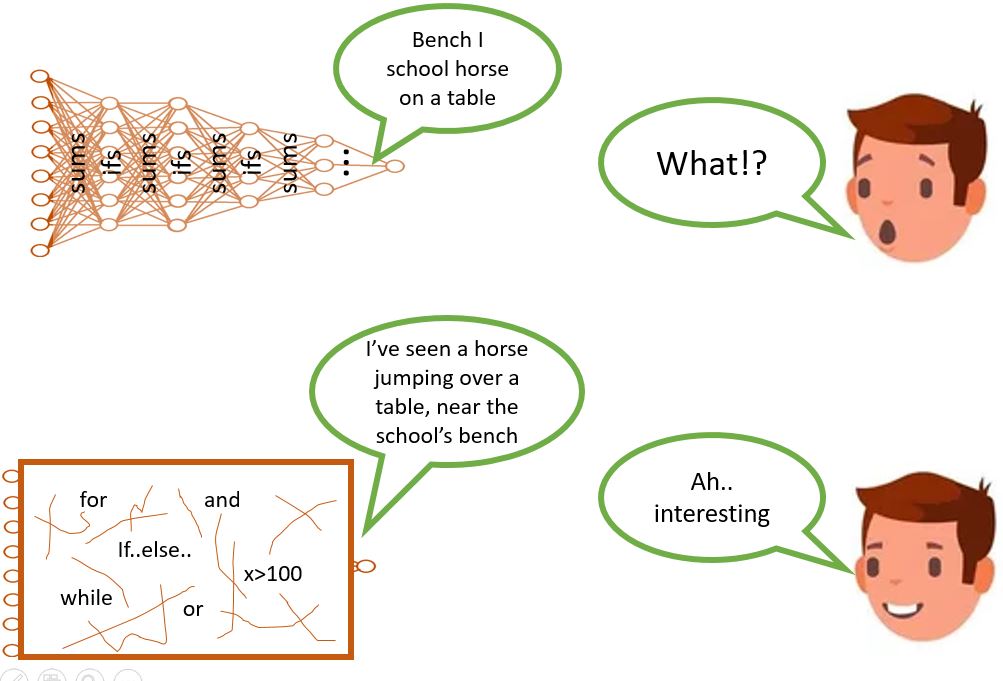}
		\caption{Caricature of interpretation in communication}
		\label{fig:caricature01}
	\end{figure}

Subsequently, we advocate what researchers like Gary Marcus \citep{marcus2003algebraic} have claimed for many years - the adoption of neuro-symbolic hybrid approach. The modeling we propose is exactly this - learnable symbolic manipulation. As the caricature in Fig.~\ref{fig:caricature01} shows, we must have symbolic explicit (and not implicit via \textit{DL}) representation of knowledge. Especially due to communicative aspects, as discussed in \citep{EasyChair:7922}.

Another shortcoming of DNNs is that the features comprising the model are unrelated to any concept, hence cannot be communicated as actions over concepts. This problem was tried to be dealt with by numerous studies, such as converting features to interpretable concepts \citep{blazek2021explainable}, the training on objects instead of pixels paper.., and dynamic evolving NN \citep{EasyChair:7922} which construct the NN neurons step-by-step by relevancy.

Another argument that \textit{DL} solely is not enough, is since we need abstraction for multiple reasons, such as analogy and other functions. One way to realize this is via symbolic manipulation as in \textit{OOP}. It is mentioned in many places, such is in \citep{Dickson} article: 
\textit{"Despite the impressive achievements of deep learning, some of the field’s problems remain unsolved. Among them are causality, compositionality, common sense, reasoning, planning, intuitive physics, and abstraction and analogy-making."}.
Though, there are studies in \textit{DL} that try to implement modularity, e.g Modular Neural Network \citep{azam2000biologically}, 
and see modular NNs in Neuro-Symbolic section/presentation...

We can view it also from a prior knowledge point-of-view: it is the trade-off between how much external prior/bias we should induce, verse how much should be learned from the data, i.e. how much of our proposed model is a prior and how much of it is flexibility and freedom to adapt. Hence, 
whatever \textit{AGI} architecture we will choose, we must be prepared to insert artificially, e.g. externally by learning, some of the prior embedded in human brain, perhaps inherited right from birth. For example, physical world priors\footnote{It may just be that physical conservation principle (object do not appear/disappear) is learned, as stated by Piaget, since it is not merely programming, but it is also descriptive (behavioral).} \citep{khatib2022learning, piloto2022intuitive}, or the self and others models, and more.
Additionally, some things will have to be guided explicitly, since \textit{AGI} do not posses a will of its own, e.g. telling it what to transfer from WM to LTM (what is important).
Moreover, some basic functionalities perhaps needed to be rule-based, i.e. written algorithmically, in the \textit{AGI}. These functionalities are probably coming from the human genome, mainly regulating the main systems or memory. Similarly to \citep{marcus2003algebraic}.

\begin{figure}[H]%
	\centering
	\subfigure[No free Lunch for highly-designed AI.]{ 
			\label{fig:a}%
			\includegraphics[width=0.5\textwidth]{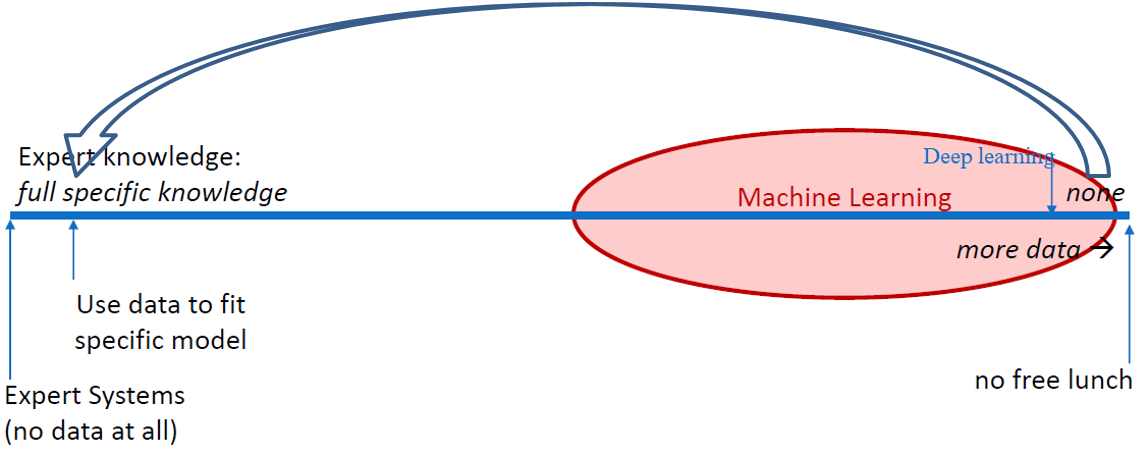}}%
		\hfill
		\subfigure[Performance verse the data scale/dimension, taken from \citep{chalapathy2019deep}.]{%
			\label{fig:b}%
			\includegraphics[width=0.45\textwidth]{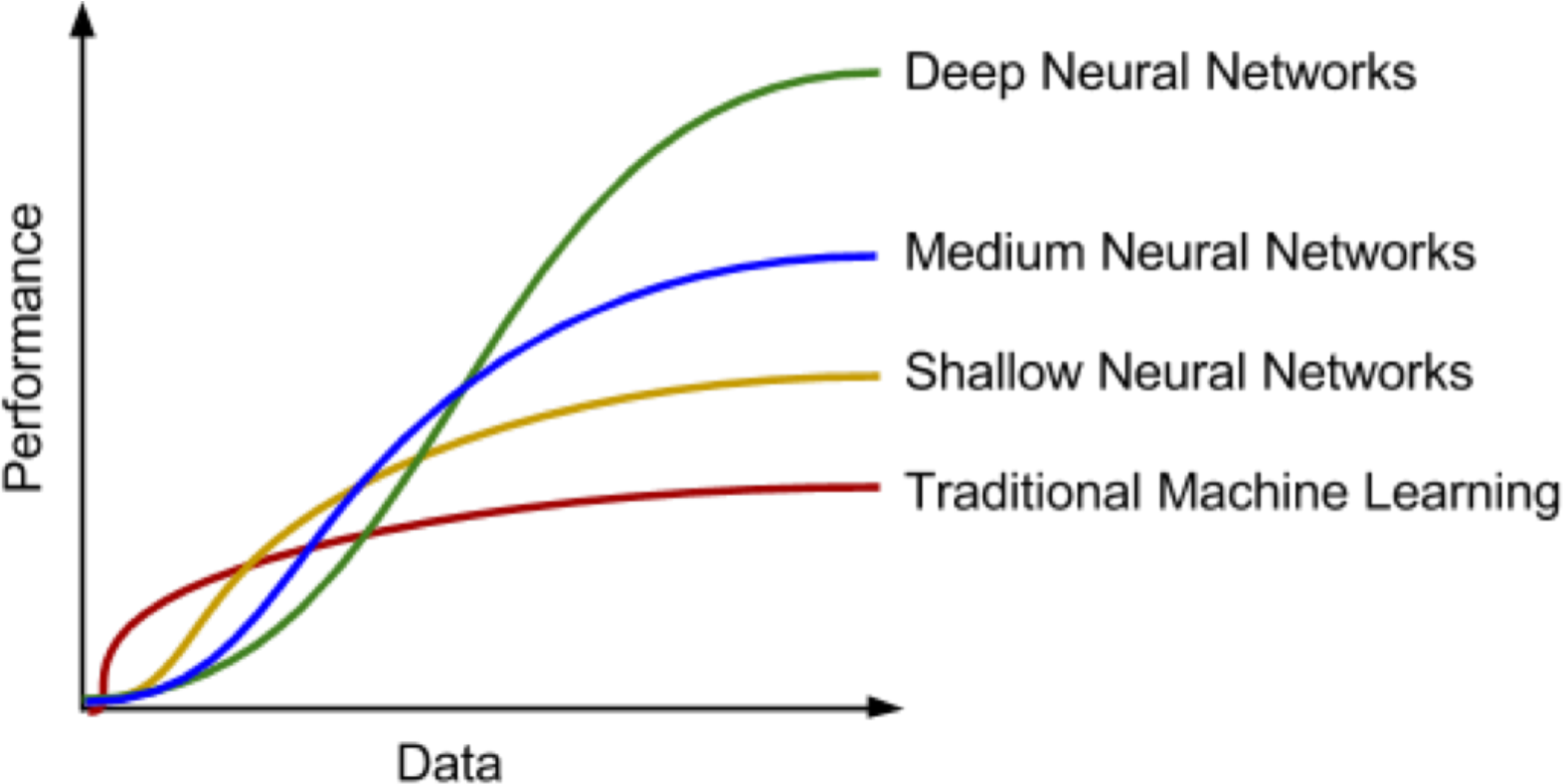}}%
		\caption{Comparison between AI methods to comprehend the environment.}
		\label{fig:AGI00}
	\end{figure}

Nowadays, Machine Learning (ML) theory assumes best performance is in \textit{DL}, see Fig.~\ref{fig:AGI00}(a).
Especially for large data set \citep{chalapathy2019deep}, see Fig.~\ref{fig:AGI00}(b), where the main effect is for big data, since then there is a difference between the models. But in small data the different sizes of NNs are all behave approximately the same.

However, in our humble opinion, unlike the place where \textit{AGI} supposed to be: in the most right part of the scale, where we almost totally dependent on data (minimal prior knowledge), it is actually the opposite - we need full knowledge for designing \textit{AGI} (lots of assumptions), which do not depend on data at all. Then the \textit{AGI} is flexible and adaptive, and is able to tune gradually and gently according to data. Whose more important here? It is not the data, but rather the assumptions and lots of planning of AI. 
This is like planning an adaptive or any control system, which is deployed only after knowing apriori all possible data/cases/scenarios there could be, encounter by the agent.

Same idea is in \citep{ioannou2018structural}: \emph{"learning ability does not come for free, and is far from automatic. It relies on very specific assumptions that are mostly encoded in the design of the NN architecture itself - here denoted as structural priors."}.

	Moreover, DNNs are not that far from symbolic manipulation, since the input is perhaps sub-symbolic, but the actions on it are pure logic: if, sum, activation function, etc.
	
	We see in the figure that DNN has very strict and rigid structure, which of course allows it to efficiently learn, but on the other hand it denies from it the freedom to model the actual relations and interactions in the data to represent it in its correct form/model. Handing the learning module more freedom (higher variance), might have larger search space for the best model, but eventually, after enough time, it will settle on the most appropriate one.
	
	This is the exact argument against the \textit{DL} immediate-results perspective \citep{Ball2022,de2016parallel}. 
	First, because there is no gradual learning in the complexity of data, as a school student experiences throughout his school years. And second, it takes years for an infant to acquire even simple skills, while in \textit{DL} it is expected to do the same and more (the performance is compared to adults), in a scale of days.

	Note, that Fig.~\ref{fig:caricature01} is not about the issue of explainability of DNNs, but rather it is about the inappropriate structure to represent the correct model, to enable correct reasoning and communication.
	Although, when learning the correct model, it is also explainable in its inner structure. But more importantly, it is explainable by communicating with.
	
	Also, due to the gap between the set of hypotheses or models that a DNN can produce and the ideal model that represent some trained data, you can notice the shortcuts that the DNN produces to learn the data, such as predicting a cow in an image, if it spots a sky. I.e. in this case it has nothing to do with a cow or its features. Similar effect, of not being able to generalize in LLMs performing multi-step composite tasks. Doing this also by shortcuts, or more specifically by pattern matching and subgraph matching, instead of comprehending the tasks \citep{dziri2023faith}. 
	
	This effect is reasonable, since there are no constraints or bias over how the DNN should arrive to truly representing approximated function. It is because there is a huge amount of suitable functions to map a limited amount of inputs to their corresponding outputs, see also in \citep{guest2023logical}. 
	This huge amount of potential functions explains over-fitting in \textit{DL}, since many possible functions can represent the training data without any respect to the general data and knowledge in the world.

	It can be solved in a system, where different models are not learned in isolation, and it learns a model with alignment and consistency-check with other models learned so far, which \textit{MOM} supplies (continual learning). That is, the learning performed in interaction and feedback from other models, in order for each of them to be maximally correct. I.e., every model should be learned without conflicts with other models, but rather in cooperation with them, which enforces each model to be aligned to its true functions.

	Moreover, any ML system has to be provided with a well-defined objective, which means that the designer needs to be very punctilious about his objective and the constraints of the problem. All this rigorous process is redundant if we use natural language, since we assume a minimalistic form of communication, where very little is said, and most of it is completed by previous knowledge and context.
	
	\subsubsection{Neuro-Symbolic AI in literature}
	
	Neuro-Symbolic AI is widely studied in ML community. For example, the Neuro-Symbolic Concept Learner for Visual-Question-Answering tasks \citep{mao2019neuro}, which is an object-based representation for a scene, and run a semantic parsing module to translate a question into executable and symbolic program, given a domain specific language (DSL), and executes it on object features to get the answer. They use (image, question, answer) tuples for supervised training. The DSL contains a set of operations over objects' features or concepts in the question. These concepts are neural operations, learned in training. Hence, the program executor is quasi-symbolic, and the program is a nested sequence of operations, e.g. \emph{Query(Shape, Filter(Red, Relate(Left, Filter(Sphere))))}. The training is end-to-end excluding the visual component that produces some embedding, to be operated upon.
	
	Differently, \citep{yi2018neural} presents purely symbolic executor, since instead of using  sub-symbolic features, it uses a scene parser that de-renders the image to obtain a structural scene representation, i.e. generate symbolic features of the objects, e.g. their size, shape, material and color categories. And they are used as parameters in program execution. Here the training is partitioned to object detection with the categories above, and then the symbolic part is trained via supervised learning on question-program tuples following with RL REINFORCE algorithm for question-answer tuples, where the answer is the reward. 
	
	It raises the question, whether we should convert the sub-symbolic to symbolic \citep{yi2018neural}, or leave it as sub-symbolic and operate symbolically on it, as in \citep{mao2019neuro}.
	
	In \citep{verma2018programmatically}, they use deep RL (DRL) to learn a neural policy network first and then do local search on programmatic policies to minimize the distance between policy generated by DRL and programmatic policy. Thus generating an interpretable policy. They claim that they perform this conversion, since program cannot be learned directly, since it is not differentiable. But we propose to find differentiable type of program learning, see Section~\ref{sec:FP}.

\subsection{How will is learned at the beginning}

Inverse reinforcement learning (IRL) \citep{zhifei2012survey} or imitation learning is an RL variation focusing on learning a reward function from human feedback. We argue that just as mirror neurons discovered by neurosciencetists, mimicking other peoples' behavior, similarly IRL and perceiving a story or a message from a sender (as described in \textit{AKREM}) is all about constructing the actual intention from the observed behavior. Hence infants, due to lack of developed memory must use visual (besides audio) input as a "story-telling" channel, to learn intentions of objects, either physical or living (e.g. animals and humans).

Mimicking is quite central in early life, such as pretending games \citep{wolf2022implications}, 
where other perspectives are learned for social interaction. Though in early childhood it may also be unrealistic perspective, or more imaginary.

However, the more fundamental question is whether will can be externally learned or it is embedded internally, inherently, or may be both.

\subsection{Knowledge representations}

Here we survey some common knowledge representations, and inspect them in comparison to \textit{MOM}.

%
%

\subsubsection{Scene Graphs and Knowledge graphs}

In AI, knowledge is expressed usually by objects, relations and attributes, via \emph{Knowledge graphs} (KG), and in \textit{DL} it is often realized via \emph{Scene Graphs} (SG) \citep{agarwal2020visual, zhu2022scene}.

SGs are uniquely describe a given image, while KGs are expressing general knowledge.

One of the SGs limitations, is that it produces very large network of too much details. It cannot distinguish between the important and the minor, between the essence and the details.

Sometimes, KGs and SGs combined together, in VQA tasks for example.

Anyway, we assume that memory data should not be in the form of merely KG or SG, since they are fixed and rigid, while we need more flexibility. In thinking we simulate different and new scenarios by applying new actions on the data.

So unlike static representations of knowledge, i.e. static maps, operational representation enables the dynamic structure of this knowledge, by applying admissible actions in the knowledge structure.

\subsubsection{OPM}

Another approach, known as Object-Process-Methodology (OPM) \citep{dori2011object}, is a conceptual modeling of systems. It contains two types of elements: objects, processes and links between them. Processes represent activity, action, and procedure.

One of OPM's limitations: it is locally designed, with arbitrary names for objects/processes, while it should be universal (like John Ball's Super KG \url{https://medium.com/pat-inc/how-semantics-enables-super-knowledge-graphs-part-1-bda3c4a4386c}). Also: Many conceptual discriminations/categorizations should be learned and not pre-defined. E.g. physical verse logical (temporal verse non-temporal), environmental verse systematic, which actually should be in a continuum, because there are several degrees of relevance to the discussed object. Also types of relations: structural verse procedural (should be learned). Or the separation of processes and relations. We consider all as actions, though we should have primitive actions to start with. We, unlike OPM, do not separate into different categories, since we cannot predict if there will not be others also along the way. Only the basic atoms of knowledge should remain: objects, attributes, and actions, all else should be learned.

Additionally, OPM based on physical reality, while \textit{AGI} is more general - it allows also abstract concepts. Hence objects could be non-spatial, and "processes" can be a private case of actions, which generally are not necessary temporal, but could be sometimes.

Moreover, OPM is largely based on previously presented representations (e.g. semantic networks). 

Furthermore, the state changing is simple assigning operation. They also distinct main object/process/function from secondary ones (like systemic verse environmental). We have it as additional attribute, of an action/object, about how relevant is one to the other, i.e. action admissibility. It is very important as prioritization among different options in the process of association. It can be the result of consolidation, where some are consolidated strongly, while others weakly. 




OPM can also model spatial data, e.g. an architecture of an apartment or a room. Conversely, we use semantic-like operators (above/below, left/right, up/down, next to, in-front-of/behind) with relative strength to model spatial data. Although \textit{DL} treats semantics as being represented in continuous space, via embedding.


Finally, OPM's zooming-in diagrams support limited WM hypothesis (or limited cognitive load) \citep{dori2011object}, 
as in \textit{AKREM} and \textit{MOM}.



\subsubsection{SciSoft model}  \label{sec:SciSoft}

This is operational representation of knowledge. It is about representing scientific theories and models, such as classic mechanics or electro-magnetism in physics. It was designed specifically to present equation development in these theories. The main motivation is to make the data highly connected, as in Wikipedia and other web-based platforms, to enable maximal understanding of any topic. It is done specifically by connecting a topic to all its relevant concepts so that nothing would be missing from the user to understand the topic to its fullest.

The final product of this idea is a knowledge base, represented by a graph, containing various types of objects: animations, images, text, mathematical expressions and visualizations (as graphs and shapes). All objects are connected by relevancy, but also have attributes, such that no piece of data, no matter how small it is, will be missing. For example, each equation should have access to all its components (e.g. variables, parameters and operations), with each component as an object have its attributes and connections.

The necessity in such detailed system comes from failures in teaching scientific subjects (from the teacher) or difficulties from the student side. The root of all these problems is our language. It is highly context-based and minimalistic, i.e. like an efficient code, we use minimal set of words to describe something - we say much less then we should, and expect everything else to be completed in the recipient's head. But unfortunately, most of the time students lack some "obvious" details, which the teacher considers they have in their model. This fact generates misunderstanding, which generates the need for Q\&A.

All this can be avoided if we have all the details in one system, for the student to observe and bounce from one concept to another, filling any gap he has.

Note that SciSoft is obviously following entropy of information principle, i.e. when I perform operations on equations, the result cannot restore its origins, because some operations are irreversible. Hence, SciSoft is a directed graph, representing the development of equations from some fundamental assumptions or axioms. Meaning, like any operation, as in \textit{MOM}, if it is irreversible, then information is lost in the process.
For example, the operation in Fig.~\ref{fig:developing_equations}(a) is irreversible, since $2(y-3)+4=10$ become $2y-2=10$ which cannot reproduce its origins: $2x+4=10$ and $x=y-3$.

Since any specific topic or subject can have multiple theories/models, sometimes in conflict, then SciSoft enable multiple parallel versions of the same thing. Similarly, it is enabled in OPM and in \textit{MOM}.



Finally, it is noticeable that most rule-based systems, like OPM and SciSoft, construct the knowledge manually. However, \textit{AGI}, or in our case \textit{MOM}, should learn it by itself.

\subsection{Bias}  \label{sec:bias}

Bias is required in our reality. We need it for survival and fast response. It is also due to consolidation, or similarly described as the function collapse in quantum theory. See more in Section~\ref{sec:consolidation}. 
We always pick patterns in an unpatterned reality at its core. Hence, it is related to the right and left brain theory, where the left hemisphere might act as unbiasing or resetting function, to take us back to all the possible options, before we selected one specific, in every topic and model.

%

Left-brain creative part, unlike right-brain pattern-recognition system, is where we set free from patterns. It is unlimited space for imagination. Hence activating it allows using non-learned actions while thinking. For example the imagining of an elephant flying with his ears, is not learned from experience.

Similar idea should be applied in art, e.g. painting. It should not be a random and a weird mash of things, as it is so often occurs in current \textit{DL} generating tools as DALL-E \citep{reddy2021dall}. Classical art should be created via structured imagination, i.e. having some sense/logic, though it may be not physical or it can be infeasible.

Back to Bias. 
It is related to our subjectivity. We born unbiased (or slightly biased due to our genes), and as we learn models, we gain our character and mold our identity, or our uniqueness. The sum of all our models is what define us, and differentiate us from others, and generate our unique perspective and our subjective interpretation. You can see it as if every model, of something, is a superposition of all possible models it could be, while we simply collapsed into one of them.

Hence unbiasing is the act of collective thinking. Where I behave less as a unique subject, or individual, and more as a part of some collective.

We cannot unbias ourselves, but rather we can either deepen our models by incorporating more details, options, and conditions, that reduce simple prejudice view over some topic (e.g. over groups of people, i.e. racism); or, we can add more versions of the same model (which is like including the models of other people in me).

But eventually, these two options are equivalent. It is simply the question of how we update our models.

Simply put, we like to have hierarchical models to abstract details away and simplify as much as possible, since we dislike to remember lots of details. We hence strive to construct hierarchical associative models. This idea can be seen here for example, \citep{friesen2021theories}, 
claiming: \textit{"It feels good when many items come together in a simple manner. It feels bad when there is an exception to some general rule."}

\subsection{Linguistic aspect}   \label{sec:Linguistic}

\begin{figure}[H]%
	\centering
	\subfigure[Syntax issue]{%
		\label{fig:a}%
		\includegraphics[width=0.35\textwidth]{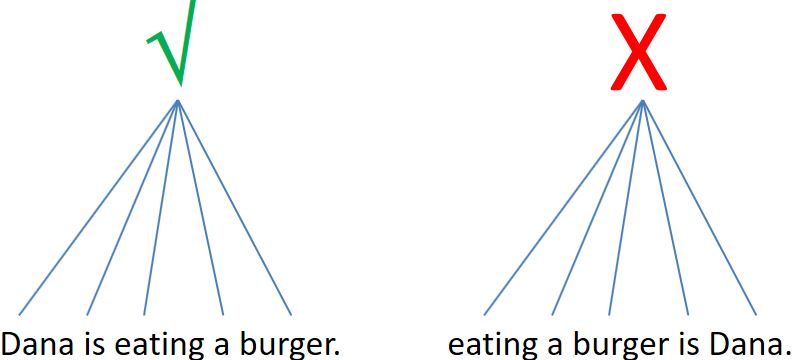}}%
	\hfill
	\subfigure[Semantics issue]{%
		\label{fig:b}%
		\includegraphics[width=0.6\textwidth]{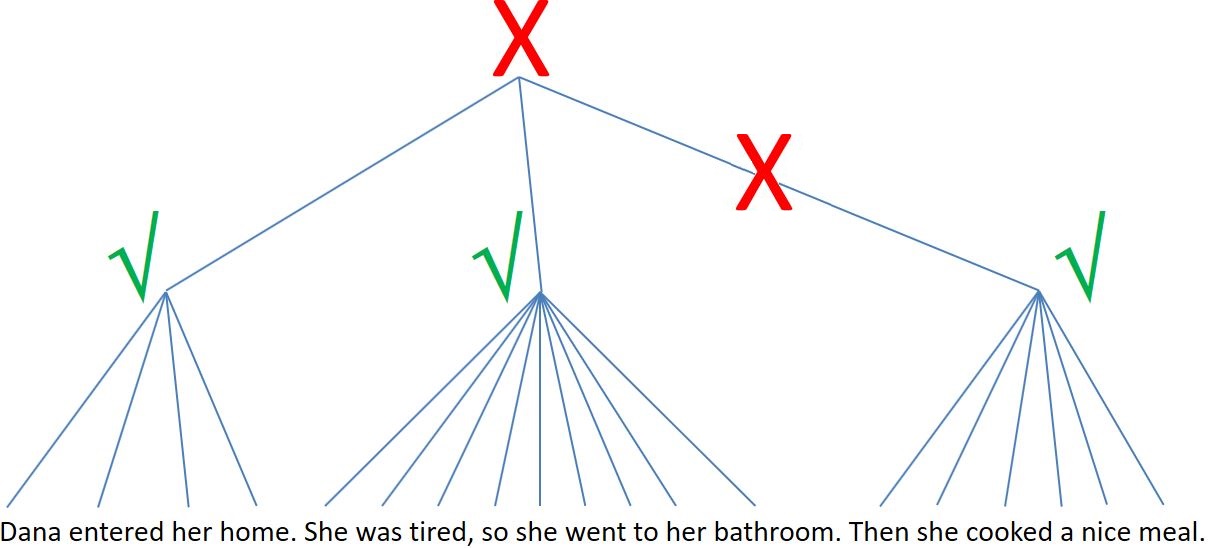}}%
	\caption{Language processing issues}
	\label{fig:Language_processing_issues}
\end{figure}

Here an advantage of \textit{AKREM} in language is proposed.
Figure~\ref{fig:Language_processing_issues} describes some issues in Natural Language Processing (NLP). Figure~\ref{fig:Language_processing_issues}(a) shows that syntax is not enough to contain also meaning (semantics), while Figure~\ref{fig:Language_processing_issues}(b) shows that local semantics or solving Coreference Resolution (finding what refers to what) is not enough to ensure logic and coherence within and between sentences.

So Figure~\ref{fig:Language_processing_issues}(b) shows that semantics is also not enough. We argue that both John Ball \citep{van2015nlp, ball2017using} and Chomsky \citep{chomsky1956three} tried to do the right thing - to construct a model from a sequential input. Syntax parsing \citep{abney1997part} only evaluate the entities themselves, while John Ball includes the relations between the entities or their roles - thus introduce the meaning of a full sentence. Though meaning reduce the number of illogical phrases and sentences, it is only local meaning. We propose to continue to check relations within and among sentences, and continue in the same manner to higher aggregating levels.
All levels in this hierarchy have purpose, just like in a military: imagine a general that wants to occupy some country to get its natural resources. He then command his officer to invade some city that is strategically blocking the access to these resources. He gives the officer the reason for the command: most of the enemy is in this city. Consequently the officer gives order to his soldiers, to attack in specific formation, while giving them the reason: given the acquired intel it is the most effective way to occupy with least casualties.

We see that every level acts with some purpose, hence the associative hierarchy is a hierarchy of will\footnote{Though in \textit{MOM} we claim it is a hierarchy by abstraction, only that it is generated by will's field, see Section~\ref{sec:summary}}.

It is a top-down modeling that constructed from the bottom, therefore while a person is hearing a story/message, he always checks for its consistency (even though the highest intention is usually revealed at the end, such as in thrillers/mysteries).

John Ball claims that syntax is combinatorial, i.e. that it is not restricted enough, so his meaning-based approach argue that it is a more restrictive model. But as it seems, this approach is not restrictive enough too. So we filter along the way, to end up with a more refined model to represent correct human language.

Ball's main statement is that semantics/meaning is not based on syntax. We claim similarly, however our meaning is beyond simple objects and relations - it also represents will in its different forms. 

Here \citep{van2021compositionality} their argument is also against the compositionality of semantics, i.e. that the meaning of a complex phrase is not derived from the meaning of its constituents, but rather from context.
We also hold for compositionality, though of a different kind.

Beyond the deficits presented above, these language models are categorizing words and relations on pre-defined classes, e.g. Verb or Actor, etc. This is another limitation, since the constructed model should be specific and appropriate to the sentence(s). Part-Of-Speech (POS) parsing deals with it simply by adding more production rules or in Ball's case enlarging the Super KG \url{https://medium.com/pat-inc/how-semantics-enables-super-knowledge-graphs-part-1-bda3c4a4386c}, both are symbolic AI approaches, which result with the famous rule-based explosion problem.

Also, John claims that language is about pattern matching or labeling, words and compositions of them (phrases). Similarly, we propose that it is about recognizing the right model. However, it is not about recognizing simple static pattern, but an operational one.

Finally, linguistics is concerned mainly about language processing/generating, but as we mentioned earlier, it cannot be separated from the full \textit{AGI} model (holisticity), e.g. language model has to consider also refining/updating knowledge models and the cognitive functions (e.g. inference).

Note, that inclusion of will on top of semantics is important also in generalization. For example, if a usual DNN learns to recognize a chair out of huge amount of chair images, it can only infer correctly on images from the same distribution. But it goes farther: even if it has seen all the possible chairs in the world, it still might be wrong when seeing a three-legged chair or a two-legged, and so on. Why? because it has not seen one. Humans, on the other hand, do not only recognize static meaningless objects, but also their purpose. In our case, that the chair is for someone to seat on it. Hence, this additional will level embedded in the concept is so important.

\subsection{DNNs' equivalence to Programming Languages} \label{sec:DNN_equivalence}

Here we will show the equivalence of different NN structures to a rule-based algorithms (or usual programming).
For example, we can see that NNs (multiple if's), decision trees (nested if-else), RNN (for loop), recursive NN (recursion) and other ML methods are all actually program-like.

Decision trees as parametric methods, are highly unstable (hence ensemble models or random forest are usually preferred), since they produce multiple solutions. In our opinion, it may be due to insufficient set of tools to express any function. They use only "if-else" blocks, as oppose to universal approximating DNNs. There is also the difference of structure: decision trees have dynamic structure (non-parametric model), while DNNs usually have fixed structure (parametric model\footnote{Parametric models have a fixed number of adaptable parameters, independent of the amount of data, while non-parametric models have a variable number of parameters, i.e. the parameters change to adapt to the amount of data.}), which constrain the solution space, for better or for worse. 

It is very hard to obtain the equivalence of DNN to programming syntax. General DNN encapsulates in it many of the programming operations mentioned above, in a declarative manner: "if-else" rules by using activation function (see in \citep{aytekin2022neural}), 
"or" rules by using sum of weighted neurons, and layers enable multiple steps of a program. RNN adds also iterative operators.

Attention is equivalent to element-wise operation, see \citep{weiss2021thinking}, i.e. matrix/vector type of manipulations, e.g. the element-wise matching of matrix/vector to matrix/vector, to perform some specific operation.

Also attention can be compared to other NN forms: unlike fully-connected layers, it changes the attended parts in the input instead of accounting for the whole input, all the time. This mechanism converts the fixed architecture of NN into a dynamic one, depending on the input, similar to hypernetwork \citep{galanti2020modularity}, where the weights determined by the input. It is also about finding relations. Whereas convolution layers \citep{li2021survey} take into account only nearby relations, hence need many layers to locate also distant relations, attention and fully-connected layers capture these distance relations more straightforward.

Moreover, transformers might be so successful in vision and language nowadays, due to these layer-after-layer self-attention, which may be equivalent to regular algorithm, that finds only specific variables from the full input set of variables, to be manipulated at each phase of calculation.


\subsection{Evolution in \textit{AKREM}}  \label{sec:AKREM_evolution}

In this section, the general learning method in \textit{AKREM} via episodic memory is presented.
It starts with defining episodic memory. Then episodic memory's interaction with LTM through waking and sleeping periods is discussed. Then, a more broader view of human development is discussed. Finally, model learning is appended to this development process.

\subsubsection{Episodic memory in \textit{AKREM}} \label{sec:Episodic_memory}

\begin{figure}[H]%
	\centering
	\subfigure[Different memory representations in some DNN.]{%
		\label{fig:a}%
		\includegraphics[width=0.47\textwidth]{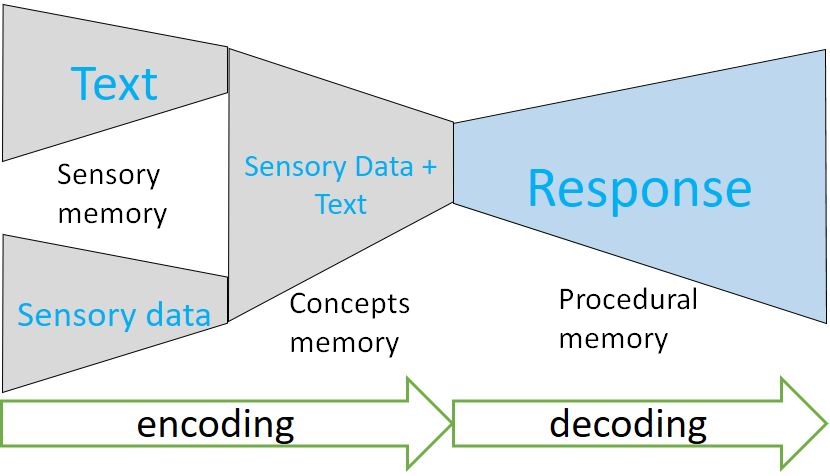}}
	\hfill
	\subfigure[Episodic memory, for storing and shaping base memories]{%
		\label{fig:b}%
		\includegraphics[width=0.47\textwidth]{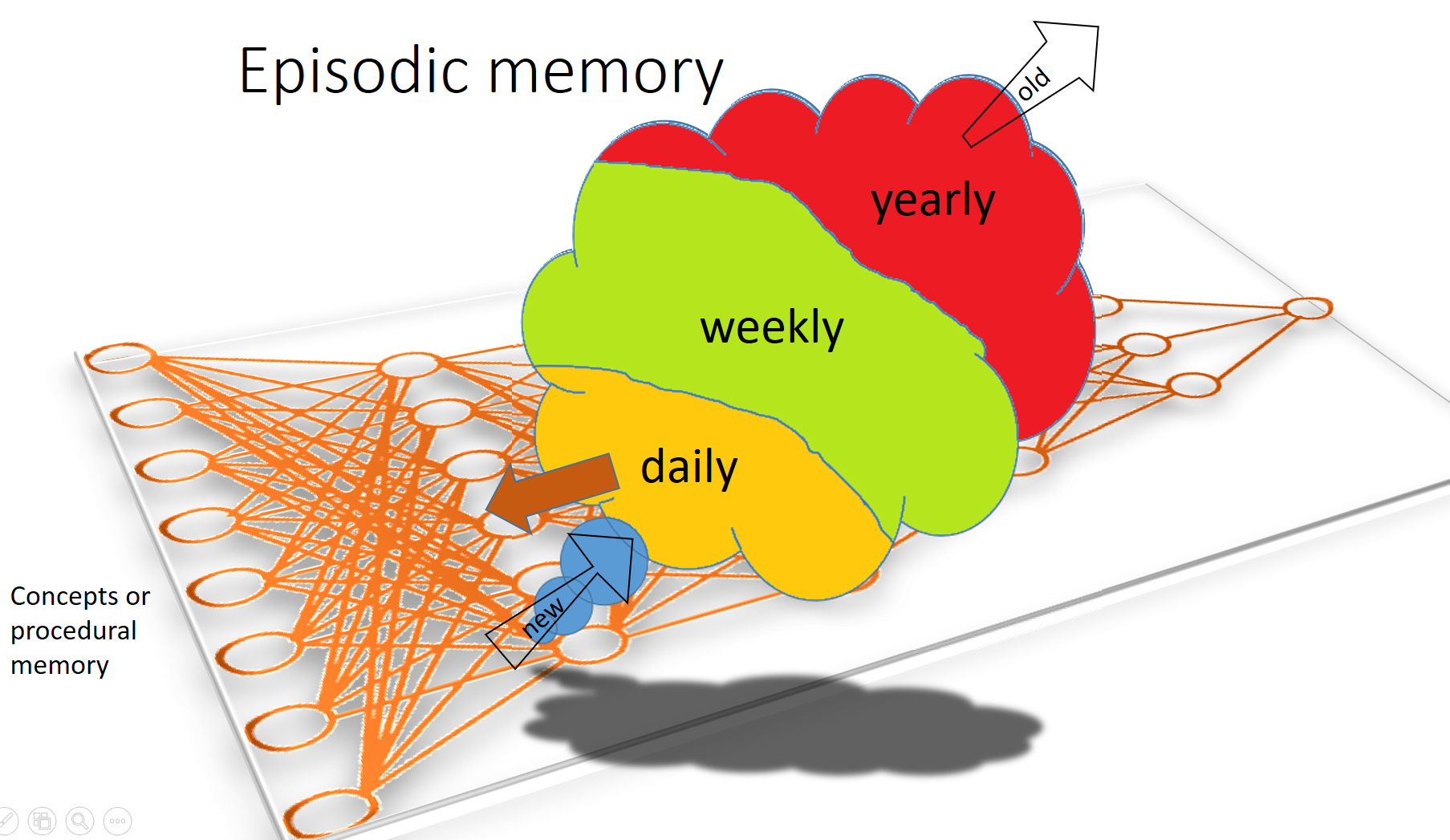}}%
	\caption{Evolution of cognition}
	\label{fig:Episodic_memory}
\end{figure}

Here some trained DNN is assumed, containing basic concepts and actions, to be treated as a base memory, upon which an episodic memory can be constructed, see Fig.~\ref{fig:Episodic_memory}(a) and similar discussion in \textit{"Issues with the proposed DLM"} section in \citep{EasyChair:7921}.

This episodic memory can be based on \textit{AKREM}, storing all the generated hierarchies.
Episodic memory's function is not only to store data. In our opinion, it can also be used for rearranging the concept memory to deal with multi-tasking. In other words, any new task requires the rearrangement of the whole concept memory, to cope with all previous tasks and the new one (consistency checking).

This episodic memory cannot create unlimited amount of memories. Therefore, a forgetting mechanism is essential. It might act like LIFO (last-in-first-out): new memories enter daily, and old mostly un-accessed ones are decay or deleted, to keep the memory fresh and with relevant information, and to respect the limited memory capacity. See Fig.~\ref{fig:Episodic_memory}(b).

Forgetting is also necessary, since in humans it helps reasoning and understanding by forcing them to generalize and abstract \citep{kuhl2007decreased}. It improves associations, relevancy, and extracting the right patterns in a given situation, since it differentiate data to be scaled from being most common to least common.

Moreover, forgetting is necessary due to the no-free-lunch principle \citep{adam2019no}, since any model cannot do best in all tasks, i.e. there is always some trade-off due to finite constraints, where it is better in some tasks while worse in others. Hence, adaptivity through forgetfulness allows for learning new tasks on-the-fly, thus putting all the weight of our current focus on the new learned task. This explains why the material after learning a course is so fresh and highly effective in preparing for exams. Meanwhile, after some period, due to the lack of using it, this material is partially forgotten, thus its utilization performance is reduced.

Finally, the base memory can consist not only from concept memory, but also from a procedural one. Because, they both are like tool memories, containing the basic elements from which not only events can be constructed, but also skills, recipies or any operational memory of how things are done.

\subsubsection{Wake-Sleep Periods}   \label{sec:wake_sleep}

\begin{figure}[H]%
\centering
\subfigure[Online Learning.]{%
	\label{fig:a}%
	\includegraphics[width=0.3\textwidth]{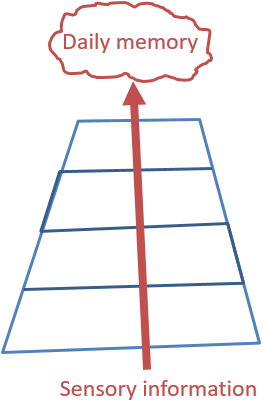}}
\hfill
\subfigure[Offline Learning]{%
	\label{fig:b}%
	\includegraphics[width=0.6\textwidth]{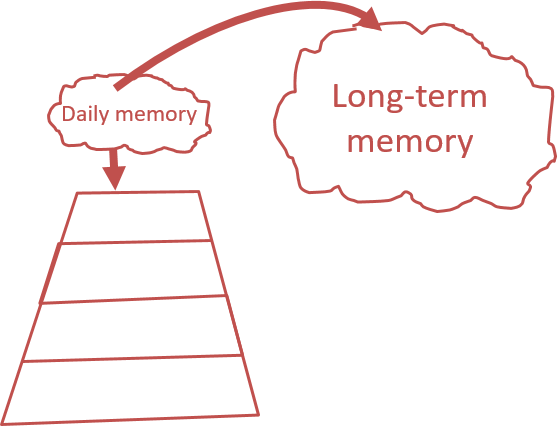}}%
\caption{Wake-Sleep Cycles}
\label{fig:online_offline_learning}
\end{figure}

In such a system, the learning occurs via two modes: waking and sleeping periods. 
During the waking period, the episodic daily memory acts like a capacitor, i.e., it is charged with events. It can also perform online learning, where a quick response is required, see Fig.~\ref{fig:online_offline_learning}(a).
At sleep, these events are discharged into LTM and perform a reorganization of the perceiving system or what is referred to as the concept and procedural memories, which are also function as the LTM. This complex and heavy learning can be regarded as offline learning, see Fig.~\ref{fig:online_offline_learning}(b).

This idea is supported by \textit{DL} \citep{liu2019human}
and by neuroscience, e.g. \citep{golden2022sleep}, 
where it is stated that \textit{“Sleep helps reorganize memories and presents them in the most efficient way”}.

In our opinion, at infancy, most of the cognitive effort is directed to the reorganization of the base memories, and less or not at all to store memories into LTM. It is because it is well known that infancy years are not remembered at adulthood. Our hypothesis is that it is because the concepts or the language by which we store these memories are not yet formalized or converged to be stable enough in order to be the same concepts as those used at adulthood. 

Another aspect can be deduced from the wake-sleep cycles:
different regimes of learning. In daytime we cannot process data too long, it must be fast real-time such as online-learning, for survival.
Whereas, in a relaxed or sleep mode, the brain is free to process memorized data.
The processing is 
of a sequence of data, which can infer cause-and-effect relationships for example.

Therefore, in survival mode we probably have access to only few state(s), representing perceived information in a sequence: few previous states (n) and few for prediction (m), due to limited resources.
While in relaxed mode we can use more memory and processing power, hence assume large $k>n$ and large $l>m$, which can produce more complex learning and abstractions.

\begin{figure}[H]
\centering
\includegraphics[width=0.7\textwidth]{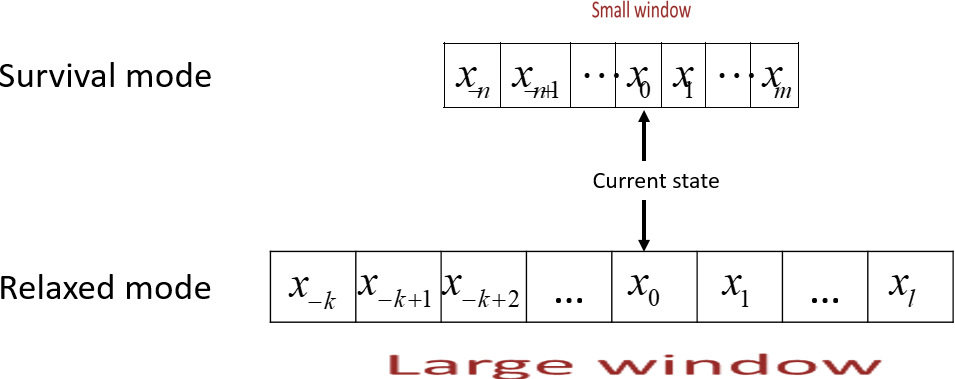}
\caption{Different modes}
\label{fig:different_modes}
\end{figure}

In other words, in relaxed mode we can learn from memory, based farther to the past (e.g. using many associations of past episodic events). And we can predict much farther to the future. See Fig.~\ref{fig:different_modes}.

In summary, learning can be divided into online and offline modes. In waking periods, i.e. when sensors are active, the learning is online and minimal since most resources are dedicated to fast response (e.g. fast optimization into local optimums). However, during sleeping periods, sensors are inactive, and previous memories can be used for improving models to make overall sense, e.g. larger time scale is used to generate causal relations in models. 
In such a case, it is a slow processing, to allow fast optimization or inference during waking periods. It is implemented e.g. as a 
Neural Architecture Search (\textit{NAS}) or a Genetic Algorithm (GA) \citep{binitha2012survey}, to get out of local minima and search for a global one.

\subsubsection{Human Development}

As we have seen in Section~\ref{sec:Episodic_memory}, the episodic and the base memories are interconnected, therefore the way I remember events is a function of the current concepts I have, or a function of the words that constructing event’s description. That is why I do not remember my infancy years. Because my concept world, or the language to describe events has changed drastically compared to the language I use as an adult. So both base memories and the episodic memory are changing frequently at infancy. 

The overall evolution is described as follows. The infant has no explicit guidance, but only some limited feedback. 
Hence, most of his learning occurs via unsupervised or even self-supervised mode.
At childhood he has some basic concepts that are last until adulthood therefore he can only vaguely recall childhood events. But still the concept world or the language is evolving in school from the various subjects that extend his concept world and change it significantly, such as Math, Sciences, and also social interactions.
Only at adulthood he can recall exact memories, with all the details, if it is not a function of natural forgetfulness.

From the assumption above we can propose that sleep functions as testing the best model considering continual learning, by testing it on all or recent memories in the episodic memory, to make sure it is consistent with all/most of them, see \url{https://medium.com/liecatcher/a-noise-model-for-why-we-sleep-and-lie-too-69bf723b3ba9}. It can be accomplished for example by sampling some of the recent memories, and performing self-supervised learning.

In parallel to the evolution of memories, there is also evolution of logic. Logic is about how consistent and comprehensible the world we experience. It is where we understand accurately how everything is operating and functioning in the world. More precisely, it is a symbolic reasoning. This logic is under-developed at infancy, and actually starts to evolve in school years, until we reach adulthood, and then it is stabilizes.

It may even be, that at infancy the episodic memory function is to organize data (e.g. at sleep), i.e. as a prediction tester, while at adulthood it is just a memory (aggregating specific events).

\begin{figure}[H]
\centering
\includegraphics[width=0.87\textwidth]{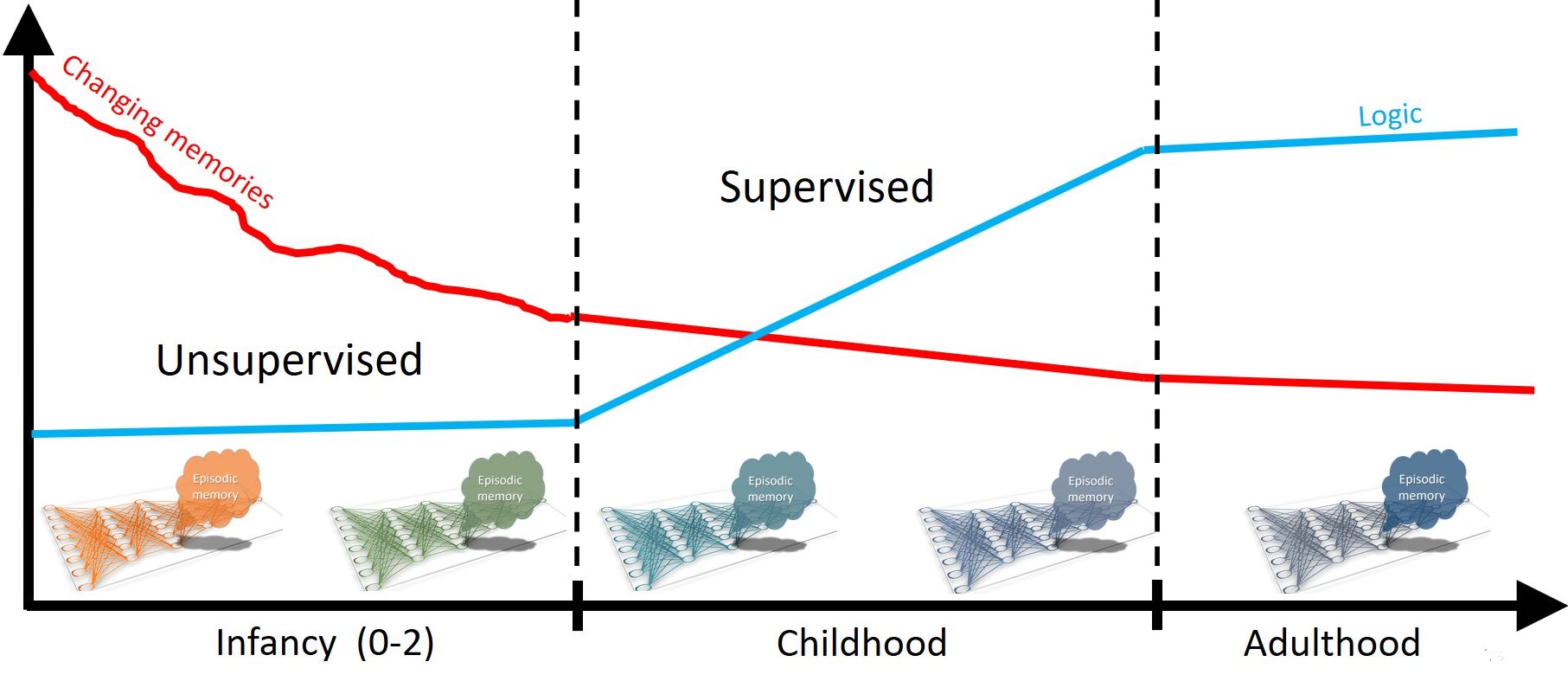}
\caption{Development plot}
\label{fig:development_plot}
\end{figure}

See this development process in Fig.~\ref{fig:development_plot}.

Finally, we assume that people perform by default unsupervised learning, while a supervised one is experienced only in guidance or tutoring.
Also, external supervised learning is performed initially via communication, and later it can be done via observation also (such as learning from videos, books, ...).

\subsubsection{Consolidation in model learning}

In model learning via \textit{DL} style, we start with random, uninterpretable features, and with time, these features become more symbolic, and perhaps represent concepts and actions. So with time the NN become more symbolic.
This idea can be presented in Figure~\ref{fig:developing}:
\begin{figure}[H]%
	\centering
	\subfigure[Human development]{%
		\label{fig:a}%
		\includegraphics[width=0.97\textwidth]{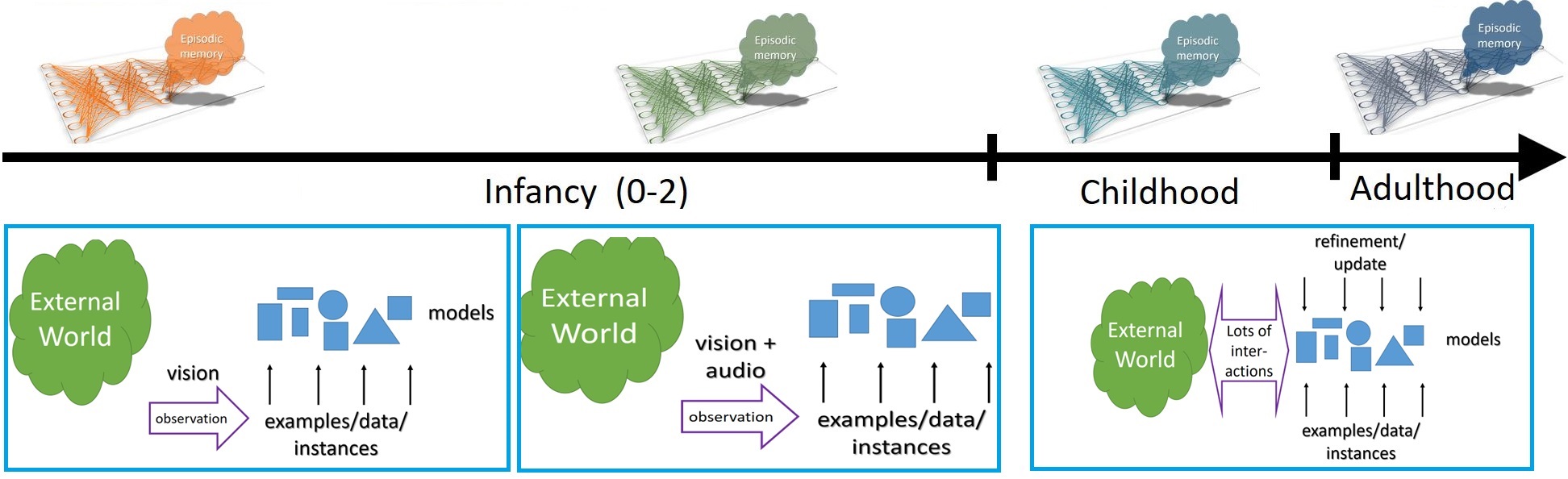}}
	\hfill
	\subfigure[Consolidation]{%
		\label{fig:b}%
		\includegraphics[width=0.63\textwidth]{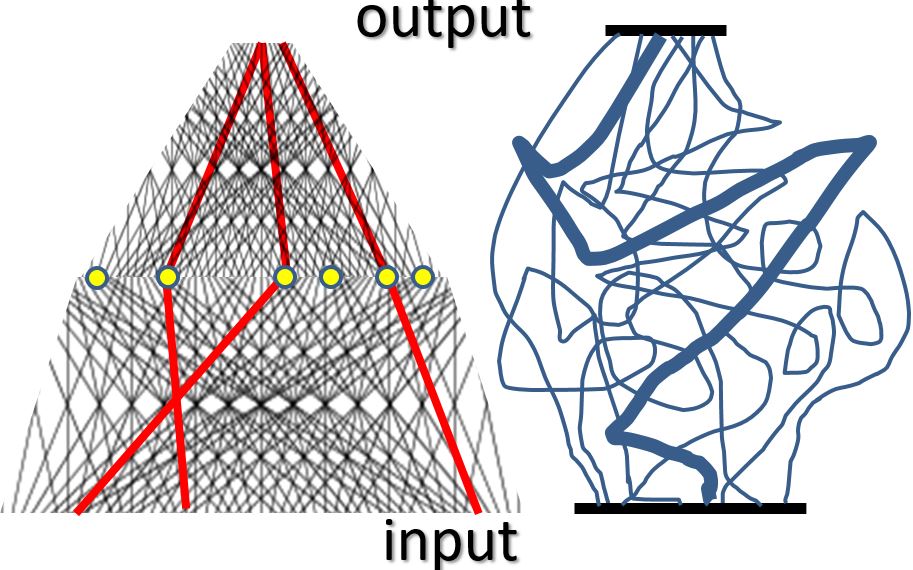}}%
	\caption{Evolution of cognition}
	\label{fig:developing}
\end{figure}

The transition from an uninterpretable representation of knowledge to a symbolic one is depicted in Figure~\ref{fig:developing}(a). In it we see the combination of the gradual interaction from infancy both from the sensory and the actuatory modules, lead to a more consolidated and less flexible knowledge.

So, we complete the developmental process of memories and logic, from Fig.~\ref{fig:development_plot}, with the inclusion of modeling development, from Figure~\ref{fig:top-down_and_bottom_up}. Resulting with a gradual evolution of interaction, starting from infancy, involving the sensory and the actuatory modules. We can see gradual extension of the infant tools, to allow more types of interaction, which mold and upgrade his models.

Interestingly, we can see in Figure~\ref{fig:developing}(a), that qualitatively, most of the changing in modeling occurs at infancy, then a bit less at childhood, and the least is at adulthood. While quantitatively (in the amount of knowledge), it is the opposite.

In Figure~\ref{fig:developing}(b) on the right, there is a DNN with consolidation  through sparsity, either over paths (in red) or/and over classes/objects in a layer (in yellow). On the left, it is a consolidation of an algorithm or a hypothesis out of many possible ones.
The sparsification in model searching is actually the manifestation of Occam's razor principal or the minimal description length, i.e., we look for the simplest algorithm as possible.
This constraint could be \textit{AGI} main's objective. It should, as we proposed, be searching for the fastest/smallest models, in order to have fast response for different situations, such as for survival, or for fewer steps of thought. 
Consequently, in the problem of combining \textit{DL} with rule-based approach, we can distinguish between soft-combination: where \textit{DL} consolidate into rule-based form, and a hard-combination: learnable model-structure right from the start.

Interestingly, similar consolidation mechanism were recently discovered in neuroscience \citep{fujimoto2023activity}, 
in what referred to as pruning. They discovered a protection/punishment machinery that would strengthen certain connections and kickoff the pruning of others, thus resulting with branches that get pruned to leave a single strong connection.

\subsection{Other implications of MOM}

\subsubsection{How Logic operations such as in/de-duction are possible in this framework?}  \label{sec:Logic}

As stated previously, in Section~\ref{sec:Learning_the_modeling}, induction and deduction may be part of the learning mechanism itself. Hence, either we treat it is prior knowledge applied in the \textit{AGI} system, or less preferred option: treating them as another operators, thus they can belong to a "logic" class, containing all relevant admissible operations of logic classes. Hence any logic class can naturally inherit these operations as part of being associated with the "logic" class.
However, for now, we assume these are not part of the learnable system, and considered as cognitive operators, separately from the learned knowledge operators.


To implement these operations we use HOL, or $\lambda$-calculus, used in semantics in NLP/NLU. It is simply the implementation of compositionality in language, i.e. where you can assign any object where there is slot for it.

For example, \textit{"I love you"} is simple $\lambda x,y.love(x,y)(I, you)$. But we could make it composed: \textit{"I love to create tools"} = $\lambda x,y.love(x,y)(I, \lambda x.create(x)(tools)$. In our case, it is simple \textit{OOP}, where we assign the "create" class to the y variable of "love" class, which is actually an action class.

Similarly, we could have compositional sentence: \textit{"I stand on a chair, that was built in a factory near my home" = stand(I, chair(origin(factory(near(home)))))}.


Another example of logic is via classes and instances, e.g. \textit{"All men are mortal. Peter Parker is a man. Therefore Peter Parker is mortal"}, is an example of updating temporarily (in WM) the "men" class to have mortal property, then creating instance of "men" called Peter Parker. Then following his object, finding that he is mortal due to inheritance. This is deductive reasoning: from general (class) to specifics (instance).

Nevertheless, even these examples demonstrate an obvious inference processes, they still require a guiding will, see \ref{sec:basic_wills}.

There is also the opposite reasoning: inductive, from specifics to general, e.g. Data: \textit{"All firetrucks I’ve seen are red"}. Hence, generalization: \textit{"Most firetrucks are red"}. This is induction directly from logic, and not from roar examples.

Another example, is in analogy tests, e.g. \textit{"Smile : mouth :: wink : eye"}, which as described earlier is accomplished by active thinking: recognizing the problem - analogy between two objects, then recognizing the source object (mouth), smile as its representing action, then searching for the same relation in the target object (eye), thus resulting with winking.
More broadly, this is an example of case-based reasoning \citep{fuchs2020case}, where source problem is used to solve a target one, by mapping and adapting source functionalities or actions to the target problem.\footnote{Based on category theory.}

The logic inference rules are discussed \citep{marquis2020guided}:
\begin{itemize}
	\item Contraposition: if A → B then NOT B → NOT A;
	\item Transitivity: if A → B and B → C then A → C;
	\item Simplification of disjunctive antecedents: if (A OR A') → B then A → B and A' → B.
\end{itemize}
Simplification is simply the function of "OR" action, while Transitivity is expressed via state-space as defined previously, where actions as vectors can be cascaded. That is, Transitivity is simply one of the features of vector state space (wills/actions).

Contraposition is expressed when problem-solving is applied backwards. It is actually how designing works: finding the problem of a goal or the cause to symptoms in health inspection. It is also referred to as abduction reasoning, which is about searching for explanations (cause) from incomplete observations (effect) and some background knowledge (such as rules).

In summary, we consider any-order logic as a stiff formalism and poor imitation of an actual flexible thinking. In our opinion it is a special framework, an abstract and mathematical one, that is learned in peoples' late years. It is not natural and fundamental as it might have being considered so far. For example, De Morgan's laws and the converting from first-order logic to conjunctive normal form (CNF) \citep{russell2010artificial} are not intuitive at all. Similar is the Resolution process and other things in logics. In our opinion, it should be simple mechanism, such as propagating in a network of states.

\subsubsection{Creativity}   \label{sec:Creativity}

\textit{MOM} enables creativity in two dimensions: either via non-admissible actions applied or via temporal abstractions.

The well known System 1 \& 2 \citep{daniel2017thinking}, combined with our attention mechanism as the projector of will, can explain how creativity sets into action.

It can be viewed by at least two processes. On the one hand, when we are contemplating about stuff while performing simple chore like walking or driving for example. In this case our attention is on the thinking process, hence it is less creative, since we use our consolidated regular logic. The less attended chore is move to the background.

On the other hand, in the opposite case, when we diverge our usual thinking and concentrate on a simple or usual chore - creativity pops in, when sudden ideas comes to mind, like during a shower\footnote{It can also be explain with quantum brain theory, which claims that thinking involves quantum effects. For example, it might just be that when we consciously searching for a solution, then we search it via classical way, i.e. via exponential growth, due to wave collapse. However, when we sleep or distracted, the search is via quantum superposition, hence it is exponential search. Therefore may yield much better solutions.}. In our framework, it is explained again with the attention. Since the simple chore is the one being focused, then the higher models are freed, this time to be handled by subconscious. This realm is not under our logical consolidated control, but it is free to "play" with the models as it wishes, without our constrained reality-based or physically-grounded thinking.
In neuroscience, there is default mode network (DMN) \citep{buckner2022brain} 
which is activated when we do not do anything, or mind-wandering, and it shows that some brain regions instead of being less active, they become more active, which may indicate self organization or creativity pops in, to produce new connections and more. 

This could be explained in other words also.
As it is well known, creativity resides in the right hemisphere, where the thinking is more divergent. The creativity comes to play mostly when we are less looking for patterns and categorization \citep{beaty2015default}. I.e., we can be biased or stuck in past patterns, when we recognize them, to find a solution to a problem. It can happen both in System 1 \& 2, where we look for the best model to resolve the current situation, as in a recognition process.
But sometimes  
we should dismiss the familiar, in order to find novel ways to solve problems \url{https://www.thinkwrongbook.com/}. In our case, it might be  
about bypassing the automatic recognition mode, and search for a more appropriate solution, via a more thorough search in different models, ending up with generating new model for the given problem.

Overall, we can unite will with attention, by stating that will projecting via attention "beacons". Hence, similarly as we have main will and sub wills, or wills with different intensity - similarly our attention is splitted into parallel beams, where there is the main one and smaller ones. 
For example, in hearing a message/story, we are mostly attentive to the incoming message in sequence, but a smaller attention is applied in stitching the different parts of the story together, as if we constructing a puzzle. Similarly in the examples above, there is main attention (e.g. thinking while walking, showering, etc) and the secondary attention (e.g. walking). This splitting of attention to parallel processing is important for survival. Because sometimes we have no choice but to handle several stuff simultaneously.
For example, to handle distractions, or as a selective attention between different sensors (such as hearing and seeing an object), or also within each sensor: as in vision we first grasp multiple objects, then concentrate only on few of them. 
Or as an alternating attention by switching from one thing to another, see \citep{abdalaziz2023rhythmic} 
about parallel multi-tasking, or as a dividing attention by focusing on several things at once, such as driving and talking at the same time. However, splitting attention has more preference on parallel non-interfering splittings instead of ones in conflict, such as splitting within the same modality.



More generally, attention splitting can be used in any problem solving, as described above in the example of message/story comprehension, by concentrating of "processing units" over the incoming objects to perform inference for filling missing information, but also for making overall sense (connect between different objects). This split of attention supported also by neuroscience, e.g. in \citep{abdalaziz2023rhythmic}, experienced memorizing of several elements simultaneously, and in \citep{howarth2012updated} 
showing an example of mental load attention splitting, e.g. between visual and audio sensory perception.

Finally, we can argue philosophically, that the attention as a top-down process revealing our will, is actually a separated entity from the physical realm that interacts with the physical brain constantly, as long as we are alive.
This idea can even have evidence from neuroscience, such as here: \url{https://iai.tv/articles/brain-scans-tell-us-nothing-about-conciousness-auid-2514}: describing a battle between physicalism (consciousness is in the physical brain) and idealism (the experience is separated from its partial expression in the physical mind), similar to person looking at me, sees only the front side of me, but not the back. Another paper \citep{pinotsis2023cytoelectric}, can only unintentionally hint this idea. It describes \textit{"Cytoelectric Coupling"}: arguing that the brain’s electrical fields, created by neural network activity, can influence the physical configuration of neurons’ sub-cellular components to optimize network stability and efficiency. In another perspective we can treat electrical brain waves as the closest side of the soul, and thus view the process described above as the soul influence the brain physical activity.

\subsubsection{Information}

Our \textit{MOM} has some basic axioms, regarding how information is interpreted.

\subsection{\textit{AGI} additional characteristics}

Here are a few small comments about the \textit{AGI} system to be constructed.

First, our opinion, is that the research of explainable AI (or XAI) is not totally correct in its core. We believe that DNN in its nature should not be explainable at all. \textbf{It is merely the vessel of intelligence, while logic and explainability are the products of intelligence.} There is no rule to require these human outcomes to reside in the cognitive system in the first place. We advocate that these two concepts to be accurately defined and separated.

Moreover, as mentioned in Section~\ref{sec:problem_solving}, many generative models that are used for explanation, e.g. Principal Component Analysis (PCA) \citep{abdi2010principal} or sparse/denoising auto-encoders \citep{bank2020autoencoders}, are themselves black-boxes. Also, most of \textit{DL} explainability methods are very narrow type of explainability, e.g. merely feature relative significance.

Next, we do not see the necessity for \textit{AGI} or any AI to be conscious. We do not comprehend why there is so much talk around this (in our opinion) insignificant topic.
\textit{AGI} should not emulate humans at all nor to be autonomous. It should only be a data processing machine, as regular PC\footnote{Personal Computer}, to comprehend natural language, learn, and accomplish tasks.
The actual meaning of things, in human perspective, should be left to humans. 

Next, \textit{AGI} should not be about strong computational power as it is nowadays. We see evidence for this in our own intelligence: we use papers to store and elaborate our ideas and thoughts, presentations, drawings, 2D/3D (interactive) visualizations, simulations, etc. Similarly it is in current generative AI: the various action GPT tools like AutoGPT, AgentGPT, BabyAGI, MetaGPT, LETI, ToolLLM, WebGLM and more - which use external applications and sources. All these are tools extending our limited capabilities, without the need to extend ourselves. Just as we use calculators to perform hard calculations, or planes to fly, the \textit{AGI} should function similarly. It should not be fast at arithmetic calculations, but rather it should be good at thinking and solving problems, while provided various tools for its disposal, as it is for humans. And similar to humans, it can create the necessary tools/technology for solving problems. See for example in \url{https://www.lesswrong.com/posts/K4urTDkBbtNuLivJx/why-i-think-strong-general-ai-is-coming-soon}.
Also see \url{https://clivethompson.medium.com/the-weird-power-of-transactive-memory-7c7e324c9425} about how we extend our memory by socially distribute skills and knowledge among members in a close community.

Next, as we embrace the holistic perspective for designing \textit{AGI}, similarly we encourage implementing both self-supervised learning and RL in the same system.
We argue here, that model learning should encapsulate these both regimes, since part of the modeling is to learn within interactive setup. Hence, there is no need for a reward to be gathered from the environment, since internal "reward" or intrinsic motivation is sufficient and represents the current will. Meanwhile the feedback is gathered simply from the sensors, i.e. the system is always stays open-loop, but during interactions it becomes (or it is considered as) closed-loop, similarly as RL. We can see these two regimes in Fig.~\ref{fig:Will_derivation_examples}.

Next, we think that current meta-learning in \textit{DL}/RL is not fit to serve as a tool for \textit{AGI}, if we consider having shared parameters learned between tasks. Because most of the skills we learn are not related to previous tasks at all. Hence it should be progressive in learning.

Next, our purpose in AI is to reach the most natural language model (LM) to perform tasks for us. However, reaching this goal requires very slow and gradual process. Thus, the evolution of AI research so far can be described in the following. 
It started with low-level machine programming code (assembly), then moved on to high-level programming languages (e.g. C\#, Python, Java), then moved to NLP LLMs such as GPT3 and DALL-E2 and others that use prompts, which are less "programmitive", but still require practice and expertise in writing the most appropriate prompts to get the desired response.
Hence, our model suggest that \textit{AGI} agent must be learned via human interaction, to learn the alignment of our wills 
from the way we express it linguistically, e.g. like a teacher demonstrates or behave in-front of his student. In other words, unlike the batch learning usual policy costumed in \textit{DL}, where the AI agent is "shoved"/bombarded with a huge amount of data, we encourage a more humane treatment - a gradual and progressive learning (via online and offline modes, but with incremental learning).

Additionally, for constructing \textit{AGI} framework we practice both the bottom-up approach (neuroscience, mainly based on Jeff Hawkins) and top-down approach (psychology, mainly based on Daniel Kahneman).
In general, this should be the duality principle we encourage in \textit{AGI} research. That is, embracing the combination of different disciplines together, to capture intelligence's many facets and features.

Finally, as computational models (automata) \citep{chater2008computational} have shown, in order to achieve a Turing machine model \citep{de2018turing}, we must have unlimited memory and unlimited access to it. Hence the \textit{AGI} model should also have these capabilities. Our proposed cognitive model express/reveals these capabilities via the use of LTM (past knowledge) and WM (current memory).


\bibliography{Shimon_paper02,Shimon5}
\end{document}